# Comprehensive Survey of Complex-Valued Neural Networks: Insights into Backpropagation and Activation Functions

M. M. Hammad

Department of Mathematics and Computer Science, Faculty of Science, Damanhour University, Egypt
Email: m_hammad@sci.dmu.edu.eg
https://orcid.org/0000-0003-0306-9719

**Abstract**

Artificial neural networks (ANNs), particularly those employing deep learning models, have found widespread application in fields such as computer vision, signal processing, and wireless communications, where complex numbers are crucial. Despite the prevailing use of real-number implementations in current ANN frameworks, there is a growing interest in developing ANNs that utilize complex numbers. This paper presents a comprehensive survey of recent advancements in complex-valued neural networks (CVNNs), focusing on their activation functions (AFs) and learning algorithms. We delve into the extension of the backpropagation algorithm to the complex domain, which enables the training of neural networks with complex-valued inputs, weights, AFs, and outputs. This survey considers three complex backpropagation algorithms: the complex derivative approach, the partial derivatives approach, and algorithms incorporating the Cauchy-Riemann equations. A significant challenge in CVNN design is the identification of suitable nonlinear Complex Valued Activation Functions (CVAFs), due to the conflict between boundedness and differentiability over the entire complex plane as stated by Liouville's theorem. We examine both fully complex AFs, which strive for boundedness and differentiability, and split AFs, which offer a practical compromise despite not preserving analyticity. This review provides an in-depth analysis of various CVAFs essential for constructing effective CVNNs. Moreover, this survey not only offers a comprehensive overview of the current state of CVNNs but also contributes to ongoing research and development by introducing a new set of CVAFs (fully complex, split and complex amplitude-phase AFs).

**Abbreviations:**

| | | | |
|---|---|---|---|
| AF | Activation Function | DNN | Deep Neural Network |
| APTF | Amplitude-Phase-Type Function | ELU | Exponential Linear Unit |
| APSF | Amplitude-Phase Sigmoidal Function | ETF | Elementary Transcendental Functions |
| BN | Batch Normalization | FC-Swish | Fully Complex Swish |
| CBP | Complex Backpropagation | FC-Mish | Fully Complex Mish |
| CVAF | Complex-Valued Activation Function | LReLU | Leaky Rectified Linear Unit |
| CVNN | Complex-Valued Neural Network | NN | Neural Network |
| CAP-PLS | Complex Amplitude-Phase Piecewise Linear Scaling | PReLU | Parametric Rectified Linear Unit |
| CAP-ES | Complex Amplitude-Phase Exponential Scaling | RVNN | Real Valued Neural Network |
| CAP-ArcTanS | Complex Amplitude-Phase ArcTan Scaling | ReLU | Rectified Linear Unit |
| CAP-Swish | Complex Amplitude-Phase Swish | RReLU | Randomized Rectified Linear Unit |
| CAP-ELU | Complex Amplitude-Phase Exponential Linear Unit | Split-STanh | Split-Sigmoidal Tanh |
| CAP-Softplus | Complex Amplitude-Phase Softplus | Split-PSigmoid | Split-Parametric Sigmoid |
| CAP-ErfA | Complex Amplitude-Phase Erf Attenuation | Split-QAM | Split-Quadrature Amplitude Modulation |

## 1. Introduction

Real-valued neural networks (RVNNs) are the most commonly used type of neural network (NN) in machine learning and artificial intelligence applications, primarily because all their parameters, inputs, and outputs are real numbers. This widespread usage can be attributed to their simplicity and well-understood mathematics, making RVNNs easier to design, implement, and troubleshoot. Additionally, operations with real numbers typically require less computational power compared to complex numbers, leading to faster training times and the ability to handle larger datasets. The vast majority of existing NN libraries and frameworks are optimized for real-valued operations, providing developers with a wide range of tools, pre-trained models, and resources tailored for RVNNs. Furthermore, RVNNs are versatile and can be applied to various domains, including image recognition, natural language processing, and speech recognition, making them a go-to choice for many machine learning tasks. Various RVNN architectures have been developed to learn abstract features from data, including multilayer perceptron, recurrent neural networks, and convolutional neural networks [1-12]. Critical aspects of RVNNs include weight initialization strategies, the formulation of loss functions, overfitting prevention techniques, learning rate schedules, and adaptive optimization algorithms [1-12].

Complex-valued neural networks (CVNNs) operate with complex numbers [13-18] for inputs, network parameters, and potentially outputs. These NNs are not as widely adopted as RVNNs for several reasons. Operations involving complex numbers are inherently more computationally intensive, leading to longer training times and higher resource consumption. Implementing and debugging CVNNs can also be more challenging due to the added complexity of dealing with both real and imaginary parts, requiring more sophisticated mathematical tools and a deeper understanding of the underlying principles. While there are some libraries and frameworks that support CVNNs, they are not as mature or widely used as those for RVNNs, which can limit the availability of pre-trained models, documentation, and community support.

CVNNs have shown considerable promise in outperforming their real-valued counterparts in various applications, notably in speech enhancement, image processing, signal processing, robotics, bioinformatics, sonar, radar, and quantum computing. This superiority primarily stems from their ability to handle and process complex-valued data, which is inherently present in these fields. Looking ahead, several factors point towards a promising future for CVNNs. As hardware technology advances, particularly in quantum computing and specialized processors, CVNNs could experience significant boosts in performance and efficiency. Algorithmic innovations are also anticipated, with ongoing research into more sophisticated learning algorithms, activation functions (AFs), and network architectures likely to yield even more powerful CVNNs. Additionally, hybrid models integrating CVNNs with RVNNs could leverage the strengths of both approaches, providing even more robust solutions to complex problems.

Over the years, researchers have made substantial strides in improving the learning algorithms tailored for CVNNs. This includes adaptations of backpropagation and optimization techniques that consider the complex nature of the weights and activations in these networks [19-30]. Complex backpropagation (CBP) and gradient descent methods designed to handle the split-complex nature of derivatives in the complex domain, have been proposed to ensure convergence and stability during training. Additionally, significant efforts have been directed towards developing suitable AFs for CVNNs [31-76]. Traditional AFs like ReLU and Sigmoid do not directly apply to complex numbers. Consequently, researchers have introduced complex-valued versions of AFs, such as complex Tanh and split AFs, which separately process the real and imaginary parts. Advanced AFs incorporating phase-based activations further help preserve and utilize phase information effectively.

Why Are CVNNs Inevitable? A complex number [13-18] has both a real part and an imaginary part, and this added dimension can be beneficial in certain contexts. CVNNs can have better convergence properties and robustness in training, especially for data naturally represented in the complex domain. Additionally, CVNNs can naturally handle data where phase and amplitude information are critical, such as in electromagnetic signal processing, and wave-related information because they can directly handle the phase and amplitude components of wave signals. Moreover, the main reason for advocating CVNNs stems from the unique properties of complex number arithmetic, particularly the multiplication operation. When complex numbers are multiplied, the result involves a phase rotation and amplitude modulation. This inherent characteristic of complex multiplication offers a significant reduction in the degrees of freedom required in the network. For example, in wave signals, the phase components represent the time course or positional differences, while the amplitude denotes the energy or power of the wave. The phase rotation and amplitude attenuation features embedded in complex numbers provide significant advantages in this context. Phase rotation, a property unique to complex numbers, allows CVNNs to naturally model and adjust for phase shifts in wave signals. Amplitude attenuation, another property of complex numbers, helps in accurately capturing the variations in wave intensity, which is essential for understanding the power and energy distribution in the signal. These features lead to more efficient and effective learning, as CVNNs can model wave phenomena more accurately by integrating phase and amplitude directly into the network parameters. This not only enhances the network's ability to learn from wave data but also improves its generalization performance.

Moreover, RVNNs have achieved extraordinary performance in the deep learning era, yet the capabilities of individual real-valued neurons remain limited. A notable example of this limitation is the XOR problem, a classic problem in NN theory. A single real-valued neuron cannot solve the XOR problem because it cannot create a non-linear decision boundary that correctly classifies the inputs. This limitation underscores a fundamental constraint of RVNNs in certain contexts. In contrast, a single complex-valued neuron can solve the XOR problem with a high degree of generalization. This capability is rooted in the unique properties of complex-valued neurons, as highlighted by Nitta's research [42-44]. Nitta introduces the concept of an orthogonal decision boundary in complex-valued neurons. In his study, two hypersurfaces intersect orthogonally, dividing the decision boundary into four distinct regions. This orthogonal property enables complex-valued neurons to handle more complex decision boundaries, which are necessary for solving problems like XOR. Additionally, the orthogonal decision boundary enhances the computational power of CVNNs. By leveraging the orthogonal property, CVNNs can address issues related to symmetry detection more effectively. This means that CVNNs can distinguish between symmetric patterns in data with greater accuracy, a task that can be challenging for RVNNs. The potential computational power of CVNNs, therefore, lies in their ability to create and manipulate complex decision boundaries using the inherent properties of complex numbers. This allows them to solve problems that are inherently non-linear and cannot be addressed by single real-valued neurons.

The classical approach to handling complex-valued signals involves splitting each complex-valued signal into two real-valued signals, typically into either the real and imaginary components or the magnitude and phase components. These real-valued representations are then processed using RVNNs. In this split complex-valued network, real-valued AFs and weights are employed to estimate the network parameters. However, this method has significant limitations. One of the primary issues is that the gradients used to update the network's free parameters do not represent the true complex-valued gradients. This discrepancy arises because real-valued networks cannot fully capture the interdependence between the real and imaginary parts of the signals or the intricate relationships between magnitude and phase. As a result, the approximation of complex-valued functions, especially the phase of the complex-valued signals, is often poor. Phase information is critical in many applications, such as communications and signal processing, where the accuracy of phase representation can significantly impact performance. The inability of split complex-valued networks to accurately capture complex-valued gradients and functions clearly indicates the need for developing fully CVNNs and their associated learning algorithms. CVNNs use Complex-Valued Activation Function (CVAFs) and complex weights, allowing them to directly handle the intricacies of complex-valued data. By working within the complex domain, these networks can better capture the relationships between the real and imaginary components and more accurately model phase information. However, fully CVNNs increase the computation complexity and face difficulty in obtaining complex-differentiable (or holomorphic) AF [20]. In contrast, CVNNs with complex-valued weights and a real-valued AF offer a solution to the phase distortion challenge. This approach, exemplified in [22], has found widespread application in various domains due to its compatibility with popular real-valued AFs [11], including ReLU, Sigmoid, and Tanh.

The development of CVNNs follows two main approaches, each addressing the challenges posed by processing complex-valued signals in different ways. The first approach seeks fully CVAFs that must navigate the conflicting requirements of boundedness and differentiability inherent in complex functions. This conflict is encapsulated in Louiville's theorem, which states that any bounded entire function (a function that is analytic, or differentiable) at every point in the complex plane $\mathbb{C}$ must be constant. Some notable fully CVAFs include the fully complex Tanh function [21, 23], the fully complex Logistic-Sigmoidal function [19], fully complex Elementary Transcendental Functions (ETFs) [23, 73, 74], and the fully complex exponential function [16, 77]. To address the unbounded nature of analytic functions in $\mathbb{C}$, researchers like Georgiou and Koutsougeras [20], and Hirose [30, 35, 62], have proposed fully CVAFs that normalize or scale the amplitude of complex signals while preserving their phase. However, these functions are limited by their inability to learn phase variations between the input and the target, a critical requirement in signal processing applications such as channel equalization or deconvolution, where both amplitude and phase information from multiple signal sources need to be resolved. In contrast, the second approach uses 'split' CVAFs, where standard real-valued AFs, such as Sigmoid [43-53] or Tanh [54-57], are applied separately to the real and imaginary parts of the complex inputs. This method circumvents the boundedness issue described by Louiville's theorem but results in AFs that are not analytic. Despite this inherent deficiency, the adaptable structure and simplicity of split CVAFs have garnered significant interest. For instance, Leung and Haylun [19], as well as Benvenuto et al. [41], have demonstrated the use of a pair of real hyperbolic tangent Sigmoid functions as a split CVAF in feedforward neural networks. This approach, although not preserving analyticity, offers a practical compromise by leveraging well-established real-valued AFs to handle complex-valued inputs, thus maintaining simplicity and adaptability in network design. Other notable split-type AFs include the split step function [42], split parametric Sigmoid [22, 23], split Sigmoid Tanh [58], split hard Tanh [59] and CReLU [60, 61].

The paper is structured to provide a comprehensive survey of CVNNs, focusing on their AFs and learning algorithms. It begins with Section 2, which introduces complex numbers and complex functions, including visualizations to aid in understanding their geometric and algebraic properties. Section 3 delves into complex calculus, specifically holomorphic or analytic functions and Wirtinger derivatives, which are essential for grasping the theoretical foundations of CVNNs. Section 4 categorizes and discusses various types of CVNNs, distinguishing between fully CVNNs in Section 4.1 and split CVNNs in Section 4.2. Section 5 examines CBP algorithms, presenting three distinct methods based on different assumptions about the existence and properties of derivatives of AF. Section 6 explores properties such as boundedness and differentiability, and their implications. Section 7 provides an in-depth examination of various CVAFs, including both established and newly introduced ones, highlighting their roles in constructing effective CVNNs. Finally, a new set of CVAFs is introduced in Section 8.

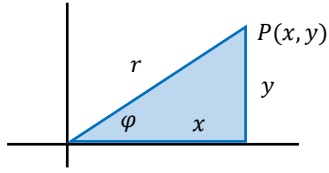

**Fig. 1.** Complex Numbers as Position Vectors in the Complex Plane. The complex number $z = x + iy$ is shown as a vector originating from the origin $(0,0)$ and terminating at the point $P(x,y)$. Here, $x$ represents the real part and $y$ represents the imaginary part of the complex number. The length of the vector, denoted as $r$, is the magnitude of the complex number and is calculated as $r = \sqrt{x^2 + y^2}$. The angle $\varphi$ represents the argument of the complex number, which is the angle between the positive real axis and the vector, measured in the counterclockwise direction. This polar representation helps in visualizing and performing operations on complex numbers geometrically.

## 2. Complex Numbers, Complex Functions, and Their Visualizations

In this section, we introduce the fundamental concepts of complex numbers, complex functions, and their visualizations. These foundational topics are essential for comprehending the advanced ideas, such as CVAFs and CBP of CVNNs, presented in this survey article.

A complex number [13-15] is of the form $a + ib$, where $a$ and $b$ are real numbers, and $i$ is the imaginary unit satisfying $i^2 = -1$, or $i^2 + 1 = 0$. The relation $i^2 + 1 = 0$ induces several equalities for any integer $k$:

$$i^{4k} = 1, i^{4k+1} = i, i^{4k+2} = -1, \text{ and } i^{4k+3} = -i. \tag{1}$$

The real part of a complex number $z = a + ib$ is denoted by $\text{Re}(z)$, or $\Re(z)$, and the imaginary part is denoted by $\text{Im}(z)$ or $\Im(z)$. For example, $\Re(3 + 2i) = 3$ and $\Im(3 + 2i) = 2$.

A complex number $z$ can be represented as an ordered pair $(\Re(z), \Im(z))$ of real numbers. This ordered pair can be interpreted as the coordinates of a point in a two-dimensional space. The complex plane is a two-dimensional plane where the horizontal axis represents the real part, $\Re(z)$, and the vertical axis represents the imaginary part, $\Im(z)$. The complex plane is also referred to as the Argand diagram. To each complex number, there corresponds one and only one point in the plane, and conversely to each point in the plane there corresponds one and only one complex number. Additionally, a complex number $z = a + ib$ can be viewed as a position vector in the complex plane, starting from the origin $(0,0)$ and ending at the point $(a, b)$. The vector represents both the magnitude (distance from the origin) and the direction (angle with the positive real axis), see Fig. 1.

When performing mathematical operations on complex numbers, we treat the real and imaginary parts separately. Here are the basic operations: Two complex numbers $a = x + iy$ and $b = u + iv$ are added by separately adding their real and imaginary parts. That is to say:

$$a + b = (x + iy) + (u + iv) = (x + u) + i(y + v). \tag{2}$$

This geometrically corresponds to moving from the origin to the point $a$, and then from there, moving to the point $b$. The resulting complex number $a + b$ is the diagonal of the parallelogram formed by the vectors representing $a$ and $b$. This method of adding complex numbers aligns with the geometric interpretation of complex addition, where the real parts represent movement along the horizontal axis, and the imaginary parts represent movement along the vertical axis in the complex plane, as illustrated in Fig. 2. Similarly, subtraction can be performed as

$$a - b = (x + iy) - (u + iv) = (x - u) + i(y - v). \tag{3}$$

Multiplication of a complex number $a = x + iy$ and a real number $r$ can be done similarly by multiplying separately $r$ and the real and imaginary parts of $a$:

$$r\,a = r(x + iy) = rx + iry. \tag{4}$$

The multiplication of two complex numbers can be performed by applying the distributive property, the commutative properties of addition and multiplication, and the defining property $i^2 = -1$. Consequently, we have:

$$(x + iy)(u + iv) = (xu - yv) + i(xv + yu). \tag{5}$$

In particular,

$$(x + iy)^2 = x^2 - y^2 + 2ixy. \tag{6}$$

An alternative to Cartesian coordinates in the complex plane is the polar coordinate system. Let $P$ be a point in the complex plane corresponding to the complex number $(x, y)$ or $x + iy$. Then we see from Fig. 1 that

$$x = r\cos\varphi \quad \text{and} \quad y = r\sin\varphi. \tag{7}$$

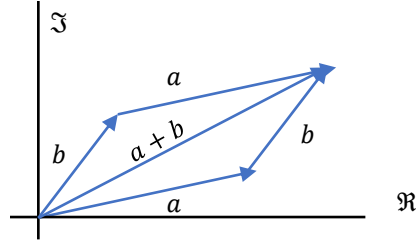

**Fig. 2.** Geometric Addition of Complex Numbers Using a Parallelogram. The complex numbers $a$ and $b$ are represented as vectors originating from the origin, with $a$ terminating at $(x, y)$ and $b$ terminating at $(u, v)$. To add these two complex numbers geometrically, a parallelogram is constructed with $a$ and $b$ as adjacent sides. The vector representing $a$ is drawn from the origin, and the vector representing $b$ is drawn starting from the endpoint of $a$. Similarly, a vector equal to $a$ is drawn starting from the endpoint of $b$. The opposite sides of the parallelogram are parallel and equal in length. The resultant vector, which represents the sum $a + b$, is the diagonal of the parallelogram originating from the origin. It terminates at the point where the opposite corners of the parallelogram meet. This point has coordinates $(x + u, y + v)$, which corresponds to the sum of the real and imaginary parts of the two complex numbers.

In polar coordinates, a complex number $z = x + iy$ is represented by its distance from the origin ($r = |z|$) and the angle $\varphi$ it makes with the positive real axis in a counterclockwise direction. The polar form of a complex number $z$ is given by

$$z = x + iy = r(\cos\varphi + i\sin\varphi), \tag{8}$$

where $r$ is the absolute value of $z$, and $\varphi$ is the argument (angle) of $z$.

For any complex number $z \neq 0$, there corresponds only one value of $\varphi$ in $0 \leq \varphi < 2\pi$. However, any other interval of length $2\pi$, for example $-\pi < \varphi \leq \pi$, can be used. Any particular choice, decided upon in advance, is called the principal range, and the value of $\varphi$ is called its principal value.

The absolute value of a complex number $z = x + iy$ is given by

$$r = |z| = \sqrt{x^2 + y^2}. \tag{9}$$

The absolute value of a complex number, as given by Pythagoras' theorem, represents the distance from the origin to the point representing the complex number in the complex plane.

The argument of a complex number $z = x + iy$ is the angle $\varphi$ that the line segment from the origin to the point $(x, y)$ makes with the positive real axis. To calculate the argument, you can use the arctangent function. The argument (written as $\arg z$) can be found using the formula:

$$\arg z = \varphi = \tan^{-1}\left(\frac{y}{x}\right), \tag{10}$$

with

$$\varphi = \arg(x + iy) = \begin{cases} \tan^{-1}(y/x), & x > 0, \text{right half plane} \\ \tan^{-1}(y/x) + \pi, & x < 0, y \geq 0, \text{upper left half plane} \\ \tan^{-1}(y/x) - \pi, & x < 0, y < 0, \text{lower left half plane} \\ \pi/2, & x = 0, y > 0, +i \text{ axis} \\ -\pi/2, & x = 0, y < 0, -i \text{ axis} \\ \text{undefined} & x = 0, y = 0, \text{origin} \end{cases}. \tag{11}$$

Let $z_1 = x_1 + iy_1 = r_1(\cos\varphi_1 + i\sin\varphi_1)$ and $z_2 = x_2 + iy_2 = r_2(\cos\varphi_2 + i\sin\varphi_2)$, because of the trigonometric identities

$$\cos a \cos b - \sin a \sin b = \cos(a + b), \tag{12.1}$$
$$\cos a \sin b + \sin a \cos b = \sin(a + b), \tag{12.2}$$

we may derive (De Moivre's Theorem:)

$$z_1 z_2 = r_1 r_2 \{\cos(\varphi_1 + \varphi_2) + i\sin(\varphi_1 + \varphi_2)\}, \tag{13.1}$$
$$\frac{z_1}{z_2} = \frac{r_1}{r_2}\{\cos(\varphi_1 - \varphi_2) + i\sin(\varphi_1 - \varphi_2)\}. \tag{13.2}$$

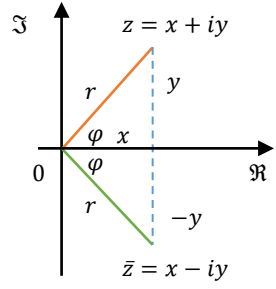

**Fig. 3.** Geometric Representation of $z$ and Its Conjugate $\bar{z}$ in the Complex Plane. The complex number $z = x + iy$ is depicted as a vector originating from the origin $(0,0)$ and terminating at the point $(x, y)$, where $x$ is the real part and $y$ is the imaginary part of $z$. The magnitude of $z$ is represented by $r$, which is calculated as $r = \sqrt{x^2 + y^2}$. The angle $\varphi$ is the argument of $z$, measured counterclockwise from the positive real axis. The conjugate of $z$, denoted as $\bar{z} = x - iy$, is also represented as a vector originating from the origin and terminating at the point $(x, -y)$. This reflects the fact that the conjugate has the same real part $x$ but the opposite sign for the imaginary part, $-y$. The magnitude $r$ and the angle $\varphi$ remain the same for both $z$ and $\bar{z}$, as the magnitudes are equal, and the angles are reflections over the real axis.

By assuming that the infinite series expansion $e^x = 1 + x + \frac{x^2}{2!} + \frac{x^3}{3!} + \frac{x^4}{4!} + \cdots$ of elementary calculus holds when $x = i\varphi$, we can arrive at the result

$$e^{i\varphi} = 1 + (i\varphi) + \frac{(i\varphi)^2}{2!} + \frac{(i\varphi)^3}{3!} + \frac{(i\varphi)^4}{4!} + \frac{(i\varphi)^5}{5!} + \frac{(i\varphi)^6}{6!} + \frac{(i\varphi)^7}{7!} \cdots$$
$$= \left(1 - \frac{\varphi^2}{2!} + \frac{\varphi^4}{4!} - \frac{\varphi^6}{6!} + \cdots\right) + i\left(\varphi - \frac{\varphi^3}{3!} + \frac{\varphi^5}{5!} - \frac{\varphi^7}{7!} \cdots\right) = \cos\varphi + i\sin\varphi, \tag{14}$$

which is called Euler's formula, for any real number $\varphi$. The functional equation implies thus that, if $x$ and $y$ are real, one has

$$e^z = e^{x+iy} = e^x e^{iy} = e^x \cos y + i e^x \sin y, \tag{15}$$

which is the decomposition of the exponential function into its real and imaginary parts.

For each $z \neq 0$, there are infinitely many possible values for $\arg z$, which all differ from each other by an integer multiple of $2\pi$.

$$e^{i(\varphi + 2k\pi)} = \cos(\varphi + 2k\pi) + i\sin(\varphi + 2k\pi) = \cos(\varphi) + i\sin(\varphi) = e^{i\varphi}, \quad k = 0, +1, +2, \ldots. \tag{16}$$

The conjugate of a complex number $z = x + iy$ is denoted by $\bar{z}$ or $z^*$, and it is obtained by changing the sign of the imaginary part, Fig. 3. Mathematically, the conjugate is given by: $x - iy$. Similarly, if $z = x - iy$, then $\bar{z} = x + iy$. The following are key properties of the complex conjugate:

$$\overline{(\bar{z})} = z, \tag{17.1}$$
$$\overline{z_1 \pm z_2} = \bar{z}_1 \pm \bar{z}_2, \tag{17.2}$$
$$\overline{z_1 \cdot z_2} = \bar{z}_1 \cdot \bar{z}_2, \tag{17.3}$$
$$\overline{z_1/z_2} = \bar{z}_1/\bar{z}_2. \tag{17.4}$$

The conjugate is useful in various mathematical operations, including finding the modulus (absolute value) of a complex number, dividing complex numbers, and simplifying expressions involving complex conjugates. Here are the basic operations:

- The reflection leaves both the real part and the magnitude of $z$ unchanged, that is

$$\Re(\bar{z}) = \Re(z) \quad \text{and} \quad |\bar{z}| = |z|. \tag{18}$$

- The imaginary part and the argument of a complex number $z$ change their sign under conjugation

$$\Im(\bar{z}) = -\Im(z) \quad \text{and} \quad \arg(\bar{z}) = -\arg(z). \tag{19}$$

- The product of a complex number $z = x + iy$ and its conjugate is known as the absolute square. It is always a non-negative real number and equals the square of the magnitude of each:

$$z \cdot \bar{z} = x^2 + y^2 = |z|^2 = |\bar{z}|^2. \tag{20}$$

- The real and imaginary parts of a complex number $z$ can be expressed using the conjugate as:

$$\Re(z) = (z + \bar{z})/2 \quad \text{and} \quad \Im(z) = (z + \bar{z})/2i. \tag{21}$$

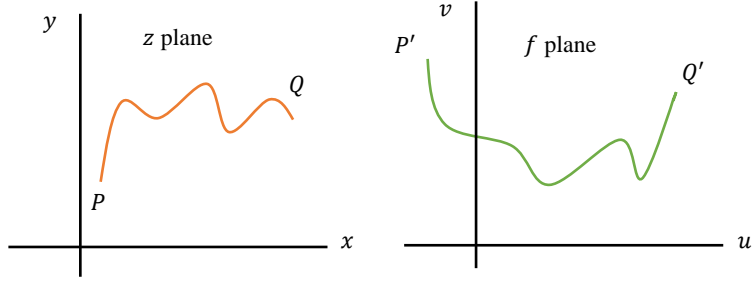

**Fig. 4.** Mapping and Transformation from $z$ Plane to $f$ Plane. Given a point $P$ in the $z$ plane with coordinates $(x, y)$, there is a corresponding point $P'$ in the $f$ plane with coordinates $(u, v)$. The set of equations defining this relationship is given by $f(z)$, which is called a transformation. Point $P$ is mapped or transformed into point $P'$ by this transformation, and $P'$ is referred to as the image of $P$. This figure visually captures the relationship between the original points and their images under a given transformation, illustrating how the transformation affects the positions of points and curves in the complex plane.

- A complex number is real if and only if it equals its own conjugate.
- Using the conjugate, the reciprocal of a nonzero complex number $z = x + iy$ can be broken into real and imaginary components

$$\frac{1}{z} = \frac{\bar{z}}{z\bar{z}} = \frac{\bar{z}}{|z|^2} = \frac{x - iy}{x^2 + y^2} = \frac{x}{x^2 + y^2} - i\frac{y}{x^2 + y^2}. \tag{22}$$

This can be used to express a division of an arbitrary complex number $w = u + iv$ by a non-zero complex number $z = x + iy$ as

$$\frac{w}{z} = \frac{w\bar{z}}{|z|^2} = \frac{(u + iv)(x - iy)}{x^2 + y^2} = \frac{ux + vy}{x^2 + y^2} + i\frac{vx - uy}{x^2 + y^2}. \tag{23}$$

A complex function is a function from complex numbers to complex numbers. In other words, it is a function that has a subset of the complex numbers as a domain and the complex numbers as a codomain. For any complex function, the values $z$ from the domain and their images $f(z)$ in the range may be separated into real and imaginary parts:

$$z = x + iy \quad \text{and} \quad f(z) = f(x + iy) = u(x, y) + iv(x, y), \tag{24}$$

where $x, y, u(x, y), v(x, y)$ are all real valued. Hence, a complex function $f: \mathbb{C} \to \mathbb{C}$ may be decomposed into

$$u: \mathbb{R}^2 \to \mathbb{R} \quad \text{and} \quad v: \mathbb{R}^2 \to \mathbb{R}, \tag{25}$$

i.e., into two real-valued functions $(u, v)$ of two real variables $(x, y)$.

If only one value of $f$ corresponds to each value of $z$, we say that $f$ is a single-valued function of $z$ or that $f(z)$ is single-valued. If more than one value of $f$ corresponds to each value of $z$, we say that $f$ is a multiple-valued or many-valued function of $z$. Whenever we speak of function, we shall, unless otherwise stated, assume a single-valued function.

For example, consider the complex function $f(z) = z^2$. If we let $z = x + iy$, then $f(z) = (x + iy)^2$. Expanding this expression yields both real and imaginary parts:

$$f(z) = x^2 - y^2 + 2ixy. \tag{26}$$

In this case,

$$u(x, y) = x^2 - y^2, \tag{27}$$

and

$$v(x, y) = 2xy, \tag{28}$$

making $f(z) = u + iv$. Thus, given a point $(x, y)$ in the $z$ plane, such as $P$ in Fig. 4, there corresponds to a point $(u, v)$ in the $f$ plane, say $P'$. The set of equations (24) [or the equivalent, $f(z)$] is called a transformation. We say that point $P$ is mapped or transformed into point $P'$ by means of the transformation and call $P'$ the image of $P$. For example, the image of a point $(1,2)$ in the $z$ plane is the point $(-3,4)$ in the $f(z) = z^2$ plane. In general, under a transformation, a set of points such as those on curve $PQ$ of Fig. 4 is mapped into a corresponding set of points, called the image, such as those on curve $P'Q'$. The particular characteristics of the image depend of course on the type of function $f(z)$, which is sometimes called a mapping function, see for example Fig. 5.

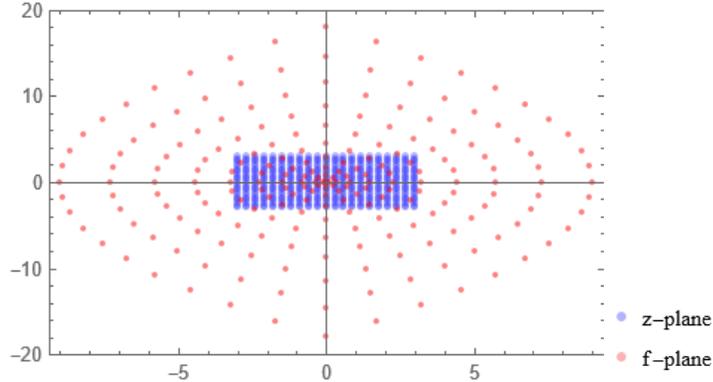

**Fig. 5.** Mapping of Points in the Complex Plane Using $f(z) = z^2$. The plot shows two sets of points: the original points in the $z$-plane and their corresponding transformed points in the $f$-plane. The points in the $z$-plane are displayed in blue with a larger point size and lower opacity, while the points in the $f$-plane are displayed in red with a smaller point size and lower opacity.

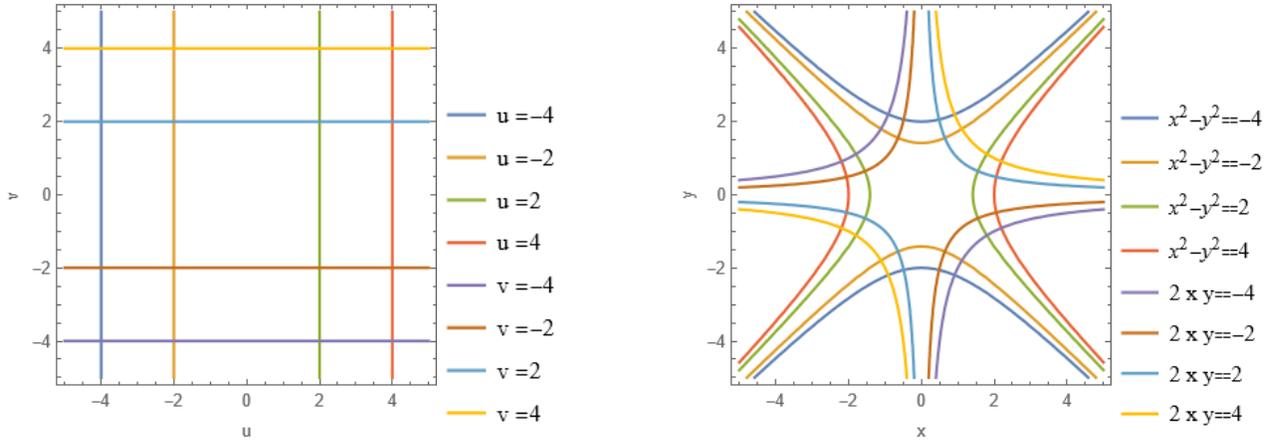

**Fig. 6.** Contour plots illustrate the real and imaginary parts of a squared complex function in the $uv$-plane, and hyperbolas defined by quadratic equations in the $xy$-plane. Left panel: The figure displays contour lines representing the real part $u$ and imaginary part $v$ of the function $f(x, y) = (x + iy)^2$. The contour lines are plotted for the values $u = -4, -2, 2, 4$ and $v = -4, -2, 2, 4$ within the range of $[-5,5]$ on both axes. Each contour line represents a constant value of $u$ or $v$, highlighting the relationship between the real and imaginary parts of the complex function in the $uv$-plane. Right panel: The figure illustrates hyperbolas defined by the equations $x^2 - y^2 = c_1$ and $2xy = c_2$ in the $xy$-plane. The contour lines represent these hyperbolas for the values $c_1 = -4, -2, 2, 4$ and $c_2 = -4, -2, 2, 4$, plotted within the range of $[-5,5]$ on both axes. Each contour line corresponds to a specific hyperbola, demonstrating the geometric shapes formed by these quadratic equations.

Now, in the above example, we will define new variables $c_1$ and $c_2$ such that $u = c_1$ and $v = c_2$. Therefore, the equations become:

$$u(x, y) = x^2 - y^2 = c_1, \qquad v(x, y) = 2xy = c_2. \tag{29}$$

These equations represent curves in the $xy$-plane, Fig. 6. The first equation $x^2 - y^2 = c_1$ represents a family of hyperbolas, and the second equation $2xy = c_2$ represents another family of curves. These curves in the $xy$-plane correspond to lines in the $uv$-plane, where $u$ is on the $c_1$ line and $v$ is on the $c_2$ line.

A graph of a real function can be drawn in two dimensions because there are two represented variables, $x$ and $y$. When visualizing complex functions, both a complex input and output are needed. Complex numbers are represented by two variables and therefore two dimensions; this means that representing a complex function (more precisely, a complex-valued function of one complex variable $f: \mathbb{C} \to \mathbb{C}$) requires the visualization of four dimensions, which is possible only in projections. Because of this, other ways of visualizing complex functions have been designed.

It is possible to add variables that keep the four-dimensional process without requiring a visualization of four dimensions. In this case, the two added variables are visual inputs such as color and brightness because they are naturally two variables easily processed and distinguished by the human eye. This assignment is called a "color function".

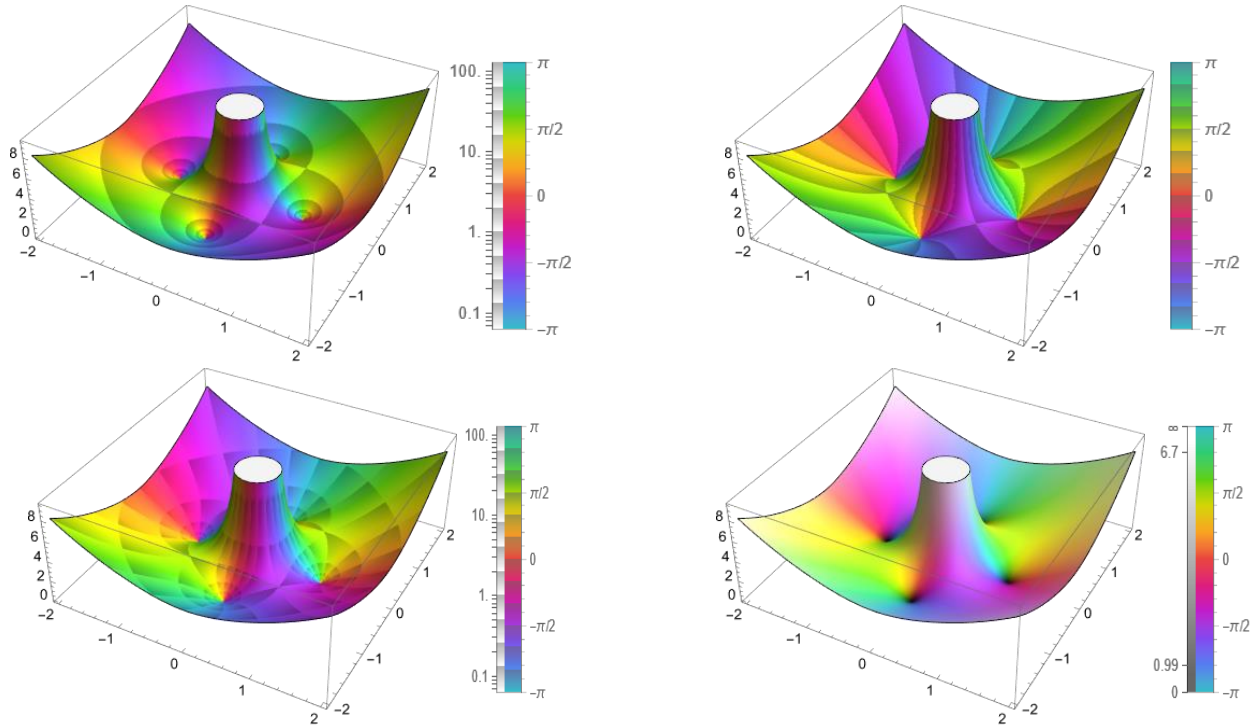

**Fig. 7.** Complex Plots of $(z^4 - 1)/z^2$ with Different Shading Functions. These figures are 3D plots of the complex function $(z^4 - 1)/z^2$ over the complex rectangle defined by the corners $-2 - 2i$ and $2 + 2i$. Each plot uses a different shading function to highlight various features of the function. The top-left panel uses CyclicLogAbs shading: the height of the surface represents the magnitude of the function, and colors are shaded cyclically based on the logarithm of the magnitude log(Abs[$f$]), creating contour-like appearances for constant magnitudes. Features visible include zeros (points where $f = 0$), poles (points where $f$ tends to infinity), and lines of constant magnitude. The top-right panel uses CyclicArg shading: the height of the surface represents the magnitude of the function, and colors are shaded cyclically based on the argument (phase) of the function Arg[$f$], highlighting regions of constant phase. Features visible include zeros, poles, and phase contours, making it easier to analyze the phase behavior of the function. The bottom-left panel uses CyclicLogAbsArg shading: the height of the surface represents the magnitude of the function, and colors are shaded cyclically based on both the logarithm of the magnitude log(Abs[$f$]) and the argument (phase) of the function Arg[$f$]. This combined shading technique highlights regions of constant magnitude and constant phase simultaneously, providing detailed contour-like appearances. The bottom-right panel uses QuantileAbs shading: the height of the surface represents the magnitude of the function, and colors are shaded from dark to light based on quantiles of Abs[$f$], with darker colors representing smaller magnitudes and lighter colors representing larger magnitudes. This shading technique emphasizes the relative magnitudes of the function values, making it easier to distinguish between low and high-magnitude regions. Each plot includes a legend that explains color mapping.

In domain coloring the output dimensions are represented by color and brightness, respectively. Each point in the complex plane as domain is ornated, typically with color representing the argument of the complex number, and brightness representing the magnitude. Dark spots mark moduli near zero, brighter spots are farther away from the origin, and the gradation may be discontinuous, but is assumed as monotonous. Domain coloring allows one to see how the function behaves across the entire complex plane, revealing patterns and structures that might not be immediately apparent in other types of plots.

Mathematica stands as a powerful tool that empowers users to explore and visualize complex numbers and functions with unprecedented ease and precision. In this survey, we generate the figures using Mathematica. In Mathematica, there are numerous functions available to represent complex functions:

- `ComplexPlot3D[f,{z,zmin,zmax}]` generates a 3D plot of Abs[f] colored by Arg[f] over the complex rectangle with corners zmin and zmax (See Fig. 7).
- `ComplexPlot[f,{z,zmin,zmax}]` generates a plot of Arg[f] over the complex rectangle with corners zmin and zmax (See Fig. 8).

Moreover, different color shading techniques can be used, such as CyclicLogAbs shading, CyclicArg shading, CyclicLogAbsArg shading, and QuantileAbs shading (see Fig. 7 and 8).

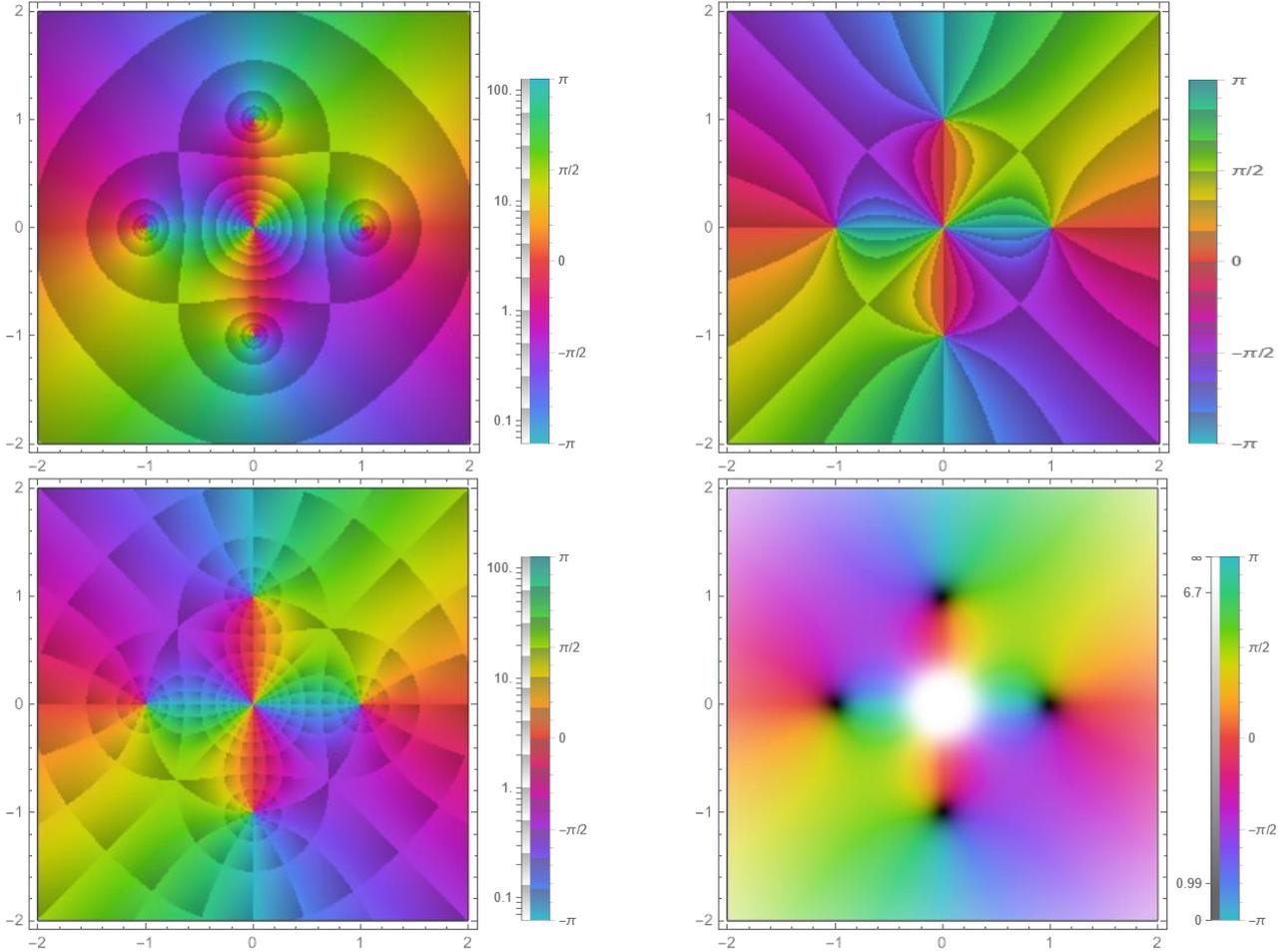

**Fig. 8.** Complex Plots of $(z^4 - 1)/z^2$ with Different Shading Functions. These figures are complex plots of the function $(z^4 - 1)/z^2$ over the complex rectangle defined by the corners $-2 - 2i$ and $2 + 2i$. Each plot uses a different shading function to visualize the argument (phase) of the function, highlighting various features such as zeros, poles, and essential singularities. The top-left panel uses CyclicLogAbs shading: the plot represents the argument of the function with colors shaded cyclically based on the logarithm of the magnitude $\log(\mathrm{Abs}[f])$. This shading creates contour-like appearances for constant magnitudes, helping to identify zeros (points where $f = 0$) and poles (points where $f$ tends to infinity). The top-right panel uses CyclicArg shading: the plot represents the argument of the function with colors shaded cyclically based on the argument itself. This shading highlights regions of constant phase, making it easier to analyze the phase behavior and locate zeros and poles. The bottom-left panel uses CyclicLogAbsArg shading: the plot combines cyclic shading based on both the logarithm of the magnitude $\log(\mathrm{Abs}[f])$ and the argument, $\mathrm{Arg}[f]$. This combined shading provides a detailed visualization of both constant magnitude and constant phase regions, enhancing the identification of zeros, poles, and essential singularities. The bottom-right panel uses QuantileAbs shading: the plot represents the argument of the function with colors shaded from dark to light based on quantiles of the magnitude $\mathrm{Abs}[f]$. Darker colors represent smaller magnitudes, and lighter colors represent larger magnitudes, emphasizing the relative magnitudes of the function values.

## 3. Complex Calculus

***Definition (Complex Differentiable):*** *Let $\mathbb{A} \subset \mathbb{C}$ be an open set. The function $f: \mathbb{A} \to \mathbb{C}$ is said to be (complex) differentiable at $z_0 \in \mathbb{A}$ if the limit*

$$f'(z_0) = \lim_{z \to z_0} \frac{f(z) - f(z_0)}{z - z_0}, \tag{30}$$

*exists independent of the manner in which $z \to z_0$. This limit is then denoted by $f'(z_0) = \left.\frac{\partial f(z)}{\partial z}\right|_{z=z_0}$ and is called the derivative of $f$ with respect to $z$ at the point $z_0$.*

***Definition (Holomorphic or Analytic):*** *Let $\mathbb{U} \subseteq \mathbb{A}$ be a nonempty open set. The function $f: \mathbb{A} \to \mathbb{C}$ is called holomorphic (or analytic) in $\mathbb{U}$, if $f$ is differentiable in $z_0$ for all $z_0 \in \mathbb{U}$.*

Basic properties for the derivative of a sum, product, and composition of two functions known from real-valued analysis remain inherently valid in the complex domain. Assume that $f(z)$ and $g(z)$ are differentiable at $z_0$. Then, the following propositions hold:

The sum $f + g$ is differentiable at $z_0$ and

$$(f + g)'(z_0) = f'(z_0) + g'(z_0). \tag{31.1}$$

The product $fg$ is differentiable at $z_0$ and

$$(fg)'(z_0) = f'(z_0)g(z_0) + f(z_0)g'(z_0). \tag{31.2}$$

If $g(z_0) \neq 0$, the quotient $f/g$ is differentiable at $z_0$ and

$$\left(\frac{f}{g}\right)'(z_0) = \frac{f'(z_0)g(z_0) - f(z_0)g'(z_0)}{g^2(z_0)}. \tag{31.3}$$

If $w = f(\xi)$ where $\xi = g(z)$ ((chain rule)) then

$$\frac{dw}{dz} = \frac{dw}{d\xi}\frac{d\xi}{dz} = f'(\xi)\frac{d\xi}{dz} = f'\big(g(z)\big)g'(z). \tag{31.4}$$

**Theorem 1:** *A necessary condition that $w = f(z) = u(x,y) + iv(x,y)$, $f: \mathbb{C} \to \mathbb{C}$, with $u, v \in \mathbb{R}$, and $z = x + iy$ with $x, y \in \mathbb{R}$ be analytic in a region $\mathbb{U}$ is that, in $\mathbb{U}$, $u$ and $v$ satisfy the Cauchy–Riemann equations* [13-18]

$$\frac{\partial u}{\partial x} = \frac{\partial v}{\partial y}, \qquad \frac{\partial u}{\partial y} = -\frac{\partial v}{\partial x}. \tag{32}$$

*If the partial derivatives in* (32) *are continuous in $\mathbb{U}$, then the Cauchy–Riemann equations are sufficient conditions that $f(z)$ be analytic in $\mathbb{U}$.*

**Proof:**

*Necessary Condition:*

In order for $f(z)$ to be analytic, the limit

$$f'(z) = \lim_{\Delta z \to 0} \frac{f(z + \Delta z) - f(z)}{\Delta z}$$
$$= \lim_{\substack{\Delta x \to 0 \\ \Delta y \to 0}} \frac{[u(x + \Delta x, y + \Delta y) + iv(x + \Delta x, y + \Delta y)] - [u(x,y) + iv(x,y)]}{\Delta x + i\Delta y},$$

must exist independent of the manner in which $\Delta z$ (or $\Delta x$ and $\Delta y$) approaches zero. We consider two possible approaches.

Case 1: $\Delta y = 0$, $\Delta x \to 0$. In this case, $\lim_{\Delta z \to 0} \frac{f(z+\Delta z)-f(z)}{\Delta z} = f'(z)$ becomes

$$\lim_{\Delta x \to 0} \frac{[u(x + \Delta x, y) + iv(x + \Delta x, y)] - [u(x,y) + iv(x,y)]}{\Delta x}$$
$$= \lim_{\Delta x \to 0} \frac{[u(x + \Delta x, y) - u(x,y)] + i[v(x + \Delta x, y) - v(x,y)]}{\Delta x}$$
$$= \lim_{\Delta x \to 0} \left(\frac{u(x + \Delta x, y) - u(x,y)}{\Delta x} + i\frac{v(x + \Delta x, y) - v(x,y)}{\Delta x}\right)$$
$$= \frac{\partial u}{\partial x} + i\frac{\partial v}{\partial x},$$

provided the partial derivatives exist.

Case 2: $\Delta x = 0$, $\Delta y \to 0$. In this case, $\lim_{\Delta z \to 0} \frac{f(z+\Delta z)-f(z)}{\Delta z} = f'(z)$ becomes

$$\lim_{\Delta y \to 0} \frac{[u(x, y + \Delta y) + iv(x, y + \Delta y)] - [u(x,y) + iv(x,y)]}{i\Delta y}$$
$$= \lim_{\Delta y \to 0} \frac{[u(x, y + \Delta y) - u(x,y)] + i[v(x, y + \Delta y) - v(x,y)]}{i\Delta y}$$

$$\begin{aligned}
&= \lim_{\Delta y \to 0} \left( \frac{u(x, y + \Delta y) - u(x, y)}{i\Delta y} + \frac{i[v(x, y + \Delta y) - v(x, y)]}{i\Delta y} \right) \\
&= \lim_{\Delta y \to 0} \left( -i\frac{u(x, y + \Delta y) - u(x, y)}{\Delta y} + \frac{[v(x, y + \Delta y) - v(x, y)]}{\Delta y} \right) \\
&= -i\frac{\partial u}{\partial y} + \frac{\partial v}{\partial y}.
\end{aligned}$$

Now $f(z)$ cannot possibly be analytic unless these two limits are identical. Thus, a necessary condition that $f(z)$ be analytic is

$$\frac{\partial u}{\partial x} + i\frac{\partial v}{\partial x} = -i\frac{\partial u}{\partial y} + \frac{\partial v}{\partial y},$$

or

$$\frac{\partial u}{\partial x} = \frac{\partial v}{\partial y}, \qquad \frac{\partial v}{\partial x} = -\frac{\partial u}{\partial y}.$$

*Sufficient Condition:*

Since $\frac{\partial u}{\partial x}$ and $\frac{\partial u}{\partial y}$ are supposed to be continuous, we have

$$\begin{aligned}
\Delta u &= u(x + \Delta x, y + \Delta y) - u(x, y) \\
&= \{u(x + \Delta x, y + \Delta y) - u(x, y + \Delta y)\} + \{u(x, y + \Delta y) - u(x, y)\} \\
&= \left\{\frac{\partial u}{\partial x} + \Delta \epsilon_1\right\} \Delta x + \left\{\frac{\partial u}{\partial y} + \Delta \eta_1\right\} \Delta y \\
&= \frac{\partial u}{\partial x} \Delta x + \Delta \epsilon_1 \Delta x + \frac{\partial u}{\partial y} \Delta y + \Delta \eta_1 \Delta y \\
&= \frac{\partial u}{\partial x} \Delta x + \frac{\partial u}{\partial y} \Delta y + \Delta \epsilon_1 \Delta x + \Delta \eta_1 \Delta y,
\end{aligned}$$

where $\Delta \epsilon_1 \to 0$ and $\Delta \eta_1 \to 0$ as $\Delta x \to 0$ and $\Delta y \to 0$. Similarly, since $\frac{\partial v}{\partial x}$ and $\frac{\partial v}{\partial y}$ are supposed to be continuous, we have

$$\begin{aligned}
\Delta v &= v(x + \Delta x, y + \Delta y) - v(x, y) \\
&= \{v(x + \Delta x, y + \Delta y) - v(x, y + \Delta y)\} + \{v(x, y + \Delta y) - v(x, y)\} \\
&= \left\{\frac{\partial v}{\partial x} + \Delta \epsilon_2\right\} \Delta x + \left\{\frac{\partial v}{\partial y} + \Delta \eta_2\right\} \Delta y \\
&= \frac{\partial v}{\partial x} \Delta x + \Delta \epsilon_2 \Delta x + \frac{\partial v}{\partial y} \Delta y + \Delta \eta_2 \Delta y \\
&= \frac{\partial v}{\partial x} \Delta x + \frac{\partial v}{\partial y} \Delta y + \Delta \epsilon_2 \Delta x + \Delta \eta_2 \Delta y,
\end{aligned}$$

where $\Delta \epsilon_2 \to 0$ and $\Delta \eta_2 \to 0$ as $\Delta x \to 0$ and $\Delta y \to 0$.

Then

$$\begin{aligned}
\Delta w &= \Delta u + i\Delta v \\
&= \frac{\partial u}{\partial x} \Delta x + \frac{\partial u}{\partial y} \Delta y + \Delta \epsilon_1 \Delta x + \Delta \eta_1 \Delta y + i\left(\frac{\partial v}{\partial x} \Delta x + \frac{\partial v}{\partial y} \Delta y + \Delta \epsilon_2 \Delta x + \Delta \eta_2 \Delta y\right) \\
&= \frac{\partial u}{\partial x} \Delta x + i\frac{\partial v}{\partial x} \Delta x + \frac{\partial u}{\partial y} \Delta y + i\frac{\partial v}{\partial y} \Delta y + \Delta \epsilon_1 \Delta x + \Delta \eta_1 \Delta y + i\Delta \epsilon_2 \Delta x + i\Delta \eta_2 \Delta y \\
&= \left(\frac{\partial u}{\partial x} + i\frac{\partial v}{\partial x}\right) \Delta x + \left(\frac{\partial u}{\partial y} + i\frac{\partial v}{\partial y}\right) \Delta y + (\Delta \epsilon_1 + i\Delta \epsilon_2) \Delta x + (\Delta \eta_1 + i\Delta \eta_2) \Delta y \\
&= \left(\frac{\partial u}{\partial x} + i\frac{\partial v}{\partial x}\right) \Delta x + \left(\frac{\partial u}{\partial y} + i\frac{\partial v}{\partial y}\right) \Delta y + \Delta \epsilon \Delta x + \Delta \eta \Delta y,
\end{aligned}$$

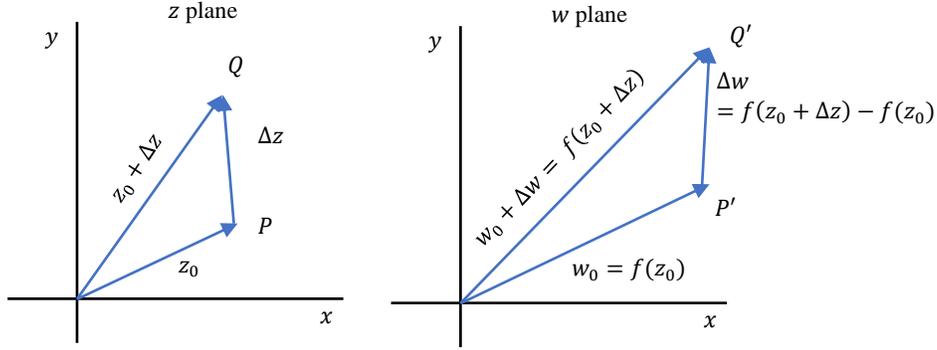

**Fig. 9.** Geometric Interpretation of the Complex Derivative. In the $z$-plane, point $P$ corresponds to $z_0$, and point $Q$ corresponds to $z_0 + \Delta z$. The change in the complex number is denoted by $\Delta z$. In the $w$-plane, point $P'$ corresponds to $w_0 = f(z_0)$, and point $Q'$ corresponds to $f(z_0 + \Delta z)$. The change in the function value is denoted by $\Delta w = f(z_0 + \Delta z) - f(z_0)$. The interpretation showcases how a small perturbation $\Delta z$ in the $z$-plane maps to a perturbation $\Delta w$ in the $w$-plane, illustrating the derivative's role in describing this transformation.

or

$$\Delta w = \left(\frac{\partial u}{\partial x} + i\frac{\partial v}{\partial x}\right)\Delta x + \left(\frac{\partial u}{\partial y} + i\frac{\partial v}{\partial y}\right)\Delta y + \Delta \epsilon \Delta x + \Delta \eta \Delta y,$$

where $\Delta \epsilon = \Delta \epsilon_1 + i\Delta \epsilon_2 \to 0$ and $\Delta \eta = \Delta \eta_1 + i\Delta \eta_2 \to 0$ as $\Delta x \to 0$ and $\Delta x \to 0$.

By the Cauchy–Riemann equations, $\Delta w$ can be written

$$\begin{aligned}\Delta w &= \left(\frac{\partial u}{\partial x} + i\frac{\partial v}{\partial x}\right)\Delta x + \left(-\frac{\partial v}{\partial x} + i\frac{\partial u}{\partial x}\right)\Delta y + \Delta \epsilon \Delta x + \Delta \eta \Delta y \\ &= \frac{\partial u}{\partial x}\Delta x + i\frac{\partial u}{\partial x}\Delta y + i\frac{\partial v}{\partial x}\Delta x - \frac{\partial v}{\partial x}\Delta y + \Delta \epsilon \Delta x + \Delta \eta \Delta y \\ &= \frac{\partial u}{\partial x}(\Delta x + i\Delta y) + i\frac{\partial v}{\partial x}(\Delta x + i\Delta y) + \Delta \epsilon \Delta x + \Delta \eta \Delta y \\ &= \left(\frac{\partial u}{\partial x} + i\frac{\partial v}{\partial x}\right)(\Delta x + i\Delta y) + \Delta \epsilon \Delta x + \Delta \eta \Delta y.\end{aligned}$$

Then, on dividing by $\Delta z = \Delta x + i\Delta y$ and taking the limit as $\Delta z \to 0$, we see that

$$\frac{dw}{dz} = f'(z) = \lim_{\Delta z \to 0} \frac{\Delta w}{\Delta z} = \frac{\partial u}{\partial x} + i\frac{\partial v}{\partial x},$$

so that the derivative exists and is unique, i.e., $f(z)$ is analytic in $\mathbb{U}$. ∎

The geometric interpretation of the complex derivative is analogous to the interpretation of the derivative for real functions but with some differences due to the fact that complex functions map from the complex plane to itself. Let $z_0$, Fig. 9, be a point $P$ in the $z$ plane and let $w_0$ be its image $P'$ in the $w$ plane under the transformation $w = f(z)$. Since we suppose that $f(z)$ is single valued, the point $z_0$ maps into only one point $w_0$. If we give $z_0$ an increment $\Delta z$, we obtain the point $Q$ of Fig. 9. This point has image $Q'$ in the $w$ plane. Thus, from Fig. 9, we see that $P'Q'$ represents the complex number $\Delta w = f(z_0 + \Delta z) - f(z_0)$. It follows that the derivative at $z_0$ (if it exists) is given by

$$\lim_{\Delta z \to 0} \frac{\Delta w}{\Delta z} = \lim_{\Delta z \to 0} \frac{f(z_0 + \Delta z) - f(z_0)}{\Delta z} = \lim_{\Delta z \to 0} \frac{P'Q'}{PQ}, \tag{33}$$

that is, the limit of the ratio $P'Q'$ to $PQ$ as point $Q$ approaches point $P$. This interpretation clearly holds when $z_0$ is replaced by any point $z$.

Let $\Delta z = dz$ be an increment given to $z$. Then $\Delta w = f(z + \Delta z) - f(z)$ is called the increment in $w = f(z)$. If $f(z)$ is continuous and has a continuous first derivative in a region, then

$$\Delta w = f'(z)\Delta z + \epsilon \Delta z = f'(z)dz + \epsilon dz, \tag{34}$$

where $\epsilon \to 0$ as $\Delta z \to 0$. The expression

$$dw = f'(z)dz, \qquad (35)$$

is called the differential of $w$ or $f(z)$, or the principal part of $\Delta w$.

**Theorem 2:** *The differential $df$ of a complex-valued function $f(z): \mathbb{A} \to \mathbb{C}$ with $\mathbb{A} \subset \mathbb{C}$ can be expressed as*

$$df = \frac{\partial f(z)}{\partial z}dz + \frac{\partial f(z)}{\partial z^*}dz^*. \qquad (36)$$

**Proof:** The total differential of the bivariate function $F(x,y): \mathbb{R}^2 \to \mathbb{C}$ and $u, v: \mathbb{R}^2 \to \mathbb{R}$ associated with the univariate function $f(z)$ via

$$z = x + iy, \qquad F(x,y) = u(x,y) + iv(x,y) = f(z) = f(z = x + iy),$$

reads as

$$dF = \frac{\partial F(x,y)}{\partial x}dx + \frac{\partial F(x,y)}{\partial y}dy.$$

Of course, the differentiability of $F(x,y)$ with respect to $x$ and $y$ in the real sense has to be imposed for the existence of the differential $dF$. This implies the differentiability of the real-valued functions $u(x,y)$ and $v(x,y)$ with respect to $x$ and $y$. Rewriting $dF$ by means of $F(x,y) = u(x,y) + iv(x,y)$ yields

$$\begin{aligned}
dF &= \frac{\partial F(x,y)}{\partial x}dx + \frac{\partial F(x,y)}{\partial y}dy \\
&= \frac{\partial}{\partial x}[u(x,y) + iv(x,y)]dx + \frac{\partial}{\partial y}[u(x,y) + iv(x,y)]dy \\
&= \frac{\partial u(x,y)}{\partial x}dx + i\frac{\partial v(x,y)}{\partial x}dx + \frac{\partial u(x,y)}{\partial y}dy + i\frac{\partial v(x,y)}{\partial y}dy.
\end{aligned}$$

Making use of

$$dz = dx + idy, \qquad dz^* = dx - idy,$$

the two differentials $dx$ and $dy$ can be expressed via

$$dx = \frac{1}{2}(dz + dz^*),$$

$$dy = \frac{1}{2i}(dz - dz^*).$$

Inserting $dx$ and $dy$ into the differential expression $dF$ and reordering the result leads to

$$\begin{aligned}
dF &= \frac{\partial u(x,y)}{\partial x}dx + i\frac{\partial v(x,y)}{\partial x}dx + \frac{\partial u(x,y)}{\partial y}dy + i\frac{\partial v(x,y)}{\partial y}dy \\
&= \frac{\partial u(x,y)}{\partial x}\frac{1}{2}(dz + dz^*) + i\frac{\partial v(x,y)}{\partial x}\frac{1}{2}(dz + dz^*) \\
&\quad + \frac{\partial u(x,y)}{\partial y}\frac{1}{2i}(dz - dz^*) + i\frac{\partial v(x,y)}{\partial y}\frac{1}{2i}(dz - dz^*) \\
&= \frac{1}{2}\left(\frac{\partial u(x,y)}{\partial x} + i\frac{\partial v(x,y)}{\partial x} - i\frac{\partial u(x,y)}{\partial y} + \frac{\partial v(x,y)}{\partial y}\right)dz \\
&\quad + \frac{1}{2}\left(\frac{\partial u(x,y)}{\partial x} + i\frac{\partial v(x,y)}{\partial x} + i\frac{\partial u(x,y)}{\partial y} - \frac{\partial v(x,y)}{\partial y}\right)dz^* \\
&= \frac{1}{2}\left(\frac{\partial}{\partial x}[u(x,y) + iv(x,y)] - i\frac{\partial}{\partial y}[u(x,y) + iv(x,y)]\right)dz \\
&\quad + \frac{1}{2}\left(\frac{\partial}{\partial x}[u(x,y) + iv(x,y)] + i\frac{\partial}{\partial y}[u(x,y) + iv(x,y)]\right)dz^*
\end{aligned}$$

$$= \frac{1}{2}\left(\frac{\partial}{\partial x}F(x,y) - i\frac{\partial}{\partial y}F(x,y)\right)dz + \frac{1}{2}\left(\frac{\partial}{\partial x}F(x,y) + i\frac{\partial}{\partial y}F(x,y)\right)dz^*$$

$$= \frac{1}{2}\left(\frac{\partial}{\partial x} - i\frac{\partial}{\partial y}\right)F(x,y)dz + \frac{1}{2}\left(\frac{\partial}{\partial x} + i\frac{\partial}{\partial y}\right)F(x,y)dz^*.$$

Finally,

$$dF = \frac{1}{2}\left(\frac{\partial}{\partial x} - i\frac{\partial}{\partial y}\right)F(x,y)dz + \frac{1}{2}\left(\frac{\partial}{\partial x} + i\frac{\partial}{\partial y}\right)F(x,y)dz^*.$$

According to the total differential for real-valued multivariate functions, the introduction of the two operators

$$\frac{\partial}{\partial z} = \frac{1}{2}\left(\frac{\partial}{\partial x} - i\frac{\partial}{\partial y}\right), \qquad \frac{\partial}{\partial z^*} = \frac{1}{2}\left(\frac{\partial}{\partial x} + i\frac{\partial}{\partial y}\right),$$

is reasonable as it leads to the very nice description of the differential $df$, where the real-valued partial derivatives are hidden.

$$df = \frac{\partial f(z)}{\partial z}dz + \frac{\partial f(z)}{\partial z^*}dz^*.$$

■

Although many important complex functions are holomorphic, including the functions $z^n$, $e^z$, $\ln(z)$, $\sin(z)$, and $\cos(z)$, and hence differentiable in the standard complex variables sense, there are commonly encountered useful functions which are not. For example, the function $f(z) = z^*$, fails to satisfy the Cauchy-Riemann conditions. Moreover, the functions $f(z) = \text{Re}(z) = \frac{z+z^*}{2} = x$ and $g(z) = \text{Im}(z) = i\frac{z-z^*}{2} = y$ fail to satisfy the Cauchy-Riemann conditions.

Note that the nonholomorphic (nonanalytic in the complex variable $z$) functions, given in the above examples, can all be written in the form $f(z, z^*)$, where they are holomorphic in $z = x + iy$ for fixed $z^*$ and holomorphic in $z^* = x - iy$ for fixed $z$. That is, if we make the substitution $w = z^*$, they are analytic in $w$ for fixed $z$, and analytic in $z$ for fixed $w$. It can be shown that this fact is true in general for any complex-valued function

$$f(z) = f(z, z^*) = f(x, y) = u(x, y) + iv(x, y). \tag{37}$$

Non-holomorphic functions, however, can be dealt with conjugate coordinates, which are related to the real coordinates by

$$z = x + iy, \qquad z^* = x - iy. \tag{38}$$

Then, one can write the two real variables as,

$$x = \frac{z + z^*}{2}, \qquad y = i\frac{z^* - z}{2}. \tag{39}$$

**Definition (Wirtinger Derivatives):** The two 'partial derivative' operators $\frac{\partial}{\partial z}$ and $\frac{\partial}{\partial z^*}$ are defined by [16-18]

$$\frac{\partial}{\partial z} = \frac{1}{2}\left(\frac{\partial}{\partial x} - i\frac{\partial}{\partial y}\right), \tag{40.1}$$

$$\frac{\partial}{\partial z^*} = \frac{1}{2}\left(\frac{\partial}{\partial x} + i\frac{\partial}{\partial y}\right), \tag{40.2}$$

and are often referred to as the Wirtinger derivatives (sometimes also called Wirtinger operators). The derivatives, (40.1) and (40.2) are called $\mathbb{R}$-derivative and conjugate $\mathbb{R}$-derivative, respectively.

Consequently,

$$\frac{\partial}{\partial x} = \frac{\partial}{\partial z} + \frac{\partial}{\partial z^*}, \qquad \frac{\partial}{\partial y} = i\left(\frac{\partial}{\partial z} - \frac{\partial}{\partial z^*}\right). \tag{41}$$

In the evaluation of $\frac{\partial f}{\partial z}$, $z^*$ is considered as a constant and vice versa. For example, with $f(z, z^*) = z^2 z^*$, we have $\frac{\partial f}{\partial z} = 2zz^*$, $\frac{\partial f}{\partial z^*} = z^2$.

Wirtinger calculus (Wirtinger operators) extends standard calculus rules to the complex domain. The attractiveness of Wirtinger calculus is that it enables us to perform all computations directly in the complex domain, and the derivatives obey all rules of conventional calculus, including the chain rule, differentiation of products, and quotients. If $\alpha$ and $\beta$ are complex numbers, we have

$$\frac{\partial}{\partial z}(\alpha f + \beta g) = \alpha \frac{\partial f}{\partial z} + \beta \frac{\partial g}{\partial z}, \tag{42.1}$$

$$\frac{\partial}{\partial z^*}(\alpha f + \beta g) = \alpha \frac{\partial f}{\partial z^*} + \beta \frac{\partial g}{\partial z^*}, \tag{42.2}$$

$$\frac{\partial}{\partial z}(f \cdot g) = \frac{\partial f}{\partial z} \cdot g + f \cdot \frac{\partial g}{\partial z}, \tag{42.3}$$

$$\frac{\partial}{\partial z^*}(f \cdot g) = \frac{\partial f}{\partial z^*} \cdot g + f \cdot \frac{\partial g}{\partial z^*}, \tag{42.4}$$

$$\left(\frac{\partial f}{\partial z}\right)^* = \frac{\partial f^*}{\partial z^*}, \tag{42.5}$$

$$\left(\frac{\partial f}{\partial z^*}\right)^* = \frac{\partial f^*}{\partial z}. \tag{42.6}$$

An elegant approach due to Wirtinger relaxes the strong requirement for differentiability (Cauchy-Riemann equations), and defines a less stringent form for the complex domain. Wirtinger derivatives allow you to treat $z$ and $z^*$ as independent variables. This is particularly useful when dealing with functions that depend on both a complex variable and its conjugate. This independence is reflected in the fact that

$$\frac{\partial z}{\partial z} = \frac{\partial z^*}{\partial z^*} = 1, \quad \frac{\partial z}{\partial z^*} = \frac{\partial z^*}{\partial z} = 0. \tag{42.7}$$

**Theorem 3:** Given $f: \mathbb{C} \to \mathbb{C}$ holomorphic with $f(z) = f(z = x + iy) = u(x,y) + iv(x,y)$ where $u, v : \mathbb{R}^2 \to \mathbb{R}$ real differentiable functions. Then

$$\frac{\partial f}{\partial z^*} = 0. \tag{43}$$

It is easy to see that the Cauchy-Riemann equations are equivalent to $\frac{\partial f}{\partial z^*} = 0$.

**Proof:** Using Wirtinger calculus

$$\frac{\partial f}{\partial z^*} = \frac{1}{2}\left(\frac{\partial}{\partial x} + i\frac{\partial}{\partial y}\right)f.$$

By definition, then:

$$\begin{aligned}
\frac{\partial f}{\partial z^*} &= \frac{1}{2}\left(\frac{\partial f}{\partial x} + i\frac{\partial f}{\partial y}\right) \\
&= \frac{1}{2}\left(\frac{\partial}{\partial x}[u(x,y) + iv(x,y)] + i\frac{\partial}{\partial y}[u(x,y) + iv(x,y)]\right) \\
&= \frac{1}{2}\left(\frac{\partial}{\partial x}u(x,y) + i\frac{\partial}{\partial x}v(x,y) + i\frac{\partial}{\partial y}u(x,y) - \frac{\partial}{\partial y}v(x,y)\right) \\
&= \frac{1}{2}\left(\left[\frac{\partial}{\partial x}u(x,y) - \frac{\partial}{\partial y}v(x,y)\right] + i\left[\frac{\partial}{\partial x}v(x,y) + \frac{\partial}{\partial y}u(x,y)\right]\right) \\
&= \frac{1}{2}\left(\left[\frac{\partial}{\partial x}u(x,y) - \frac{\partial}{\partial x}u(x,y)\right] + i\left[\frac{\partial}{\partial x}v(x,y) - \frac{\partial}{\partial x}v(x,y)\right]\right) \\
&= 0.
\end{aligned}$$

Because $f$ is holomorphic then the Cauchy-Riemann equations $\frac{\partial u}{\partial x} = \frac{\partial v}{\partial y}, \frac{\partial u}{\partial y} = -\frac{\partial v}{\partial x}$ applies, making $\frac{\partial f}{\partial z^*}$ equal to zero. ∎

The condition $\frac{\partial f}{\partial z^*} = 0$ is true for an $\mathbb{R}$-differentiable function $f$ if and only the Cauchy-Riemann conditions are satisfied. Thus, a function $f$ is holomorphic (complex-analytic in $z$) if and only if it does not depend on the complex conjugate variable $z^*$. This obviously provides a simple and powerful characterization of holomorphic and nonholomorphic functions and shows the elegance of the Wirtinger calculus formulation based on the use of conjugate coordinates $(z, z^*)$. Note that the two Cauchy-Riemann conditions are replaced by the single condition $\frac{\partial f}{\partial z^*} = 0$.

Optimizations in machine learning [10] are targeted on the minimization of a cost. The cost functions are real-valued. On account of this, we focus on functions $f(z): \mathbb{U} \to \mathbb{R}$ having complex-valued arguments $z \in \mathbb{U} \subset \mathbb{C}$ that are mapped to real-valued scalars $f(z) \in \mathbb{R}$.

First of all, it is obvious that the only possibility of a real-valued function $f(z)$ with complex argument $z$ for being analytic is that $f(z)$ is constant for all $z$ of its domain. This follows from the Cauchy-Riemann equations, since $v(x, y) = 0$ for real-valued $f(z)$. This leads to the following proposition.

***Lemma 1:*** *All non-trivial (not constant) real-valued functions $f(z)$ mapping $z \in \mathbb{A} \subset \mathbb{C}$ onto $\mathbb{R}$ are non-analytic functions and therefore not complex differentiable.*

***Theorem 4:*** *Let $\mathbb{A} \subset \mathbb{C}$, and $f: \mathbb{A} \to \mathbb{R}$ be a real-valued function. The total differential of $f$ is given by*

$$\mathrm{d}f = 2\Re\left(\frac{\partial f(z)}{\partial z} dz\right) = 2\Re\left(\frac{\partial f(z)}{\partial z^*} dz^*\right). \tag{44}$$

**Proof:** From the definition of the Wirtinger differentials and partial derivatives we obtain

$$\frac{\partial f(z)}{\partial z} dz = \frac{1}{2}\left(\frac{\partial f}{\partial x} - i\frac{\partial f}{\partial y}\right)(\mathrm{d}x + i\mathrm{d}y)$$
$$= \frac{1}{2}\left(\frac{\partial f}{\partial x} \mathrm{d}x + \frac{\partial f}{\partial y} \mathrm{d}y\right) + \frac{i}{2}\left(\frac{\partial f}{\partial x} \mathrm{d}y - \frac{\partial f}{\partial y} \mathrm{d}x\right).$$

Hence,

$$2\Re\left(\frac{\partial f(z)}{\partial z} dz\right) = \frac{\partial f}{\partial x} \mathrm{d}x + \frac{\partial f}{\partial y} \mathrm{d}y = \mathrm{d}f.$$

The analog statement holds for the conjugates

$$\frac{\partial f(z)}{\partial z^*} dz^* = \frac{1}{2}\left(\frac{\partial f}{\partial x} + i\frac{\partial f}{\partial y}\right)(\mathrm{d}x - i\mathrm{d}y) = \frac{1}{2}\left(\frac{\partial f}{\partial x} \mathrm{d}x + \frac{\partial f}{\partial y} \mathrm{d}y\right) + \frac{i}{2}\left(\frac{\partial f}{\partial y} \mathrm{d}x - \frac{\partial f}{\partial x} \mathrm{d}y\right).$$

Hence, $2\Re\left(\frac{\partial f(z)}{\partial z^*} dz^*\right) = \frac{\partial f}{\partial x} \mathrm{d}x + \frac{\partial f}{\partial y} \mathrm{d}y = \mathrm{d}f.$ ∎

## 4. Classification of CVNNs

CVNNs have emerged as a powerful tool in the realm of NN architectures, introducing a unique approach to processing complex-valued inputs. There are two types of CVNNs [31-34]: fully complex CVNNs and split-CVNNs.

### *4.1 Fully-CVNNs*

To deal with complex-valued signals, fully-CVNNs have been proposed by [35] where both of the weights and AFs are in the complex domain. Fully CVNNs preserve the information carried by phase components. However, fully CVNNs increase the computation complexity and face difficulty in obtaining complex-differentiable (or holomorphic) AF [20].

### *4.2 Split-CVNNs*

Split-CVNNs are commonly used to avoid the singularity problems in complex-valued functions and their derivatives. The fundamental idea behind split-CVNNs involves the division of complex-valued inputs, comprising real and imaginary parts, into distinct real-valued components. This split, occurring in either rectangular (real and imaginary) or polar coordinates (phase and magnitude), as shown in (45.1) and (45.2), respectively, forms the basis for subsequent processing in RVNNs.

$$\{z_1, z_2\} = \{(x_1, y_1), (x_2, y_2)\}, \tag{45.1}$$
$$\{z_1, z_2\} = \{(r_1, \varphi_1), (r_2, \varphi_2)\}, \tag{45.2}$$

where $z_1, z_2 \in \mathbb{C}$, $z = x + iy$, in which $x$ and $y$ are the real and imaginary components of the complex-valued input, respectively, while $r$ and $\varphi$ are the magnitude and phase components of the complex-valued input, respectively. However, split-CVNNs are categorized into two distinct types: those with both real-valued weights and a real-valued AF, and those with complex-valued weights and a real-valued AF.

*4.2.1 Split-CVNNs: Real-valued weights and real-valued AFs*

The complex inputs (containing both real and imaginary parts) and targets (desired outputs) are separated into two sets of real-valued data. This separation leads to an increase in dimensionality. For instance, a problem with two complex-valued inputs and one complex-valued output becomes a problem with four real-valued inputs and two real-valued outputs (doubling the number of data points). This approach allows us to utilize existing RVNN architectures like the multi-layer perceptron with backpropagation. However, it's important to consider:

- Computational cost: This split-CVNN, while conceptually straightforward, presents challenges such as increased input dimension, learnable parameters, and network size. The increased dimensionality can lead to higher computational demands.
- Potential loss of information: Splitting complex data might discard some of the inherent relationships encoded within the complex numbers. It introduces phase distortion and accuracy issues during network updates, as real-valued gradients may not accurately represent the true complex gradient.

*4.2.2 Split-CVNNs: Complex-valued weights and real-valued AFs*

In contrast, CVNNs with complex-valued weights and a real-valued AF offer a solution to the phase distortion challenge. This approach, exemplified in [22], has found widespread application in various domains due to its compatibility with popular real-valued AFs, including ReLU, Sigmoid, and Tanh. The ease of computation and avoidance of phase distortion make this variant of CVNNs particularly attractive, as evidenced in various applications [35-39]. Fig. 10 shows the structure of a single neuron in the split-CVNN with complex-valued weights and a real-valued AF. It shows that the output of a neuron in the split-CVNN is equal to the output of the AF with the complex weighting summation of all its inputs being argument. The real-valued AF handle the real and imaginary parts separately, $\sigma_{\text{Real–imaginary}}(z) = \sigma^{\Re}(\Re[z]) + i\sigma^{\Re}(\Im[z])$, these functions preserve essential complex information. In Fig. 11, a complex-valued neuron with $n$-inputs is equivalent to two real-valued neurons with $2n$-inputs. We shall refer to a real-valued neuron corresponding to the real part $X$ of an output of a complex-valued neuron as a Real-part Neuron, and a real-valued neuron corresponding to the imaginary part $Y$ as an Imaginary-part Neuron.

Before concluding this section, let us explain the primary reason for advocating CVNNs. The main rationale for promoting CVNNs stems from the unique properties of complex number arithmetic, particularly the multiplication operation. In NNs, input signals are subjected to summation processes where connection weights are multiplied and added up. To unveil potential distinctions between real and complex representations, let us examine the summation of four real numbers ($x_1$, $x_2$, $y_1$, and $y_2$), i.e., sum on $x$ ($x_1 + x_2$), sum on $y$ ($y_1 + y_2$) and compare it with the summation of two complex numbers ($z_1 = x_1 + iy_1$ and $z_2 = x_2 + iy_2$), i.e., $(x_1 + x_2) + i(y_1 + y_2)$. Upon evaluating these summations, we observe that the basis of summation remains unchanged between the real and complex domains. The real and imaginary components ($x_n$ and $y_n$) are summed separately in both cases, highlighting a commonality in this fundamental operation. The crucial divergence emerges when we explore the weighting processes, particularly the multiplication of signals by connection weights. In the context of real numbers, the product is straightforward, involving the multiplication of individual real values (multiplication on $x$ ($x_1 x_2$) and multiplication on $y$ ($y_1 y_2$)). However, in the complex domain, the multiplication of two complex numbers ($z_1 z_2$) yields a more intricate result:

$$z_1 z_2 = (x_1 + iy_1)(x_2 + iy_2) = (x_1 x_2 - y_1 y_2) + i(x_1 y_2 + x_2 y_1). \tag{46}$$

This complex product involves a combination of real and imaginary components, deviating significantly from the simplicity of real-number multiplication [30].

To comprehend the intricacies of complex multiplication, we can reinterpret the process in terms of amplitude and phase. If we represent complex numbers as $z_1 = r_1 e^{i\theta_1}$ and $z_2 = r_2 e^{i\theta_2}$, the complex multiplication $z_1 z_2$ can be expressed as:

$$z_1 z_2 = r_1 r_2 e^{i(\theta_1 + \theta_2)}. \tag{47}$$

This representation demonstrates that complex multiplication is equivalent to combining amplitude multiplication and phase addition. This fact suggests that CVNNs can be more meaningful in applications where they should adopt the amplitude-phase-type AF.

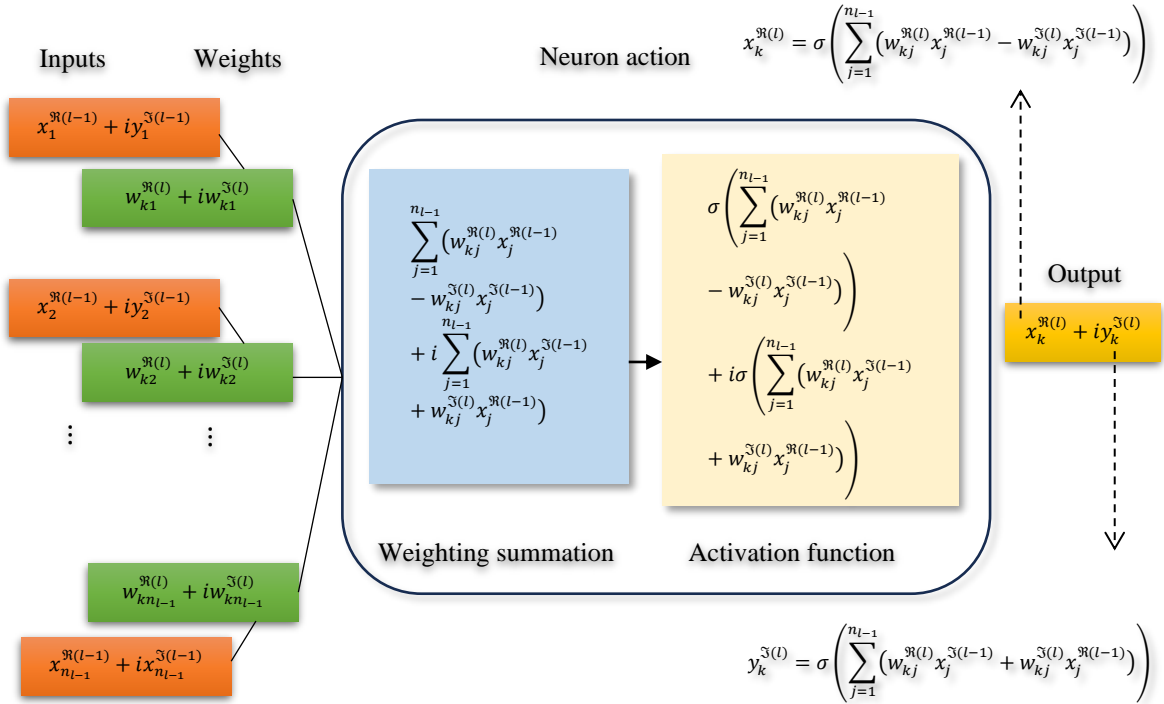

**Fig. 10.** Structure of a neuron in a split CVNN.

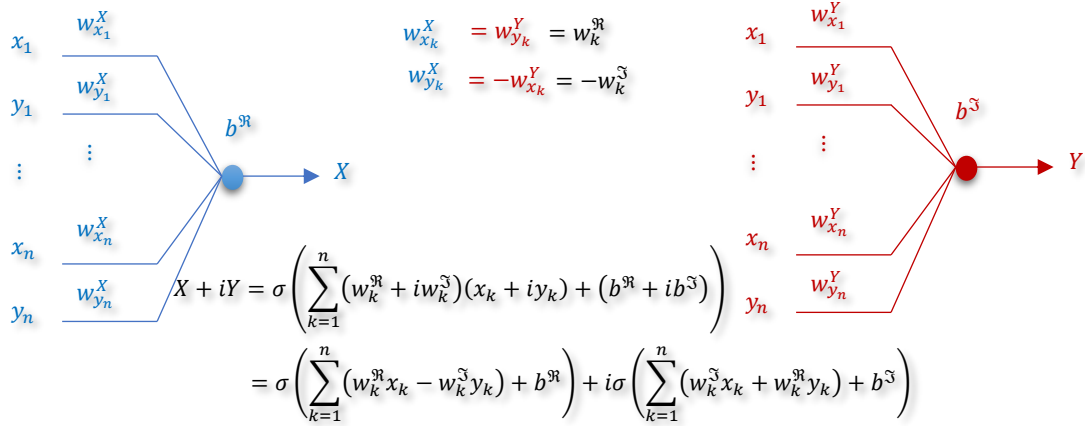

**Fig. 11.** Two real-valued neurons which are equivalent to a complex-valued neuron.

## 5. Backpropagation Algorithms for CVNN

In the realm of CVNNs, AFs are classified into two main categories: Split AFs and Fully Complex AFs. Split AFs handle complex-valued inputs by applying a real-valued AF (Split Real-Imaginary AF) to the real and imaginary parts separately, and the Split Phase-Amplitude AF utilizing the magnitude and phase of the complex number. For more details see section 6. These AFs, while not analytic, offer design flexibility by focusing on meaningful partial derivatives. In contrast, fully complex AFs treat the real and imaginary parts of a complex number as a single entity, preserving the inherent relationships between them. For example, the complex Sigmoid function, despite its computational challenges, promises more integrated network behaviors. Training CVNNs necessitates the use of CBP. This involves adapting traditional backpropagation methods to accommodate complex-valued functions. Various techniques, including the Complex Derivative Approach, the Partial Derivatives Approach, and algorithms that incorporate the Cauchy-Riemann equations, have been developed for this purpose. This section provides an in-depth exploration of these three CBP algorithms.

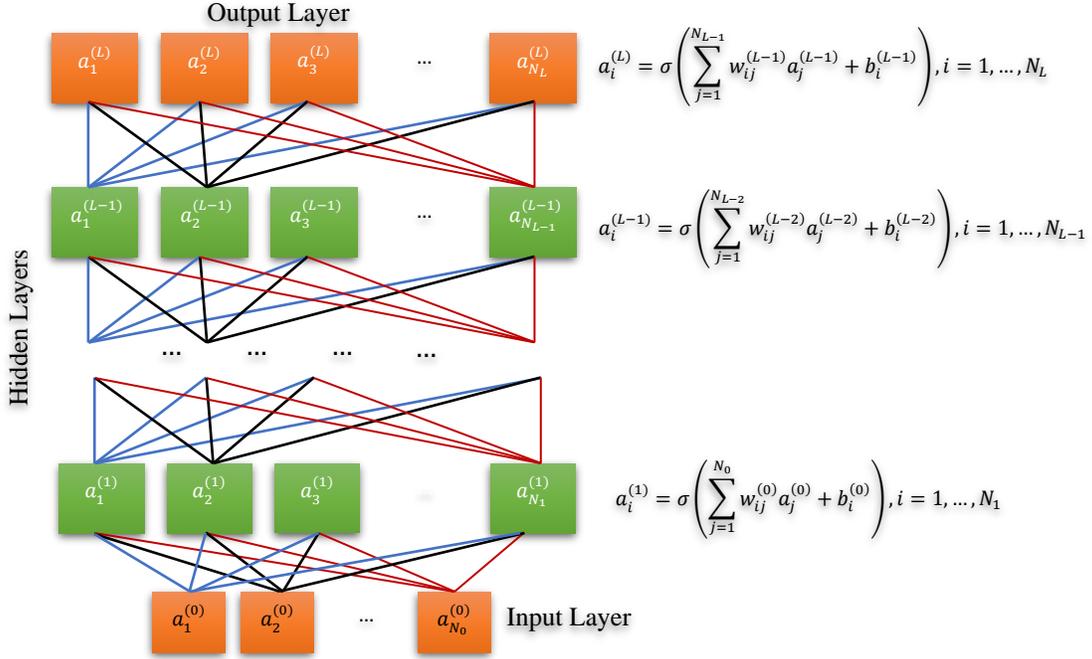

**Fig. 12.** Structure of a MLP. The diagram represents a MLP with one input layer, several hidden layers, and one output layer. The input layer consists of nodes $a_1^{(0)}$, $a_2^{(0)}$, …, $a_{N_0}^{(0)}$, which receive the input features. Each hidden layer $m$ has nodes $a_1^{(m)}$, $a_2^{(m)}$, …, $a_{N_m}^{(m)}$, where each node applies a weighted sum of the inputs from the previous layer, adds a bias term, and passes the result through an AF ($\sigma$). The output layer nodes $a_1^{(L)}$, $a_2^{(L)}$, …, $a_{N_L}^{(L)}$ produce the final output of the network. This structure enables the MLP to learn and model complex patterns in the data through a series of linear transformations and non-linear activations.

### 5.1 The derivation of CBP assumes the existence of the complex derivative $\sigma'(z) = d\sigma(z)/dz$ of the AF $\sigma(z)$

A multilayer perceptron consists of many adaptive linear combiners, each of which has a nonlinearity at its output as shown in Fig. 12. The input/output relationship of such a unit is characterized by the nonlinear recursive difference equation

$$a_i^{(l+1)} = \sigma\left(\sum_{j=1}^{N_l} w_{ij}^{(l)} a_j^{(l)} + b_i^{(l)}\right), \tag{48}$$

and this relation is generalized to all units in multilayer perceptron as listed in Fig. 12.

The error signal $\epsilon_j$, required for adaptation is defined as the difference between the desired response and the output of the perceptron:

$$\epsilon_j(n) = d_j(n) - o_j(n), \qquad j = 1,2,\ldots,N_L, \tag{49}$$

where $d_j(n)$ is the desired response at the $j$-th node of the output layer at time $n$; $o_j(n) = a_j^{(L)}(n)$ is the output at the $j$-th node of the output layer, and $N_L$ is the number of nodes in the $L$-th layer or the output layer. Hence, the sum of error squares produced by the network is [19]

$$\mathcal{L}(n) = \sum_{j=1}^{N_L} \epsilon_j(n)\epsilon_j^*(n). \tag{50}$$

The BP algorithm minimizes the cost functional $\mathcal{L}(n)$ by recursively altering the coefficient $\{w_{ij}^{(l)}, b_i^{(l)}\}$ based on the gradient search technique. Thus, finding the gradient vector of $\mathcal{L}(n)$ is the main idea of deriving the BP algorithm. We first find the partial derivative of $\mathcal{L}(n)$ with respect to the coefficients of the output layer, and then extend to the coefficient of all hidden units. Since $\mathcal{L}(n)$ is a real-valued function which is not analytic, we need to derive the partial derivative of $\mathcal{L}(n)$ with respect to the real and imaginary parts of the coefficients separately. Writing $w_{ij}^{(l)}(n)$ as

$$w_{ij}^{(l)}(n) = w_{ij}^{\Re,(l)}(n) + iw_{ij}^{\Im,(l)}(n), \tag{51}$$

our purpose is to obtain $\partial \mathcal{L}(n)/\partial w_{ij}^{\Re,(l)}$ and $\partial \mathcal{L}(n)/\partial w_{ij}^{\Im,(l)}$. First, let us consider the update rule of the output layer

$$w_{ij}^{\Re,(L-1)}(n+1) = w_{ij}^{\Re,(L-1)}(n) - \frac{1}{2}\alpha \frac{\partial \mathcal{L}(n)}{\partial w_{ij}^{\Re,(L-1)}(n)}, \tag{52.1}$$

$$w_{ij}^{\Im,(L-1)}(n+1) = w_{ij}^{\Im,(L-1)}(n) - \frac{1}{2}\alpha \frac{\partial \mathcal{L}(n)}{\partial w_{ij}^{\Im,(L-1)}(n)}. \tag{52.2}$$

Combining (52.1) and (52.2), we have

$$w_{ij}^{\Re,(L-1)}(n+1) + iw_{ij}^{\Im,(L-1)}(n+1) = w_{ij}^{\Re,(L-1)}(n) + iw_{ij}^{\Im,(L-1)}(n) - \frac{1}{2}\alpha \left( \frac{\partial \mathcal{L}(n)}{\partial w_{ij}^{\Re,(L-1)}(n)} + i\frac{\partial \mathcal{L}(n)}{\partial w_{ij}^{\Im,(L-1)}(n)} \right), \tag{53.1}$$

or

$$w_{ij}^{(L-1)}(n+1) = w_{ij}^{(L-1)}(n) - \frac{1}{2}\alpha \left( \frac{\partial \mathcal{L}(n)}{\partial w_{ij}^{\Re,(L-1)}(n)} + i\frac{\partial \mathcal{L}(n)}{\partial w_{ij}^{\Im,(L-1)}(n)} \right). \tag{53.2}$$

Thus, the next step is to find some expressions for the partial derivative:

$$\frac{\partial \mathcal{L}(n)}{\partial w_{ij}^{\Re,(L-1)}} = \frac{\partial \mathcal{L}(n)}{\partial o_i} \frac{\partial o_i}{\partial z_i^{(L)}} \frac{\partial z_i^{(L)}}{\partial w_{ij}^{\Re,(L-1)}} + \frac{\partial \mathcal{L}(n)}{\partial o_i^*} \frac{\partial o_i^*}{\partial z_i^{(L)*}} \frac{\partial z_i^{(L)*}}{\partial w_{ij}^{\Re,(L-1)}}, \tag{54.1}$$

where

$$o_i = \sigma(z_i^{(L)}), \tag{54.2}$$

$$z_i^{(L)} = \sum_{j=1}^{N_{L-1}} w_{ij}^{(L-1)} a_j^{(L-1)} + b_i^{(L-1)}, \tag{54.3}$$

and $\sigma$ can be any nonlinear function in principle. Let $\sigma$ be defined as the sigmoidal function $1/1 + \exp(-z_i)$. The exponential function is now utilized in its complex form, employing principal values. It is important to note that the use of the exponential function necessitates careful handling of singularities. This issue is circumvented by scaling the input data to a specific region in the complex plane.

One point that should be noted is the absence of the partial derivatives $\partial o_i/\partial z_i^*$ and $\partial o_i^*/\partial z_i$, in (54.1). More precisely, there should be two more terms in (54.1) when the chain rule is applied. When we look at the definition of $o_i$, the function $\sigma$ is a sigmoidal function which does not contain any complex operation. That is, $\sigma$ maps a real number to a real number and a complex number to a complex number, but it will not map a real number to a complex number. Thus

$$a_i^{*(l+1)} = \sigma(z_i^{*(l+1)}), \quad l = 0,1,\ldots,L-1, \tag{55}$$

and hence those two derivatives are equal to zero.

Evaluating all the partial derivatives in (54.1) yields

$$\frac{\partial \mathcal{L}(n)}{\partial w_{ij}^{\Re,(L-1)}} = -(d_i^* - o_i^*)\sigma'(z_i)a_j^{(L-1)} - (d_i - o_i)\sigma'(z_i^*)a_j^{*(L-1)}. \tag{56.1}$$

Similarly

$$\frac{\partial \mathcal{L}(n)}{\partial w_{ij}^{\Im,(L-1)}} = -i(d_i^* - o_i^*)\sigma'(z_i)a_j^{(L-1)} + i(d_i - o_i)\sigma'(z_i^*)a_j^{*(L-1)}. \tag{56.2}$$

Combining (56.1) and (56.2) yields

$$\frac{\partial \mathcal{L}(n)}{\partial w_{ij}^{\Re,(L-1)}} + i\frac{\partial \mathcal{L}(n)}{\partial w_{ij}^{\Im,(L-1)}} = -2(d_i - o_i)\sigma'(z_i^*)a_j^{*(L-1)}. \tag{57}$$

Substituting (57) into (53.2) yields the adaptation rule for the output layer:

$$w_{ij}^{(L-1)}(n+1) = w_{ij}^{(L-1)}(n) + \alpha\big(d_i(n) - o_i(n)\big)\sigma'\big(z_i^*(n)\big)a_j^{*(L-1)}(n),$$
$$i = j, \ldots, N_L;$$
$$j = 1,2,\ldots, N_{L-1}, \tag{58}$$

where $N_L$ and $N_{L-1}$, are the numbers of nodes in layer $L$ (output layer) and layer $L-1$ (the hidden layer converted to the output layer), respectively. The same procedure is now applied to the $(L-1)$-th layer. The generalization to other hidden layers is trivial and will be given at the end of this section. The adaptation rule for this hidden layer is the same as (53.2). Thus, what we need to do is to find the partial derivatives

$$\frac{\partial \mathcal{L}(n)}{\partial w_{ij}^{\Re,(L-2)}} = \frac{\partial \mathcal{L}(n)}{\partial a_i^{(L-1)}} \frac{\partial a_i^{(L-1)}}{\partial z_i^{(L-1)}} \frac{\partial z_i^{(L-1)}}{\partial w_{ij}^{\Re,(L-2)}} + \frac{\partial \mathcal{L}(n)}{\partial a_i^{*(L-1)}} \frac{\partial a_i^{*(L-1)}}{\partial z_i^{*(L-1)}} \frac{\partial z_i^{*(L-1)}}{\partial w_{ij}^{\Re,(L-2)}}. \tag{59}$$

Since $\mathcal{L}(n)$ is not related to $a_i^{(L-1)}$ or $a_i^{*(L-1)}$ explicitly, the evaluation of the above equation requires the use of the chain rule again. Consider

$$\frac{\partial \mathcal{L}(n)}{\partial a_i^{(L-1)}} = -\sum_k (d_k^* - o_k^*) \frac{\partial o_k}{\partial a_i^{(L-1)}}, \tag{60}$$

where

$$o_k = \sigma(z_k^{(L)}) = \sigma\left(\sum_{l=1}^{N_{L-1}} w_{kl}^{(L-1)} a_i^{(L-1)} + b_k^{(L-1)}\right). \tag{61}$$

Hence

$$\frac{\partial \mathcal{L}(n)}{\partial a_i^{(L-1)}} = -\sum_k (d_k^* - o_k^*)\sigma'(z_k^{(L)}) w_{ki}^{(L-1)}. \tag{62.1}$$

Similarly,

$$\frac{\partial \mathcal{L}(n)}{\partial a_i^{*(L-1)}} = -\sum_k (d_k - o_k)\sigma'(z_k^{*(L)}) w_{ki}^{*(L-1)}. \tag{62.2}$$

Substituting (62.1) and (62.2) into (59) yields

$$\frac{\partial \mathcal{L}(n)}{\partial w_{ij}^{\Re,(L-2)}} = \left[-\sum_k (d_k^* - o_k^*)\sigma'(z_k^{(L)}) w_{ki}^{(L-1)}\right]\sigma'(z_i^{(L-1)})a_j^{(L-2)}$$
$$+ \left[-\sum_k (d_k - o_k)\sigma'(z_k^{*(L)}) w_{ki}^{*(L-1)}\right]\sigma'(z_i^{*(L-1)})a_j^{*(L-2)}. \tag{63.1}$$

The partial derivative with respect to the imaginary part is then

$$\frac{\partial \mathcal{L}(n)}{\partial w_{ij}^{\Im,(L-2)}} = i\left[-\sum_k (d_k^* - o_k^*)\sigma'(z_k^{(L)}) w_{ki}^{(L-1)}\right]\sigma'(z_i^{(L-1)})a_j^{(L-2)}$$
$$- i\left[-\sum_k (d_k - o_k)\sigma'(z_k^{*(L)}) w_{ki}^{*(L-1)}\right]\sigma'(z_i^{*(L-1)})a_j^{*(L-2)}. \tag{63.2}$$

Combining (63.1) and (63.2) into complex form yields the adaptation rule

$$w_{ij}^{(L-2)}(n+1) = w_{ij}^{(L-2)}(n) + \alpha \left[\sum_k (d_k - o_k)\sigma'(z_k^{*(L)}) w_{ki}^{*(L-1)}\right]\sigma'(z_i^{*(L-1)})a_j^{*(L-2)}. \tag{64}$$

Equations (64) and (58) are the two basic update equations for the CBP algorithm. The logic of deriving the update formula for other hidden units is exactly the same.

**5.2 The derivation of CBP is based on the assumption that the partial derivatives $\frac{\partial u}{\partial x}, \frac{\partial u}{\partial y}, \frac{\partial v}{\partial x},$ and $\frac{\partial v}{\partial y}$ of the AF $\sigma(z) = u(x,y) + iv(x,y)$ exists. This assumption is not sufficient that $\sigma'(z)$ exists.**

Let the neuron AF be

$$\sigma(z) = u(x, y) + iv(x, y), \quad z = x + iy. \tag{65}$$

The functions $u$ and $v$ are the real and imaginary parts of $\sigma$, respectively. Likewise, $x$ and $y$ are the real and imaginary parts of $z$. For now, it is sufficient to assume that the partial derivatives $u_x = \partial u/\partial x$, $u_y = \partial u/\partial y$, $v_x = \partial v/\partial x$, and $v_y = \partial v/\partial y$ exist for all $z \in \mathbb{C}$. This assumption is not sufficient that $\sigma'(z)$ exists. If it exists, the partial derivatives also have to satisfy the Cauchy-Riemann equations.

The error signal $\epsilon_k$, required for adaptation is defined as the difference between the desired output at the $k$-th neuron of the output layer, $d_k$, and the actual output of the $k$-th output neuron, $o_k$:

$$\epsilon_k = d_k - o_k. \tag{66}$$

Hence, the sum of error squares produced by the network is

$$\begin{aligned}
\mathcal{L} &= \frac{1}{2}\sum_{k=1}|\epsilon_k|^2 = \frac{1}{2}\sum_{k=1}\epsilon_k \epsilon_k^* \\
&= \frac{1}{2}\sum_{k=1}(d_k - o_k)(d_k - o_k)^* \\
&= \frac{1}{2}\sum_{k=1}(d_k - o_k)(d_k^* - o_k^*) \\
&= \frac{1}{2}\sum_{k=1} d_k d_k^* - o_k d_k^* - d_k o_k^* + o_k o_k^* \\
&= \frac{1}{2}\sum_{k=1}(d_k d_k^* + o_k o_k^*) - (o_k d_k^* + d_k o_k^*).
\end{aligned} \tag{67}$$

It should be noted that the error $\mathcal{L}$ is a real scalar function. In what follows, the subscripts $\mathfrak{R}$ and $\mathfrak{I}$ indicate the real and imaginary parts, respectively. The output $o_j$ of neuron $j$ in the network is [20]

$$o_j = \sigma(z_j) = u_j + iv_j, \tag{68}$$

and

$$z_j = x_j + iy_j = \sum_{l=1} w_{jl} a_{jl}, \tag{69.1}$$

where the $w_{jl}$'s are the (complex) weights of neuron $j$

$$w_{jl} = w_{jl}^{\mathfrak{R}} + iw_{jl}^{\mathfrak{I}}, \tag{69.2}$$

and $a_{jl}$'s their corresponding (complex) inputs

$$a_{jl} = a_{jl}^{\mathfrak{R}} + ia_{jl}^{\mathfrak{I}}. \tag{69.3}$$

Hence,

$$z_j = x_j + iy_j = \sum_{l=1}(w_{jl}^{\mathfrak{R}} + iw_{jl}^{\mathfrak{I}})(a_{jl}^{\mathfrak{R}} + ia_{jl}^{\mathfrak{I}}) = \sum_{l=1}(w_{jl}^{\mathfrak{R}} a_{jl}^{\mathfrak{R}} - w_{jl}^{\mathfrak{I}} a_{jl}^{\mathfrak{I}}) + i(w_{jl}^{\mathfrak{I}} a_{jl}^{\mathfrak{R}} + w_{jl}^{\mathfrak{R}} a_{jl}^{\mathfrak{I}}). \tag{70}$$

We get,

$$x_j = \sum_{l=1} w_{jl}^{\mathfrak{R}} a_{jl}^{\mathfrak{R}} - w_{jl}^{\mathfrak{I}} a_{jl}^{\mathfrak{I}}, \tag{71.1}$$

$$y_j = \sum_{l=1} w_{jl}^{\mathfrak{I}} a_{jl}^{\mathfrak{R}} + w_{jl}^{\mathfrak{R}} a_{jl}^{\mathfrak{I}}. \tag{71.2}$$

A bias weight, having permanent input $(1,0)$, may also be added to each neuron. We note the following partial derivatives:

$$\frac{\partial x_j}{\partial w_{jl}^{\mathfrak{R}}} = a_{jl}^{\mathfrak{R}}, \quad \frac{\partial y_j}{\partial w_{jl}^{\mathfrak{R}}} = a_{jl}^{\mathfrak{I}}, \quad \frac{\partial x_j}{\partial w_{jl}^{\mathfrak{I}}} = -a_{jl}^{\mathfrak{I}}, \quad \frac{\partial y_j}{\partial w_{jl}^{\mathfrak{I}}} = a_{jl}^{\mathfrak{R}}. \tag{72}$$

In order to use the chain rule to find the gradient of the error function $\mathcal{L}$ with respect to the real part of $w_{jl}$, we have to observe the variable dependencies: The real function $\mathcal{L}$ is a function of both $u_j(x_j, y_j)$ and $v_j(x_j, y_j)$, and $x_j$ and $y_j$ are both functions of $w_{jl}^{\mathfrak{R}}$ (and $w_{jl}^{\mathfrak{I}}$). Thus, the gradient of the error function with respect to the real part of $w_{jl}$ can be written as

$$\frac{\partial \mathcal{L}}{\partial w_{jl}^{\Re}} = \frac{\partial \mathcal{L}}{\partial u_j}\left(\frac{\partial u_j}{\partial x_j}\frac{\partial x_j}{\partial w_{jl}^{\Re}} + \frac{\partial u_j}{\partial y_j}\frac{\partial y_j}{\partial w_{jl}^{\Re}}\right) + \frac{\partial \mathcal{L}}{\partial v_j}\left(\frac{\partial v_j}{\partial x_j}\frac{\partial x_j}{\partial w_{jl}^{\Re}} + \frac{\partial v_j}{\partial y_j}\frac{\partial y_j}{\partial w_{jl}^{\Re}}\right)$$
$$= \frac{\partial \mathcal{L}}{\partial u_j}\left(\frac{\partial u_j}{\partial x_j}a_{jl}^{\Re} + \frac{\partial u_j}{\partial y_j}a_{jl}^{\Im}\right) + \frac{\partial \mathcal{L}}{\partial v_j}\left(\frac{\partial v_j}{\partial x_j}a_{jl}^{\Re} + \frac{\partial v_j}{\partial y_j}a_{jl}^{\Im}\right)$$
$$= -\delta_j^{\Re}\left(\frac{\partial u_j}{\partial x_j}a_{jl}^{\Re} + \frac{\partial u_j}{\partial y_j}a_{jl}^{\Im}\right) - \delta_j^{\Im}\left(\frac{\partial v_j}{\partial x_j}a_{jl}^{\Re} + \frac{\partial v_j}{\partial y_j}a_{jl}^{\Im}\right), \tag{73}$$

with $\delta_j^{\Re} = -\partial \mathcal{L}/\partial u_j$ and $\delta_j^{\Im} = -\partial \mathcal{L}/\partial v_j$ and consequently

$$\delta_j = \delta_j^{\Re} + i\delta_j^{\Im} = -\frac{\partial \mathcal{L}}{\partial u_j} - i\frac{\partial \mathcal{L}}{\partial v_j}. \tag{74}$$

Likewise, the gradient of the error function with respect to the imaginary part of $w_{jl}$ is

$$\frac{\partial \mathcal{L}}{\partial w_{jl}^{\Im}} = \frac{\partial \mathcal{L}}{\partial u_j}\left(\frac{\partial u_j}{\partial x_j}\frac{\partial x_j}{\partial w_{jl}^{\Im}} + \frac{\partial u_j}{\partial y_j}\frac{\partial y_j}{\partial w_{jl}^{\Im}}\right) + \frac{\partial \mathcal{L}}{\partial v_j}\left(\frac{\partial v_j}{\partial x_j}\frac{\partial x_j}{\partial w_{jl}^{\Im}} + \frac{\partial v_j}{\partial y_j}\frac{\partial y_j}{\partial w_{jl}^{\Im}}\right)$$
$$= \frac{\partial \mathcal{L}}{\partial u_j}\left(\frac{\partial u_j}{\partial x_j}(-a_{jl}^{\Im}) + \frac{\partial u_j}{\partial y_j}a_{jl}^{\Re}\right) + \frac{\partial \mathcal{L}}{\partial v_j}\left(\frac{\partial v_j}{\partial x_j}(-a_{jl}^{\Im}) + \frac{\partial v_j}{\partial y_j}a_{jl}^{\Re}\right)$$
$$= -\delta_j^{\Re}\left(\frac{\partial u_j}{\partial x_j}(-a_{jl}^{\Im}) + \frac{\partial u_j}{\partial y_j}a_{jl}^{\Re}\right) - \delta_j^{\Im}\left(\frac{\partial v_j}{\partial x_j}(-a_{jl}^{\Im}) + \frac{\partial v_j}{\partial y_j}a_{jl}^{\Re}\right). \tag{75}$$

Combining (73) and (75), we can write the gradient of the error function $\mathcal{L}$ with respect to the complex weight $w_{jl}$ as [20]

$$\nabla_{w_{jl}}\mathcal{L} = \frac{\partial \mathcal{L}}{\partial w_{jl}^{\Re}} + i\frac{\partial \mathcal{L}}{\partial w_{jl}^{\Im}} = -\delta_j^{\Re}\left(\frac{\partial u_j}{\partial x_j}a_{jl}^{\Re} + \frac{\partial u_j}{\partial y_j}a_{jl}^{\Im}\right) - \delta_j^{\Im}\left(\frac{\partial v_j}{\partial x_j}a_{jl}^{\Re} + \frac{\partial v_j}{\partial y_j}a_{jl}^{\Im}\right)$$
$$-i\delta_j^{\Re}\left(\frac{\partial u_j}{\partial x_j}(-a_{jl}^{\Im}) + \frac{\partial u_j}{\partial y_j}a_{jl}^{\Re}\right) - i\delta_j^{\Im}\left(\frac{\partial v_j}{\partial x_j}(-a_{jl}^{\Im}) + \frac{\partial v_j}{\partial y_j}a_{jl}^{\Re}\right)$$
$$= -\delta_j^{\Re}\frac{\partial u_j}{\partial x_j}a_{jl}^{\Re} - \delta_j^{\Re}\frac{\partial u_j}{\partial y_j}a_{jl}^{\Im} - \delta_j^{\Im}\frac{\partial v_j}{\partial x_j}a_{jl}^{\Re} - \delta_j^{\Im}\frac{\partial v_j}{\partial y_j}a_{jl}^{\Im}$$
$$+i\delta_j^{\Re}\frac{\partial u_j}{\partial x_j}a_{jl}^{\Im} - i\delta_j^{\Re}\frac{\partial u_j}{\partial y_j}a_{jl}^{\Re} + i\delta_j^{\Im}\frac{\partial v_j}{\partial x_j}a_{jl}^{\Im} - i\delta_j^{\Im}\frac{\partial v_j}{\partial y_j}a_{jl}^{\Re}$$
$$= \delta_j^{\Re}\left(-\frac{\partial u_j}{\partial x_j}a_{jl}^{\Re} - \frac{\partial u_j}{\partial y_j}a_{jl}^{\Im} + i\frac{\partial u_j}{\partial x_j}a_{jl}^{\Im} - i\frac{\partial u_j}{\partial y_j}a_{jl}^{\Re}\right)$$
$$+\delta_j^{\Im}\left(-\frac{\partial v_j}{\partial x_j}a_{jl}^{\Re} - \frac{\partial v_j}{\partial y_j}a_{jl}^{\Im} + i\frac{\partial v_j}{\partial x_j}a_{jl}^{\Im} - i\frac{\partial v_j}{\partial y_j}a_{jl}^{\Re}\right)$$
$$= \delta_j^{\Re}\Theta_1 + \delta_j^{\Im}\Theta_2.$$

Hence, we have

$$\nabla_{w_{jl}}\mathcal{L} = \delta_j^{\Re}\Theta_1 + \delta_j^{\Im}\Theta_2, \tag{76.1}$$

where,

$$\Theta_1 = \left(-\frac{\partial u_j}{\partial x_j}a_{jl}^{\Re} - \frac{\partial u_j}{\partial y_j}a_{jl}^{\Im} + i\frac{\partial u_j}{\partial x_j}a_{jl}^{\Im} - i\frac{\partial u_j}{\partial y_j}a_{jl}^{\Re}\right)$$
$$= \left(-\left[\frac{\partial u_j}{\partial x_j} + i\frac{\partial u_j}{\partial y_j}\right]a_{jl}^{\Re} + \frac{1}{i}\left[-i\frac{\partial u_j}{\partial y_j} - \frac{\partial u_j}{\partial x_j}\right]a_{jl}^{\Im}\right)$$
$$= -\left[\frac{\partial u_j}{\partial x_j} + i\frac{\partial u_j}{\partial y_j}\right](a_{jl}^{\Re} - ia_{jl}^{\Im})$$
$$= -\left[\frac{\partial u_j}{\partial x_j} + i\frac{\partial u_j}{\partial y_j}\right]\bar{a}_{jl},$$

$$\tag{76.2}$$

and

$$\begin{aligned}
\Theta_2 &= \left(-\frac{\partial v_j}{\partial x_j}a_{jl}^{\Re} - \frac{\partial v_j}{\partial y_j}a_{jl}^{\Im} + i\frac{\partial v_j}{\partial x_j}a_{jl}^{\Im} - i\frac{\partial v_j}{\partial y_j}a_{jl}^{\Re}\right) \\
&= \left(-\left[\frac{\partial v_j}{\partial x_j} + i\frac{\partial v_j}{\partial y_j}\right]a_{jl}^{\Re} + \frac{1}{i}\left[-i\frac{\partial v_j}{\partial y_j} - \frac{\partial v_j}{\partial x_j}\right]a_{jl}^{\Im}\right) \\
&= -\left[\frac{\partial v_j}{\partial x_j} + i\frac{\partial v_j}{\partial y_j}\right](a_{jl}^{\Re} - ia_{jl}^{\Im}) \\
&= -\left[\frac{\partial v_j}{\partial x_j} + i\frac{\partial v_j}{\partial y_j}\right]\bar{a}_{jl}.
\end{aligned}$$

(76.3)

Finally, the gradient of the error function $\mathcal{L}$ with respect to the complex weight becomes

$$\begin{aligned}
\nabla_{w_{jl}}\mathcal{L} &= -\left[\frac{\partial u_j}{\partial x_j} + i\frac{\partial u_j}{\partial y_j}\right]\bar{a}_{jl}\delta_j^{\Re} - \left[\frac{\partial v_j}{\partial x_j} + i\frac{\partial v_j}{\partial y_j}\right]\bar{a}_{jl}\delta_j^{\Im} \\
&= -\bar{a}_{jl}\left\{\left[\frac{\partial u_j}{\partial x_j} + i\frac{\partial u_j}{\partial y_j}\right]\delta_j^{\Re} + \left[\frac{\partial v_j}{\partial x_j} + i\frac{\partial v_j}{\partial y_j}\right]\delta_j^{\Im}\right\}.
\end{aligned}$$

(77)

To minimize the error $\mathcal{L}$, each complex weight $w_{jl}$ should be changed by a quantity $\Delta w_{jl}$ proportional to the negative gradient:

$$\begin{aligned}
\Delta w_{jl} &= w_{jl,\text{new}} - w_{jl,\text{old}} \\
&= -\alpha\nabla_{w_{jl}}\mathcal{L} \\
&= \alpha\bar{a}_{jl}\left\{\left[\frac{\partial u_j}{\partial x_j} + i\frac{\partial u_j}{\partial y_j}\right]\delta_j^{\Re} + \left[\frac{\partial v_j}{\partial x_j} + i\frac{\partial v_j}{\partial y_j}\right]\delta_j^{\Im}\right\},
\end{aligned}$$

(78)

where $\alpha$ is the learning rate, a real positive constant. A momentum term may be added to the above learning equation. When the weight $w_{jl}$ belongs to an output neuron, then $\delta_j^{\Re}$, and $\delta_j^{\Im}$ in (77) have the values

$$\delta_j^{\Re} = -\frac{\partial \mathcal{L}}{\partial u_j} = \epsilon_j^{\Re} = d_j^{\Re} - u_j \quad \text{and} \quad \delta_j^{\Im} = -\frac{\partial \mathcal{L}}{\partial v_j} = \epsilon_j^{\Im} = d_j^{\Im} - v_j,$$

(79)

or more compactly:

$$\begin{aligned}
\delta_j = \epsilon_j &= \epsilon_j^{\Re} + i\epsilon_j^{\Im} \\
&= d_j^{\Re} - u_j + i(d_j^{\Im} - v_j) \\
&= (d_j^{\Re} + id_j^{\Im}) - (u_j + iv_j) \\
&= d_j - o_j.
\end{aligned}$$

(80)

When weight $w_{jl}$ belongs to a hidden neuron, i.e., when the output of the neuron is fed to other neurons in subsequent layers, in order to compute $\delta_j$, or equivalently $\delta_j^{\Re}$ and $\delta_j^{\Im}$, we have to use the chain rule. Let the index $k$ indicate a neuron that receives input from neuron $j$. Then the net input $z_k$ to neuron $k$ is

$$\begin{aligned}
z_k &= x_k + iy_k \\
&= \sum_l (u_l + iv_l)(w_{kl}^{\Re} + iw_{kl}^{\Im}) \\
&= \sum_l (u_l w_{kl}^{\Re} - v_l w_{kl}^{\Im}) + i(v_l w_{kl}^{\Re} + u_l w_{kl}^{\Im}),
\end{aligned}$$

(81)

where the index $l$ runs through the neurons from which neuron $k$ receives input. Hence

$$x_k = \sum_l u_l w_{kl}^{\Re} - v_l w_{kl}^{\Im},$$

(82.1)

$$y_k = \sum_l v_l w_{kl}^{\Re} + u_l w_{kl}^{\Im}.$$

(82.2)

Thus, we have the following partial derivatives:

$$\frac{\partial x_k}{\partial u_j} = w_{kj}^{\Re}, \quad \frac{\partial y_k}{\partial u_j} = w_{kj}^{\Im}, \quad \frac{\partial x_k}{\partial v_j} = -w_{kj}^{\Im}, \quad \frac{\partial y_k}{\partial v_j} = w_{kj}^{\Re}.$$

(83)

Using the chain rule, we compute $\delta_j^{\Re}$ [20]:

$$\begin{aligned}
\delta_j^{\Re} &= -\frac{\partial \mathcal{L}}{\partial u_j} \\
&= -\sum_k \frac{\partial \mathcal{L}}{\partial u_k}\left(\frac{\partial u_k}{\partial x_k}\frac{\partial x_k}{\partial u_j} + \frac{\partial u_k}{\partial y_k}\frac{\partial y_k}{\partial u_j}\right) - \sum_k \frac{\partial \mathcal{L}}{\partial v_k}\left(\frac{\partial v_k}{\partial x_k}\frac{\partial x_k}{\partial u_j} + \frac{\partial v_k}{\partial y_k}\frac{\partial y_k}{\partial u_j}\right) \\
&= -\sum_k \frac{\partial \mathcal{L}}{\partial u_k}\left(\frac{\partial u_k}{\partial x_k}w_{kj}^{\Re} + \frac{\partial u_k}{\partial y_k}w_{kj}^{\Im}\right) - \sum_k \frac{\partial \mathcal{L}}{\partial v_k}\left(\frac{\partial v_k}{\partial x_k}w_{kj}^{\Re} + \frac{\partial v_k}{\partial y_k}w_{kj}^{\Im}\right) \\
&= \sum_k \delta_k^{\Re}\left(\frac{\partial u_k}{\partial x_k}w_{kj}^{\Re} + \frac{\partial u_k}{\partial y_k}w_{kj}^{\Im}\right) + \sum_k \delta_k^{\Im}\left(\frac{\partial v_k}{\partial x_k}w_{kj}^{\Re} + \frac{\partial v_k}{\partial y_k}w_{kj}^{\Im}\right),
\end{aligned} \qquad (84)$$

where the index $k$ runs through the neurons that receive input from neuron $j$. In a similar manner, we can compute $\delta_j^{\Im}$:

$$\begin{aligned}
\delta_j^{\Im} &= -\frac{\partial \mathcal{L}}{\partial v_j} \\
&= -\sum_k \frac{\partial \mathcal{L}}{\partial u_k}\left(\frac{\partial u_k}{\partial x_k}\frac{\partial x_k}{\partial v_j} + \frac{\partial u_k}{\partial y_k}\frac{\partial y_k}{\partial v_j}\right) - \sum_k \frac{\partial \mathcal{L}}{\partial v_k}\left(\frac{\partial v_k}{\partial x_k}\frac{\partial x_k}{\partial v_j} + \frac{\partial v_k}{\partial y_k}\frac{\partial y_k}{\partial v_j}\right) \\
&= -\sum_k \frac{\partial \mathcal{L}}{\partial u_k}\left(\frac{\partial u_k}{\partial x_k}(-w_{kj}^{\Im}) + \frac{\partial u_k}{\partial y_k}w_{kj}^{\Re}\right) - \sum_k \frac{\partial \mathcal{L}}{\partial v_k}\left(\frac{\partial v_k}{\partial x_k}(-w_{kj}^{\Im}) + \frac{\partial v_k}{\partial y_k}w_{kj}^{\Re}\right) \\
&= \sum_k \delta_k^{\Re}\left(\frac{\partial u_k}{\partial x_k}(-w_{kj}^{\Im}) + \frac{\partial u_k}{\partial y_k}w_{kj}^{\Re}\right) + \sum_k \delta_k^{\Im}\left(\frac{\partial v_k}{\partial x_k}(-w_{kj}^{\Im}) + \frac{\partial v_k}{\partial y_k}w_{kj}^{\Re}\right).
\end{aligned} \qquad (85)$$

Combining (84) and (85), we arrive at the following expression for $\delta_j$:

$$\begin{aligned}
\delta_j &= \delta_j^{\Re} + i\delta_j^{\Im} \\
&= \sum_k \delta_k^{\Re}\left(\frac{\partial u_k}{\partial x_k}w_{kj}^{\Re} + \frac{\partial u_k}{\partial y_k}w_{kj}^{\Im}\right) + \sum_k \delta_k^{\Im}\left(\frac{\partial v_k}{\partial x_k}w_{kj}^{\Re} + \frac{\partial v_k}{\partial y_k}w_{kj}^{\Im}\right) \\
&\quad + i\sum_k \delta_k^{\Re}\left(\frac{\partial u_k}{\partial x_k}(-w_{kj}^{\Im}) + \frac{\partial u_k}{\partial y_k}w_{kj}^{\Re}\right) + i\sum_k \delta_k^{\Im}\left(\frac{\partial v_k}{\partial x_k}(-w_{kj}^{\Im}) + \frac{\partial v_k}{\partial y_k}w_{kj}^{\Re}\right) \\
&= \sum_k \left(\delta_k^{\Re}\frac{\partial u_k}{\partial x_k}w_{kj}^{\Re} + \delta_k^{\Re}\frac{\partial u_k}{\partial y_k}w_{kj}^{\Im} + \delta_k^{\Im}\frac{\partial v_k}{\partial x_k}w_{kj}^{\Re} + \delta_k^{\Im}\frac{\partial v_k}{\partial y_k}w_{kj}^{\Im}\right) \\
&\quad + \sum_k \left(-i\delta_k^{\Re}\frac{\partial u_k}{\partial x_k}w_{kj}^{\Im} + i\delta_k^{\Re}\frac{\partial u_k}{\partial y_k}w_{kj}^{\Re} - i\delta_k^{\Im}\frac{\partial v_k}{\partial x_k}w_{kj}^{\Im} + i\delta_k^{\Im}\frac{\partial v_k}{\partial y_k}w_{kj}^{\Re}\right) \\
&= \sum_k \left(\delta_k^{\Re}\frac{\partial u_k}{\partial x_k}[w_{kj}^{\Re} - iw_{kj}^{\Im}] + \delta_k^{\Re}\frac{\partial u_k}{\partial y_k}[w_{kj}^{\Im} + iw_{kj}^{\Re}] + \delta_k^{\Im}\frac{\partial v_k}{\partial x_k}[w_{kj}^{\Re} - iw_{kj}^{\Im}] + \delta_k^{\Im}\frac{\partial v_k}{\partial y_k}[w_{kj}^{\Im} + iw_{kj}^{\Re}]\right) \\
&= \sum_k \left(\delta_k^{\Re}\frac{\partial u_k}{\partial x_k}[w_{kj}^{\Re} - iw_{kj}^{\Im}] + i\delta_k^{\Re}\frac{\partial u_k}{\partial y_k}[w_{kj}^{\Re} - iw_{kj}^{\Im}] + \delta_k^{\Im}\frac{\partial v_k}{\partial x_k}[w_{kj}^{\Re} - iw_{kj}^{\Im}] + i\delta_k^{\Im}\frac{\partial v_k}{\partial y_k}[w_{kj}^{\Re} - iw_{kj}^{\Im}]\right) \\
&= \sum_k [w_{kj}^{\Re} - iw_{kj}^{\Im}]\left(\left[\frac{\partial u_k}{\partial x_k} + i\frac{\partial u_k}{\partial y_k}\right]\delta_k^{\Re} + \left[\frac{\partial v_k}{\partial x_k} + i\frac{\partial v_k}{\partial y_k}\right]\delta_k^{\Im}\right) \\
&= \sum_k \overline{w}_{kj}\left(\left[\frac{\partial u_k}{\partial x_k} + i\frac{\partial u_k}{\partial y_k}\right]\delta_k^{\Re} + \left[\frac{\partial v_k}{\partial x_k} + i\frac{\partial v_k}{\partial y_k}\right]\delta_k^{\Im}\right),
\end{aligned}$$

Hence, we have

$$\delta_j = \sum_k \overline{w}_{kj}\left(\left[\frac{\partial u_k}{\partial x_k} + i\frac{\partial u_k}{\partial y_k}\right]\delta_k^{\Re} + \left[\frac{\partial v_k}{\partial x_k} + i\frac{\partial v_k}{\partial y_k}\right]\delta_k^{\Im}\right). \qquad (86)$$

Training of a feed-forward network with the CBP algorithm is done in a similar manner as in the usual real BP.

- First, the weights are initialized to small random complex values.
- Until an acceptable output error level is arrived at, each input vector is presented to the network.
- The corresponding output and output error are calculated (forward pass).
- Then the error is backpropagated to each neuron in the network and the weights are adjusted accordingly (backward pass).

- More precisely, for the input pattern $a_j$, $\delta_j$ for neuron $j$ is computed by starting at the neurons in the output layer using (80) and then for neurons in hidden layers by recursively using (86). As soon as $\delta_j$ is computed for neuron $j$, its weights are changed according to (78).

## 5.3 Cauchy-Riemann equations and the CBP algorithm

Cauchy-Riemann equations can be used to simplify the CBP algorithm derived in subsection 5.2. Cauchy-Riemann equations are the necessary condition for a complex function to be analytic at a point $z \in \mathbb{C}$ and can be written by noting that the partial derivatives of $\sigma(z) = u(x,y) + iv(x,y)$, where $z = x + iy$, should be equal along the real and imaginary axes:

$$\sigma'(z) = \frac{\partial u}{\partial x} + i\frac{\partial v}{\partial x} = \frac{\partial v}{\partial y} - i\frac{\partial u}{\partial y}. \tag{87}$$

Equating the real and imaginary parts in (87), we obtain the Cauchy-Riemann equations: $\frac{\partial u}{\partial x} = \frac{\partial v}{\partial y}, \frac{\partial v}{\partial x} = -\frac{\partial u}{\partial y}$. Also note that this enables (87) to be expressed more concisely as (using Wirtinger derivatives) [21]

$$\begin{aligned}\sigma'(z) = \frac{\partial \sigma}{\partial z} &= \frac{1}{2}\left(\frac{\partial \sigma}{\partial x} - i\frac{\partial \sigma}{\partial y}\right) \\
&= \frac{1}{2}\left(\frac{\partial}{\partial x}[u(x,y) + iv(x,y)] - i\frac{\partial}{\partial y}[u(x,y) + iv(x,y)]\right) \\
&= \frac{1}{2}\left(\left[\frac{\partial}{\partial x}u(x,y) + i\frac{\partial}{\partial x}v(x,y)\right] - i\left[-\frac{\partial}{\partial x}v(x,y) + i\frac{\partial}{\partial x}u(x,y)\right]\right) \\
&= \frac{1}{2}\left(\frac{\partial}{\partial x}u(x,y) + i\frac{\partial}{\partial x}v(x,y) + i\frac{\partial}{\partial x}v(x,y) + \frac{\partial}{\partial x}u(x,y)\right) \\
&= \frac{\partial}{\partial x}(u(x,y) + iv(x,y)) \\
&= \frac{\partial \sigma}{\partial x},\end{aligned}$$

and

$$\begin{aligned}\sigma'(z) = \frac{\partial \sigma}{\partial z} &= \frac{1}{2}\left(\frac{\partial \sigma}{\partial x} - i\frac{\partial \sigma}{\partial y}\right) \\
&= \frac{1}{2}\left(\left[\frac{\partial}{\partial x}u(x,y) + i\frac{\partial}{\partial x}v(x,y)\right] - i\left[\frac{\partial}{\partial y}u(x,y) + i\frac{\partial}{\partial y}v(x,y)\right]\right) \\
&= \frac{1}{2}\left(\left[\frac{\partial}{\partial y}v(x,y) - i\frac{\partial}{\partial y}u(x,y)\right] - i\left[\frac{\partial}{\partial y}u(x,y) + i\frac{\partial}{\partial y}v(x,y)\right]\right) \\
&= \frac{1}{2}\left(\frac{\partial}{\partial y}v(x,y) - i\frac{\partial}{\partial y}u(x,y) - i\frac{\partial}{\partial y}u(x,y) + \frac{\partial}{\partial y}v(x,y)\right) \\
&= -i\frac{\partial}{\partial y}(u(x,y) + iv(x,y)) \\
&= -i\frac{\partial \sigma}{\partial y}.\end{aligned}$$

Therefore, we arrive at the following equation:

$$\sigma'(z) = \frac{\partial \sigma}{\partial x} = -i\frac{\partial \sigma}{\partial y}. \tag{88.1}$$

Similarly,

$$\begin{aligned}\bar{\sigma}'(z) = \overline{\left(\frac{\partial \sigma}{\partial z}\right)} &= \frac{1}{2}\left(\frac{\partial \sigma}{\partial x} - i\frac{\partial \sigma}{\partial y}\right)^* \\
&= \frac{1}{2}\left(\frac{\partial}{\partial x}[u(x,y) + iv(x,y)]^* + i\frac{\partial}{\partial y}[u(x,y) + iv(x,y)]^*\right) \\
&= \frac{1}{2}\left(\left[\frac{\partial}{\partial x}u(x,y) - i\frac{\partial}{\partial x}v(x,y)\right] + i\left[\frac{\partial}{\partial y}u(x,y) - i\frac{\partial}{\partial y}v(x,y)\right]\right)\end{aligned}$$

$$\begin{aligned}
&= \frac{1}{2}\left(\left[\frac{\partial}{\partial x}u(x,y) - i\frac{\partial}{\partial x}v(x,y)\right] + i\left[-\frac{\partial}{\partial x}v(x,y) - i\frac{\partial}{\partial x}u(x,y)\right]\right) \\
&= \frac{1}{2}\left(\frac{\partial}{\partial x}u(x,y) - i\frac{\partial}{\partial x}v(x,y) - i\frac{\partial}{\partial x}v(x,y) + \frac{\partial}{\partial x}u(x,y)\right) \\
&= \frac{\partial}{\partial x}(u(x,y) - iv(x,y)) = \frac{\partial \bar{\sigma}}{\partial x},
\end{aligned}$$

and

$$\begin{aligned}
\bar{\sigma}'(z) &= \overline{\left(\frac{\partial \sigma}{\partial z}\right)} = \frac{1}{2}\left(\frac{\partial \sigma}{\partial x} - i\frac{\partial \sigma}{\partial y}\right)^* \\
&= \frac{1}{2}\left(\frac{\partial}{\partial x}[u(x,y) + iv(x,y)]^* + i\frac{\partial}{\partial y}[u(x,y) + iv(x,y)]^*\right) \\
&= \frac{1}{2}\left(\left[\frac{\partial}{\partial x}u(x,y) - i\frac{\partial}{\partial x}v(x,y)\right] + i\left[\frac{\partial}{\partial y}u(x,y) - i\frac{\partial}{\partial y}v(x,y)\right]\right) \\
&= \frac{1}{2}\left(\left[\frac{\partial}{\partial y}v(x,y) + i\frac{\partial}{\partial y}u(x,y)\right] + i\left[\frac{\partial}{\partial y}u(x,y) - i\frac{\partial}{\partial y}v(x,y)\right]\right) \\
&= \frac{1}{2}\left(\frac{\partial}{\partial y}v(x,y) + i\frac{\partial}{\partial y}u(x,y) + i\frac{\partial}{\partial y}u(x,y) + \frac{\partial}{\partial y}v(x,y)\right) \\
&= i\frac{\partial}{\partial y}(u(x,y) - iv(x,y)) = i\frac{\partial \bar{\sigma}}{\partial y}.
\end{aligned}$$

As a result, we obtain the following equation:

$$\bar{\sigma}'(z) = \frac{\partial \bar{\sigma}}{\partial x} = i\frac{\partial \bar{\sigma}}{\partial y}. \tag{88.2}$$

Further applying the Cauchy-Riemann equations to (77), a more compact representation for the gradient of the error function is obtained using the simple derivative form given in (88.1)

$$\begin{aligned}
\nabla_{w_{jl}}\mathcal{L} &= -\bar{a}_{jl}\left\{\left[\frac{\partial u_j}{\partial x_j} + i\frac{\partial u_j}{\partial y_j}\right]\delta_j^{\Re} + \left[\frac{\partial v_j}{\partial x_j} + i\frac{\partial v_j}{\partial y_j}\right]\delta_j^{\Im}\right\} \\
&= -\bar{a}_{jl}\left\{\left[\frac{\partial u_j}{\partial x_j} - i\frac{\partial v_j}{\partial x_j}\right]\delta_j^{\Re} + \left[-\frac{\partial u_j}{\partial y_j} + i\frac{\partial v_j}{\partial y_j}\right]\delta_j^{\Im}\right\} \\
&= -\bar{a}_{jl}\left\{\frac{\partial}{\partial x_j}[u_j - iv_j]\delta_j^{\Re} - \frac{\partial}{\partial y_j}[u_j - iv_j]\delta_j^{\Im}\right\} \\
&= -\bar{a}_{jl}\left\{\frac{\partial \bar{\sigma}}{\partial x_j}\delta_j^{\Re} - \frac{\partial \bar{\sigma}}{\partial y_j}\delta_j^{\Im}\right\} \\
&= -\bar{a}_{jl}\left\{\frac{\partial \bar{\sigma}}{\partial x_j}\delta_j^{\Re} + \left(i\frac{\partial \bar{\sigma}}{\partial y_j}\right)(i\delta_j^{\Im})\right\} \\
&= -\bar{a}_{jl}\left\{\frac{\partial \bar{\sigma}}{\partial x_j}\delta_j^{\Re} + \frac{\partial \bar{\sigma}}{\partial x_j}(i\delta_j^{\Im})\right\} \\
&= -\bar{a}_{jl}\frac{\partial \bar{\sigma}}{\partial x_j}\{\delta_j^{\Re} + i\delta_j^{\Im}\} = -\bar{a}_{jl}\bar{\sigma}'\delta_j,
\end{aligned}$$

where, $\bar{\sigma}' = \frac{\partial \bar{\sigma}}{\partial x_j} = i\frac{\partial \bar{\sigma}}{\partial y_j}$. So, we have [21]

$$\nabla_{w_{jl}}\mathcal{L} = -\bar{a}_{jl}\bar{\sigma}'(z)\delta_j. \tag{89}$$

Complex weight update $\Delta w_{jl}$ is proportional to the negative gradient:

$$\Delta w_{jl} = \alpha \bar{a}_{jl}\bar{\sigma}'(z)\delta_j, \tag{90}$$

where $\alpha$ is a real positive learning rate. When the complex weight belongs to an output neuron:

$$\delta_j = \epsilon_j = d_j - o_j. \tag{91}$$

When weight $w_{jl}$ belongs to a hidden neuron, i.e., when the output of the neuron is fed to other neurons in subsequent layers, in order to compute $\delta_j$, or equivalently $\delta_j^{\Re}$ and $\delta_j^{\Im}$, we have to use the chain rule. Let the index $k$ indicate a neuron that receives input from neuron $j$. Using (88.1) with (86), the following expression is obtained similarly for the weight update function (90) using (89), [21]:

$$\begin{aligned}
\delta_j = \delta_j^{\Re} + i\delta_j^{\Im} &= \sum_k \bar{w}_{kj} \left( \left[\frac{\partial u_k}{\partial x_k} + i\frac{\partial u_k}{\partial y_k}\right]\delta_k^{\Re} + \left[\frac{\partial v_k}{\partial x_k} + i\frac{\partial v_k}{\partial y_k}\right]\delta_k^{\Im} \right) \\
&= \sum_k \bar{w}_{kj} \left( \left[\frac{\partial u_k}{\partial x_k} - i\frac{\partial v_k}{\partial x_k}\right]\delta_k^{\Re} + \left[-\frac{\partial u_k}{\partial y_k} + i\frac{\partial v_k}{\partial y_k}\right]\delta_k^{\Im} \right) \\
&= \sum_k \bar{w}_{kj} \left( \frac{\partial}{\partial x_k}[u_k - iv_k]\delta_k^{\Re} - \frac{\partial}{\partial y_k}[u_k - iv_k]\delta_k^{\Im} \right) \\
&= \sum_k \bar{w}_{kj} \left( \frac{\partial \bar{\sigma}}{\partial x_k}\delta_k^{\Re} - \frac{\partial \bar{\sigma}}{\partial y_k}\delta_k^{\Im} \right) \\
&= \sum_k \bar{w}_{kj} \left( \frac{\partial \bar{\sigma}}{\partial x_k}\delta_k^{\Re} + \left(i\frac{\partial \bar{\sigma}}{\partial y_k}\right)(i\delta_k^{\Im}) \right) \\
&= \sum_k \bar{w}_{kj} \left( \frac{\partial \bar{\sigma}}{\partial x_k}\delta_k^{\Re} + \frac{\partial \bar{\sigma}}{\partial x_j}(i\delta_k^{\Im}) \right) \\
&= \sum_k \bar{w}_{kj} \frac{\partial \bar{\sigma}}{\partial x_k}(\delta_k^{\Re} + i\delta_k^{\Im}) = \sum_k \bar{w}_{kj} \bar{\sigma}' \delta_k,
\end{aligned}$$

where, $\bar{\sigma}' = \frac{\partial \bar{\sigma}}{\partial x_j} = i\frac{\partial \bar{\sigma}}{\partial y_j}$. So, we have

$$\delta_j = \sum_k \bar{w}_{kj} \bar{\sigma}'(z) \delta_k. \tag{92}$$

Compared to the complex activation representation as $\sigma(z) = u(x,y) + iv(x,y)$, the split complex AF is a special case and can be represented as $\sigma(z) = u(x) + iv(y)$. This indicates that $u_y = v_x = 0$ for the split CBP algorithm. Removing these zero terms from (77), we obtain the following gradient and (complex) weight updates:

$$\nabla_{w_{jl}}\mathcal{L} = -\bar{a}_{jl} \left\{ \left[\frac{\partial u_j}{\partial x_j} + i\frac{\partial u_j}{\partial y_j}\right]\delta_j^{\Re} + \left[\frac{\partial v_j}{\partial x_j} + i\frac{\partial v_j}{\partial y_j}\right]\delta_j^{\Im} \right\} = -\bar{a}_{jl} \left\{ \frac{\partial u_j}{\partial x_j}\delta_j^{\Re} + i\frac{\partial v_j}{\partial y_j}\delta_j^{\Im} \right\}, \tag{93}$$

and

$$\Delta w_{jl} = \alpha \bar{a}_{jl} \left\{ \frac{\partial u_j}{\partial x_j}\delta_j^{\Re} + i\frac{\partial v_j}{\partial y_j}\delta_j^{\Im} \right\}. \tag{94}$$

As before, for output layer neurons, $\delta_j = \epsilon_j = d_j - o_j$. For the input and hidden layer, (86) becomes

$$\delta_j = \sum_k \bar{w}_{kj} \left( \left[\frac{\partial u_k}{\partial x_k} + i\frac{\partial u_k}{\partial y_k}\right]\delta_k^{\Re} + \left[\frac{\partial v_k}{\partial x_k} + i\frac{\partial v_k}{\partial y_k}\right]\delta_k^{\Im} \right) = \sum_k \bar{w}_{kj} \left( \frac{\partial u_k}{\partial x_k}\delta_k^{\Re} + i\frac{\partial v_k}{\partial y_k}\delta_k^{\Im} \right). \tag{95}$$

For further details on CBP algorithms, refer to [22-30].

## 6. Properties of Complex AFs

As with any innovative approach, challenges arise during the development and optimization of the CVNNs. One such critical challenge revolves around the choice of AFs within the complex-valued context. AFs play a pivotal role in shaping the non-linear transformations applied to the input data, enabling NNs to learn complex patterns and representations. The standard practice of extending real-valued AFs to the complex domain has been questioned, particularly when applying the Sigmoid function to create a complex-valued saturation function for CVNNs.

The crux of the issue lies in the non-analytic nature of certain CVAF. A complex function is considered analytic (or holomorphic) if it is differentiable at every point within its domain. If $\sigma(z)$ is analytic at all points $z \in \mathbb{C}$, it is called entire. The concept of differentiability at each point categorizes a function as regular at that point, while points, where differentiability fails, are termed singular points. The significance of analyticity lies in its implications for the analytical study of neural dynamics, such as learning, self-organization, and processing. One of the most challenging aspects of CVNNs is designing a suitable AF for CVNNs, as dictated by Liouville's theorem.

***Theorem 5 (Liouville Theorem):*** *If $\sigma(z)$ is entire and bounded on the complex plane, then $\sigma(z)$ is a constant function.*

Since a suitable $\sigma(z)$ must be bounded, it follows from Liouville's theorem that if in addition $\sigma(z)$ is entire, then $\sigma(z)$ is constant, clearly not a suitable AF. In other words, a function that is bounded and analytic everywhere in the complex plane is not a suitable CVAF. In fact, the requirement that $\sigma(z)$ be entire alone imposes considerable structure on it, e.g., the Cauchy-Riemann equations should be satisfied. Therefore, most of the AFs in CVNNs are designed by selecting one property (i.e., boundedness or analytic).

A suitable AF $\sigma(z)$ must possess two properties [20]: (1) $\sigma(z)$ is bounded, and (2) $\sigma(z)$ is such that $\delta_j \neq (0,0)$ and $x_{jl} \neq (0,0)$ imply $\nabla_{w_{jl}} \mathcal{L} \neq (0,0)$. Non-compliance with the latter condition is undesirable since it would imply that even in the presence of both non-zero input $x_{jl} \neq (0,0)$ and non-zero error $\delta_j \neq (0,0)$, it is still possible that $\Delta w_{jl} = 0$, i.e., no learning takes place.

***Lemma 2:*** *If $\sigma(z) = u + iv$, and*

$$\frac{\partial u}{\partial x}\frac{\partial v}{\partial y} = \frac{\partial v}{\partial x}\frac{\partial u}{\partial y}, \tag{96}$$

*then $\sigma(z)$ is not a suitable AF.*

**Proof:** We would like to show that an AF with the above property violates condition (2). Suppose that $x_{jl} \neq (0,0)$. We will show that there exists $\delta_j \neq (0,0)$ such that $\nabla_{w_{jl}} \mathcal{L} = 0$. From (77) we see that $\nabla_{w_{jl}} \mathcal{L} = 0$ when

$$\left\{ \left[\frac{\partial u_j}{\partial x_j} + i\frac{\partial u_j}{\partial y_j}\right]\delta_j^{\Re} + \left[\frac{\partial v_j}{\partial x_j} + i\frac{\partial v_j}{\partial y_j}\right]\delta_j^{\Im} \right\} = 0,$$

or equivalently, when the real and imaginary parts of the equation equal to zero:

$$\frac{\partial u_j}{\partial x_j}\delta_j^{\Re} + \frac{\partial v_j}{\partial x_j}\delta_j^{\Im} = 0,$$

$$\frac{\partial u_j}{\partial y_j}\delta_j^{\Re} + \frac{\partial v_j}{\partial y_j}\delta_j^{\Im} = 0.$$

A nontrivial solution of the above homogeneous system of equations in $\delta_j^{\Re}$ and $\delta_j^{\Im}$, i.e., $\delta_j^{\Re}$ and $\delta_j^{\Im}$ not both zero, can be found if and only if the determinant of the coefficient matrix is zero:

$$\frac{\partial u_j}{\partial x_j}\frac{\partial v_j}{\partial y_j} - \frac{\partial u_j}{\partial y_j}\frac{\partial v_j}{\partial x_j} = 0,$$

or $\frac{\partial u_j}{\partial x_j}\frac{\partial v_j}{\partial y_j} = \frac{\partial u_j}{\partial y_j}\frac{\partial v_j}{\partial x_j}$. ∎

The Sigmoid function, $\sigma_{\text{Sigmoid}}(z) = 1/(1 + e^{-z})$, as well as other commonly used AFs like Tanh(z) and $e^{-z^2}$, are indeed unbounded when their domain is extended from the real line to the complex plane. It can be easily verified that when $z$ approaches any value in the set $\{(0 \pm i(2n + 1)\pi: n \text{ is any integer}\}$, then $|\sigma_{\text{Sigmoid}}(z)| \to \infty$, and thus $\sigma_{\text{Sigmoid}}(z)$ is unbounded. Similarly, one can verify that $|\text{Tanh}(z)| \to \infty$ as $z$ approaches a value in the set $\{(0 \pm i((2n + 1)/2)\pi: n \text{ is any integer}\}$. Additionally, $|e^{-z^2}| \to \infty$ when $z = 0 + iy$ and $y \to \infty$.

In order to avoid the problem of singularities in the Sigmoid function $\sigma_{\text{Sigmoid}}(z)$, it was suggested in [19] to scale "the input data to some region in the complex plane. Scaling the input data to some region in the complex plane could potentially control the magnitude of $z$. However, backpropagation, being a weak optimization procedure, may not enforce constraints on the weights effectively. The weights can still assume arbitrary values, and the combination of inputs and weights, $z$, can result in values that cause numerical issues.

It is clear that some other AF must be found for CBP. In the derivation of CBP, we only assumed that the partial derivatives $\frac{\partial u}{\partial x}$, $\frac{\partial v}{\partial y}$, $\frac{\partial v}{\partial x}$, and $\frac{\partial u}{\partial y}$ exist. Other important properties the AF $\sigma(z) = u(x,y) + iv(x,y)$ should possess are the following [20]:

1. The AF $\sigma(z) = u(x,y) + iv(x,y)$ should be nonlinear in both $x$ and $y$. This is a fundamental requirement for NNs to have the capacity to learn and represent complex, nonlinear relationships in the data.
2. The AF should be bounded. Boundedness is crucial to avoid numerical instability during training. If the AF is unbounded, it might lead to issues like exploding gradient, making the training process difficult. This is true if and only if both $u$ and $v$ bounded. Since both $u$ and $v$ are used during training (forward pass), they must be bounded. If either one was unbounded, then a software overflow could occur.
3. The partial derivatives $\frac{\partial u}{\partial x}$, $\frac{\partial v}{\partial y}$, $\frac{\partial v}{\partial x}$, and $\frac{\partial u}{\partial y}$ must exist and be bounded. This is necessary for the backpropagation algorithm to compute gradients during training. Bounded derivatives contribute to stable learning.
4. $\sigma(z)$ is not entire. See Theorem 5.
5. The relationship $\frac{\partial u}{\partial x}\frac{\partial v}{\partial y} \neq \frac{\partial v}{\partial x}\frac{\partial u}{\partial y}$ should hold. See Lemma 2.

Considering all these properties helps in defining an AF that is suitable for CBP.

Note that by Louiville's theorem, the second and the fourth conditions are redundant, i.e., a bounded nonlinear function in $\mathbb{C}$ cannot be an entire function. Using this fact, You and Hong [40] reduced the above conditions into the four conditions given below to introduce a split complex AF:

1. The AF $\sigma(z) = u(x,y) + iv(x,y)$ is nonlinear in $x$ and $y$.
2. For the stability of a system, $\sigma(z)$ should have no singularities and be bounded for all $z$ in a bounded set.
3. The partial derivatives $\frac{\partial u}{\partial x}$, $\frac{\partial v}{\partial y}$, $\frac{\partial v}{\partial x}$, and $\frac{\partial u}{\partial y}$ should be continuous and bounded.
4. The relationship $\frac{\partial u}{\partial x}\frac{\partial v}{\partial y} \neq \frac{\partial v}{\partial x}\frac{\partial u}{\partial y}$ if not, then $\sigma(z)$ is not a suitable AF except in the following cases:

$$\frac{\partial u}{\partial x} = \frac{\partial v}{\partial x} = 0, \text{and } \frac{\partial u}{\partial y} \neq 0, \frac{\partial v}{\partial y} \neq 0, \tag{97.1}$$

$$\frac{\partial u}{\partial y} = \frac{\partial v}{\partial y} = 0, \text{and } \frac{\partial u}{\partial x} \neq 0, \frac{\partial v}{\partial x} \neq 0. \tag{97.2}$$

Note that both sets of conditions above emphasize the boundedness of an AF and its partial derivatives, even when the function is defined in a local domain of interest. By Louiville theorem, however, the cost of this restriction is that a bounded AF cannot be analytic. The Tanh(z) function violates the second and third boundedness requirements of both sets of conditions above. Instead of boundedness, Tanh(z) function has well-defined but not necessarily bounded first order derivatives almost everywhere in $\mathbb{C}$. Since they are bounded almost everywhere, the rare existence of singular points hardly poses a problem in learning, and the singular points can be handled separately. Therefore, the above boundedness requirements are unnecessary for the fully complex AFs that are almost everywhere bounded [21].

Several complex AFs have been proposed in the literature. These AFs are mainly classified into the following two categories.

## 6.1 Split AF

The initial idea of split AF is proposed by [22, 41], where they adopt the real-valued AF (for example, Tanh and Sigmoid function) as shown in (98.1) and (98.2).

- Split real–imaginary AF

$$\sigma_{\text{Real-imaginary}}(z) = \sigma^{\Re}(\Re[z]) + i\sigma^{\Re}(\Im[z]). \tag{98.1}$$

- Split phase-amplitude AF

$$\sigma_{\text{Phase-amplitude}}(z) = \sigma^{\Re}(|z|)\exp(i\arg(z)). \tag{98.2}$$

Where $z = x + iy \in \mathbb{C}$, $\sigma^{\Re}(\cdot)$ is the real-valued AF (e.g., Sigmoid, Tanh), and $\sigma_{\text{Real-imaginary}}(z) \in \mathbb{C}$, $\sigma_{\text{Phase-amplitude}}(z) \in \mathbb{C}$. Such AF is bounded and not analytic. In contrast to RVNNs, where analyticity has been a cornerstone for rigorous investigations, CVNNs adopt a more pragmatic approach. The attention has shifted towards constructing the dynamics of learning and self-organization based on meaningful partial derivatives. This departure from the requirement of analyticity does not diminish the significance of the analyses but rather opens up avenues for more flexible and creative network design.

An interesting consequence of this shift in thinking is the emergence of coordinate-dependent neural dynamics. Unlike the general principle that mechanics, including NNs, should be coordinate-independent, CVNNs embrace a certain level of coordinate dependence. This choice is not a drawback but a deliberate strategy to enhance the CVNNs. Two widely used AFs exemplify this departure from traditional thinking – one that focuses on the real and imaginary components (98.1) and another that manipulates amplitude and phase (98.2). The coordinate dependence of CVNN dynamics becomes an asset when interacting with the real world. Considerations of amplitude, phase, and frequency become crucial in applications involving wave control or periodic motion. A CVNN is uniquely positioned to utilize this coordinate dependence as an advantage, aligning its dynamics with the specific properties it directly deals with in real-world interactions.

*6.2 Fully Complex AF*

Where real and imaginary parts are treated as a single entity. There are only a few works of literature on fully complex AF compared to split-type due to the computation complexity and the difficulty in fulfilling Livoullie's theorem.

$$\sigma_{\text{Fully Complex}}(z) = \sigma^{\mathbb{C}}(z), \tag{99}$$

where $z = x + iy \in \mathbb{C}$, $\sigma^{\mathbb{C}}(\cdot)$ denotes, for example, a complex Sigmoid. For instance, in [21, 24], ETFs (Tanh(z), Sinh(z), Sin(z), ArcSin(z), Tan(z), ArcTan(z), ArcTanh(z), ArcCos(z)) are studied where the functions are entire and either bounded in a desired domain or bounded almost everywhere (i.e., unbounded on a set of points at complex plane).

## 7. Complex AFs

AFs are crucial in NNs for determining how effectively a network learns and models data. In CVNNs, which handle both real and imaginary components, AFs must be tailored to manage this complexity. This section introduces three main types of complex-valued Afs. 1. Split Type: These functions apply separately to the real and imaginary parts of the complex number. Examples include the Split-Step, Split-Sigmoid, and Split-Tanh functions. 2. Amplitude-Phase Type: These functions modulate the amplitude of the input complex number while preserving its phase. Examples include the Amplitude-Phase-Type Function. 3. Fully Complex Type: These functions process the entire complex number simultaneously. Examples include the Fully Complex Exponential Function. Each type offers unique advantages for different applications in CVNNs, enhancing the network's capability to handle complex-valued data effectively.

*7.1 Split-Step Function*

The Split-Step function [42] is a CVAF defined by complex combining a real-valued step function applied separately to the real and imaginary components of a complex variable $z = x + iy$. The mathematical expression for the Split-Step Function is given by:

$$\begin{aligned}\sigma_{\text{Split-StepF}}(z) &= \sigma^{\mathfrak{R}}_{\text{Split-StepF}}(x) + i\sigma^{\mathfrak{R}}_{\text{Split-StepF}}(y)\\ &= \sigma^{\mathfrak{R}}_{\text{Split-StepF}}(\mathfrak{R}(z)) + i\sigma^{\mathfrak{R}}_{\text{Split-StepF}}(\mathfrak{I}(z)),\end{aligned} \tag{100.1}$$

where $z = x + iy$ and $\sigma^{\mathfrak{R}}_{\text{Split-StepF}}$ is a real-valued step function defined on $\mathbb{R}$; that is,

$$\sigma^{\mathfrak{R}}_{\text{Split-StepF}}(u) = \begin{cases} 1, & u \geq 0 \\ 0, & \text{otherwise}\end{cases}, \tag{100.2}$$

for any $u \in \mathbb{R}$.

The Split-Step function introduces a threshold activation behavior, as illustrated in Fig. 13. If the real part or the imaginary part of the complex variable is non-negative, the corresponding real or imaginary part of the AF becomes 1; otherwise, it becomes 0. This AF operates independently on the real and imaginary components of the complex variable, allowing it to capture different aspects of information. The output of each component of the Split-Step function is binary, either 0 or 1, making it particularly suitable for scenarios where a binary decision or representation is desired.

The XOR problem and symmetry detection problem require non-linear decision boundaries that a single real-valued neuron in a two-layered RVNN cannot provide. However, a single complex-valued neuron in a two-layered CVNN, utilizing the Split-Step function, can solve these problems by forming orthogonal decision boundaries in the complex plane. This demonstrates the potent computational power of complex-valued neurons, allowing them to efficiently handle complex, non-linear tasks that are challenging for real-valued neurons.

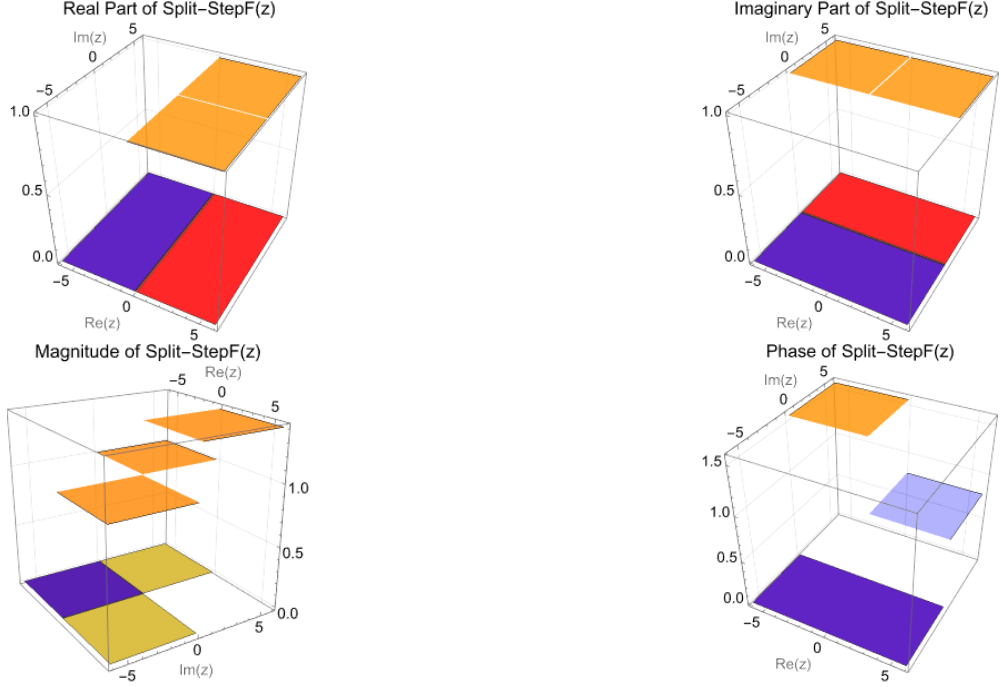

**Fig. 13.** Visualizations of the real, imaginary, magnitude, and phase parts of the Split-Step function. These figures provide a comprehensive visualization of the Split-Step function, $\sigma_{\text{Split-StepF}}(z)$, where $z = x + iy$, illustrating its real part, imaginary part, magnitude, and phase over a 3D grid with $x$ and $y$ ranging from $-6$ to $6$. Each plot uses blue and orange mesh shading for clarity and includes a slice contour plot at $z = 0$ to enhance understanding of the function's behavior at this specific plane.

### 7.2 Split-Sigmoid Function

The Split-Sigmoid function [43-53] is a CVAF defined by the expression:

$$\sigma_{\text{Split-SigmoidF}}(z) = \sigma^{\Re}_{\text{Split-SigmoidF}}(x) + i\sigma^{\Re}_{\text{Split-SigmoidF}}(y)$$
$$= \frac{1}{1 + e^{-\Re(z)}} + i\frac{1}{1 + e^{-\Im(z)}}, \quad (101.1)$$

where $z = x + iy$ and $\sigma^{\Re}_{\text{Split-SigmoidF}}$ is a real-valued function. This AF exhibits sigmoidal behavior on the real axis, as described by:

$$\sigma^{\Re}_{\text{Split-SigmoidF}}(u) = \frac{1}{1 + e^{-u}}, \quad (101.2)$$

for any $u \in \mathbb{R}$. Remarkably, this AF is a unique fusion of real-valued Sigmoid functions applied separately to the real and imaginary components of the complex variable $z$.

The real and imaginary parts of the Split-Sigmoid function are bounded for any complex number, a direct consequence of the properties of the Sigmoid function used in their definitions, as illustrated in Fig. 14. The Sigmoid function ensures that both the imaginary and real parts of the Split-Sigmoid CVAF remain bounded between 0 and 1. This characteristic is crucial in NNs as it contributes to stability during forward propagation, preventing exploding activations that can impede training.

The Split-Sigmoid CVAF exhibits line symmetry with respect to both the real and imaginary axes, Fig. 14. This symmetry implies that replacing a complex number $z_1 = (a, b)$ with $z_2 = (a, -b)$ leaves the real part unchanged and replacing ($z_1 = (a, b)$ with $z_2 = (-a, b)$ leaves the imaginary part unchanged. This simplifies the analysis of the function, enhancing its interpretability and reducing complexity.

The line symmetry exhibited by the Split-Sigmoid CVAF is a key feature with significant implications for neural dynamics. This symmetry simplifies analysis by establishing a relationship between values on one side of the real or imaginary axes and their mirrored counterparts. The Split-Sigmoid CVAF's characteristics make it particularly suitable for scenarios where NNs encounter information with inherent symmetries. The line symmetry along the real and imaginary axes implies that the function excels in processing complex information and displaying specific symmetrical patterns.

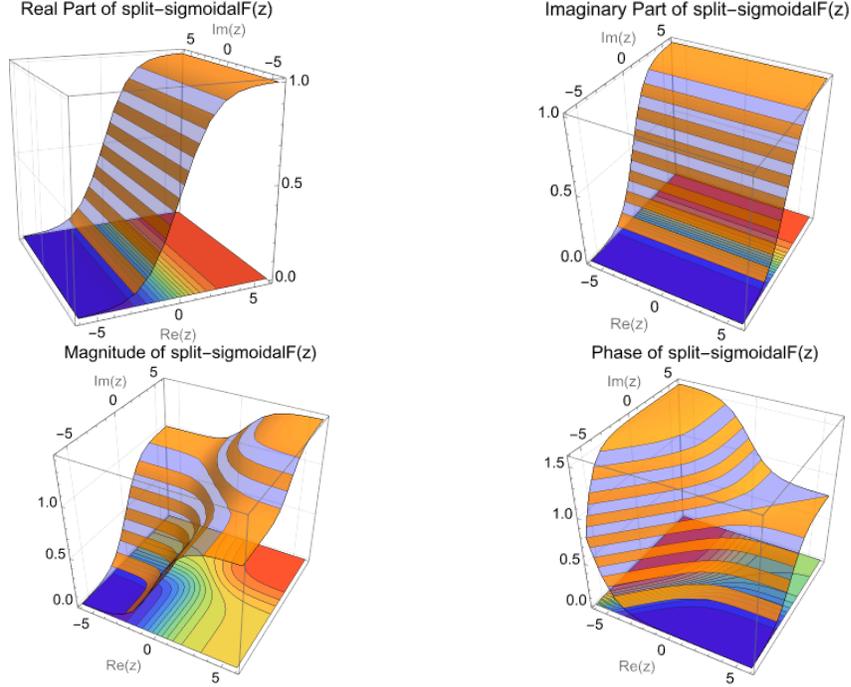

**Fig. 14.** Visualizations of the real, imaginary, magnitude, and phase parts of the Split-Sigmoid function.

The analogy drawn between the Split-Sigmoid CVAF and RVNNs processing $2n$-dimensional real-valued information provides valuable insights into the neural dynamics associated with this CVAF. The separate processing of real and imaginary components parallels the operations in RVNNs, offering a bridge for understanding and potentially leveraging existing knowledge from real-valued domains.

### 7.3 Split-Parametric Sigmoid Function

The Split-Parametric Sigmoid function was introduced as a CVAF [22, 23], expressed mathematically as:

$$\sigma_{\text{Split-PSigmoidF}}(z) = \sigma^{\Re}_{\text{Split-PSigmoidF}}(x) + i\sigma^{\Re}_{\text{Split-PSigmoidF}}(y) \\ = \frac{2c_1}{1 + e^{-c_2 \Re(z)}} - c_1 + i\left(\frac{2c_1}{1 + e^{-c_2 \Im(z)}} - c_1\right), \quad (102.1)$$

where $z = x + iy$ and $\sigma^{\Re}_{\text{Split-PSigmoidF}}$ is a real-valued function. It has sigmoidal behavior defined on $\mathbb{R}$; that is,

$$\sigma^{\Re}_{\text{Split-PSigmoidF}}(u) = \frac{2c_1}{1 + e^{-c_2 u}} - c_1, \quad (102.2)$$

for any $u \in \mathbb{R}$, with $c_1$, and $c_2$ suitable real parameters. The derivative of $\sigma^{\Re}_{\text{Split-PSigmoidF}}(u)$ has a simple expression:

$$\frac{\partial}{\partial u}\sigma^{\Re}_{\text{Split-PSigmoidF}}(u) = \frac{c_2}{2c_1}\left[c_1^2 - \left(\sigma^{\Re}_{\text{Split-PSigmoidF}}(u)\right)^2\right]. \quad (102.3)$$

A notable feature of the Split-PSigmoidF is its inherent boundedness, Fig. 15. The AF's behavior is a two-dimensional extension of the Sigmoid function applied to the real axis. The Sigmoid behavior ensures that the Split-PSigmoidF remains within certain limits, contributing to stability during NN training.

The parameters $c_1$ and $c_2$ give practitioners control over the amplitude, shape, sensitivity, and convergence properties of the Split-Parametric Sigmoidal function, allowing for fine-tuning and adaptation to specific NN requirements, Fig. 15. Adjusting these parameters may involve a balance between achieving convergence, avoiding saturation, and capturing intricate patterns in the data. $c_1$ is a real-valued parameter that scales the output of the sigmoidal function in the real- and imaginary- part of the Split-PSigmoidF. Increasing $c_1$ amplifies the output of the real- and imaginary- part, effectively increasing the amplitude of the Split-PSigmoidF. On the other hand, $c_2$ is a real-valued parameter that controls the rate of the exponential function in the sigmoidal part of the Split-PSigmoidF. Higher values of $c_2$ lead to a steeper sigmoidal curve, causing the AF to saturate more quickly. Lower values result in a more gradual saturation. Moreover, $c_2$ affects the sensitivity of the CVAF to changes in the input. A higher $c_2$ makes the function more sensitive, while a lower $c_2$ makes it less sensitive. The $c_2$ parameter directly appears in the derivative expression, influencing the rate of change of the AF.

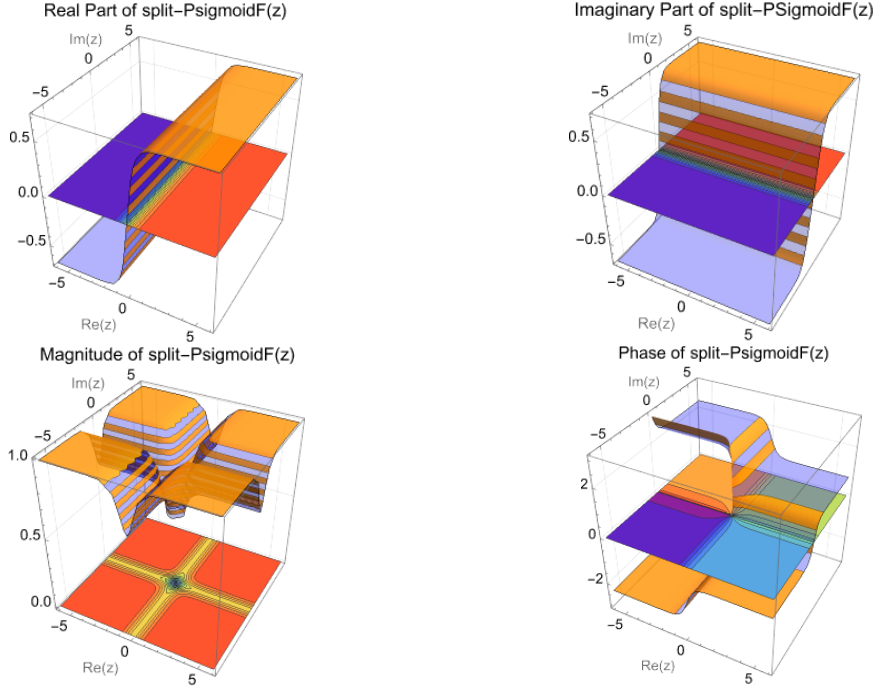

**Fig. 15.** Visualizations of the real, imaginary, magnitude, and phase parts of the Split-Parametric Sigmoid function.

### 7.4 Split-Tanh

One of the widely used CVAFs [54-57] is Split-Tanh. Split-Tanh CVAF is defined as

$$\sigma_{\text{Split-Tanh}}(z) = \sigma^{\Re}_{\text{Split-Tanh}}(x) + i\sigma^{\Re}_{\text{Split-Tanh}}(y) \\ = \text{Tanh}(\Re(z)) + i\,\text{Tanh}(\Im(z)), \tag{103.1}$$

where $z = x + iy$ and $\sigma^{\Re}_{\text{Split-Tanh}}$ is a real-valued function,

$$\sigma^{\Re}_{\text{Split-Tanh}}(u) = \text{Tanh}(u), \tag{103.2}$$

for any $u \in \mathbb{R}$. This CVAF is a unique fusion of real-valued hyperbolic tangent functions applied separately to the real and imaginary components of the complex variable.

The real- and imaginary- part of the Split-Tanh CVAF is bounded for any complex number, see Fig. 16. This boundedness is a result of the properties of the hyperbolic tangent function that is used in the definition of the Split-Tanh CVAF. The hyperbolic tangent function has a range between $-1$ and $1$. Specifically: $-1 < \text{Tanh}(u) < 1$. In the case of the real- and imaginary- part of the Split-Tanh AF, where $\text{Tanh}(\Re(z))$ and $\text{Tanh}(\Im(z))$ are involved, the real- and imaginary- parts are also bounded between $-1$ and $1$. This boundedness is important in the context of NNs because it helps in controlling the scale of values during the forward propagation of information through the network. It contributes to the stability of the network and prevents the activation values from growing too large, a phenomenon known as exploding activations, which can make training difficult.

The Split-Tanh CVAF exhibits line symmetry with respect to both the real and imaginary axes. The real- and imaginary- parts of Split-Tanh AF are visually depicted in Fig. 16, showcasing its characteristic shape. The symmetry of the real part of the split-Tanh function with respect to the real axis ($\Im(z) = 0$) implies that if you replace $z_1 = (a, b)$ with $z_2 = (a, -b)$ the real part of the split-Tanh function value remains the same. On the other hand, the symmetry of the imaginary part of the split-Tanh function with respect to the imaginary axis ($\Re(z) = 0$) implies that if you replace $z_1 = (a, b)$ with $z_2 = (-a, b)$ the imaginary part of the split-Tanh function value remains the same. This symmetry implies that the neural dynamics associated with this AF hold special significance along the axes of $\Im(z) = 0$ and $\Re(z) = 0$, i.e., the real and imaginary parts. Line symmetry simplifies the analysis of the function, making it easier to understand and work with. It reduces the complexity of studying the behavior of the function by

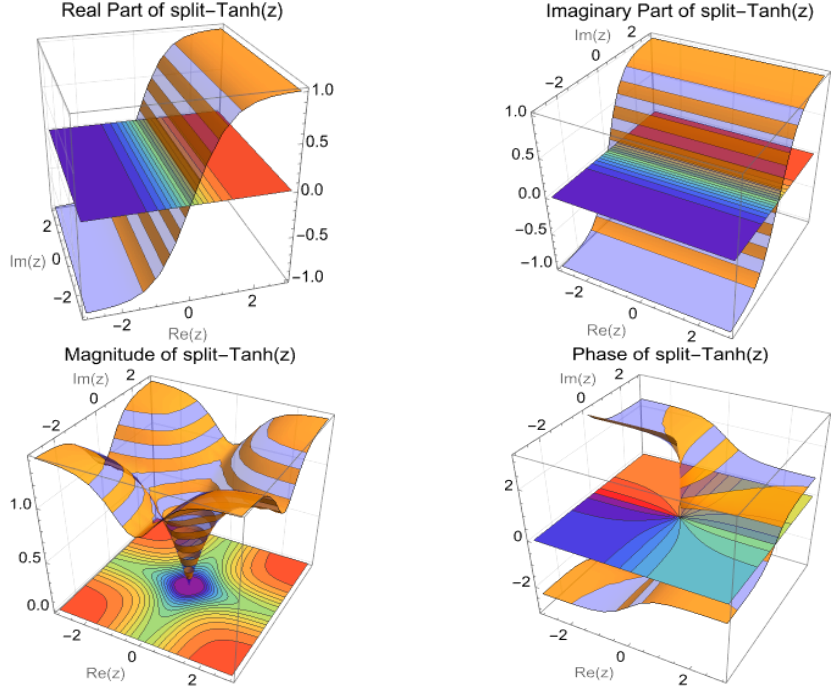
**Fig. 16.** Visualizations of the real, imaginary, magnitude, and phase parts of the Split-Tanh function.

providing a clear relationship between the values on one side of the real axis or imaginary axis and their mirrored counterparts on the other side. The line symmetry of the real-imaginary-type CVAF suggests that it excels when dealing with complex information that possesses symmetry or holds specific meanings along the real and imaginary axes. This symmetry implies that the neural dynamics of a network using this AF may resemble, to some extent, those of a RVNN processing $2n$-dimensional real-valued information. This analogy arises from the separate and independent processing of real and imaginary components, as demonstrated in (103.1) and (103.2).

We can also observe characteristic changes on and around the real and imaginary axes in figures of the amplitude and the phase of the Split-Tanh AF, Fig. 16. The symmetry gives the two axes a special meaning in neural dynamics.

### *7.5 Split- Sigmoid Tanh Function*

The Split- Sigmoidal Tanh (Split-STanh) CVAF [58] was defined as follows:

$$\sigma_{\text{Split-STanh}}(z) = \sigma^{\Re}_{\text{Split-STanh}}(x) + i\sigma^{\Re}_{\text{Split-STanh}}(y)$$
$$= \frac{\text{Tanh}(\Re(z))}{1 - (\Re(z) - 3)e^{-\Re(z)}} + i\frac{\text{Tanh}(\Im(z))}{1 - (\Im(z) - 3)e^{-\Im(z)}}, \quad (104.1)$$

where $z = x + iy$ and $\sigma^{\Re}_{\text{Split-STanh}}$ is a real-valued function,

$$\sigma^{\Re}_{\text{Split-STanh}}(u) = \frac{\text{Tanh}(u)}{1 - (u - 3)e^{-u}}, \quad (104.2)$$

for any $u \in \mathbb{R}$. The real- and imaginary- parts of Split-STanh AF, as well as the amplitude and the phase of the AF, are visually depicted in Fig. 17, showcasing its characteristic shape.

The Split-STanh CVAF is designed to ensure boundedness for both its real and imaginary parts. The maximum value of the real part of the Split-STanh AF is approximately 1.01802 and occurs when the input $x$ is around 4.06725, and the minimum value of the real part is approximately $-0.0715838$ and occurs when the input $x$ is around $-0.67288$. Similarly, the maximum and minimum values of the imaginary part of the Split-STanh AF approximately 1.01802 and $-0.0715838$ occur when the input $y$ is around 4.06725, and $-0.67288$, respectively.

The AF exhibits line symmetry with respect to both the real and imaginary axes, see Fig. 17.

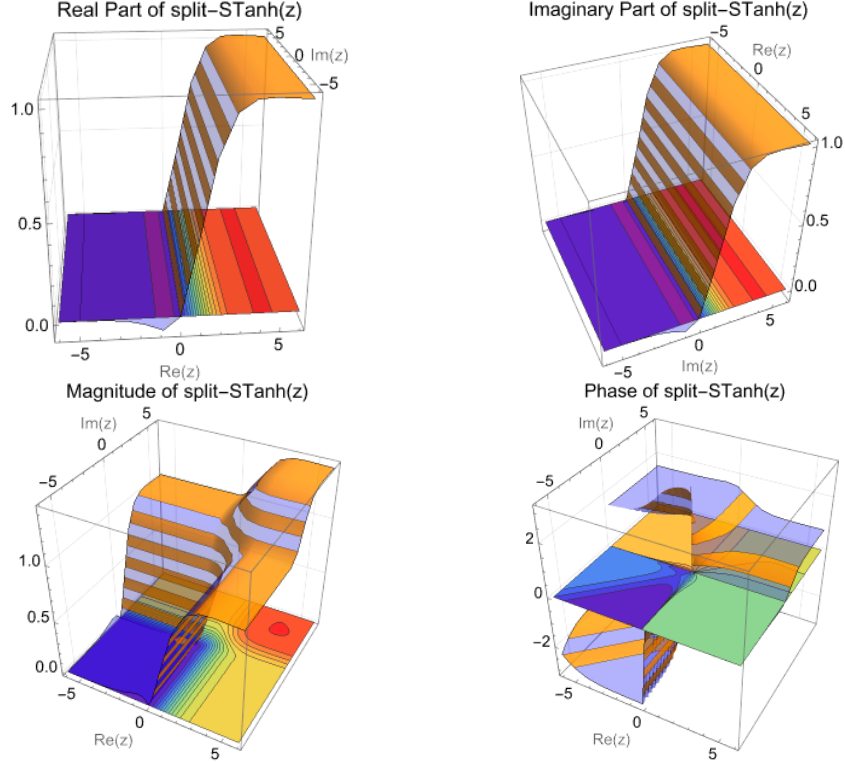

**Fig. 17.** Visualizations of the real, imaginary, magnitude, and phase parts of the Split-STanh function.

## 7.6 Split-Hard Tanh (Split-absolute value)

The Split-Hard Tanh [59] is a non-smooth function used in place of a Split-Tanh function. The Split-Hard Tanh retains the basic shape of Split-Tanh but uses simpler functions. Split-Hard Tanh AF is defined as

$$\sigma_{\text{Split-Hard Tanh}}(z) = \sigma^{\Re}_{\text{Split-Hard Tanh}}(x) + i\sigma^{\Re}_{\text{Split-Hard Tanh}}(y)$$
$$= \frac{1}{2}(|\Re(z) + 1| - |\Re(z) - 1|) + i\frac{1}{2}(|\Im(z) + 1| - |\Im(z) - 1|), \quad (105.1)$$

where $z = x + iy$ and $\sigma^{\Re}_{\text{Split-Hard Tanh}}$ is a real-valued function,

$$\sigma^{\Re}_{\text{Split-Hard Tanh}}(u) = \frac{1}{2}(|u + 1| - |u - 1|), \quad (105.2)$$

for any $u \in \mathbb{R}$. The function, given by $\sigma^{\Re}_{\text{Split-Hard Tanh}}(u) = \frac{1}{2}(|u + 1| - |u - 1|)$, is particularly interesting due to its connection with HardTanh AF. Let us explore its behavior across various ranges of real numbers. For $u > 1$, both $u + 1$ and $u - 1$ will be positive. Thus, the function simplifies to $\sigma^{\Re}_{\text{Split-Hard Tanh}}(u) = \frac{1}{2}(u + 1 - (u - 1)) = 1$. For $u < -1$, both $u + 1$ and $u - 1$ will be negative. In this case, the function becomes $\sigma^{\Re}_{\text{Split-Hard Tanh}}(u) = \frac{1}{2}(-(u + 1) - [-(u - 1)]) = -1$. For $0 < u < 1$, $u + 1$ will be positive and $u - 1$ will be negative. The function then simplifies to $\sigma^{\Re}_{\text{Split-Hard Tanh}}(u) = \frac{1}{2}((u + 1) - (-(u - 1))) = u$. For $-1 < u < 0$, $u + 1$ will be positive and $u - 1$ will be negative. Thus, the function is $\sigma^{\Re}_{\text{Split-Hard Tanh}}(u) = \frac{1}{2}((u + 1) - (-(u - 1))) = u$. Therefore, for both $-1 < u < 0$ and $0 < u < 1$, the function $\sigma^{\Re}_{\text{Split-Hard Tanh}}(u)$ evaluates to $u$. This specific behavior showcases a linear relationship between the function's output and the real variable $u$ in this range. Mathematically, the $\sigma^{\Re}_{\text{Split-Hard Tanh}}(u)$ function can be defined as follows:

$$\sigma^{\Re}_{\text{Split-Hard Tanh}}(u) = \begin{cases} -1, & u < -1 \\ 1, & u > 1. \\ u, & -1 \leq u \leq 1 \end{cases} \quad (106)$$

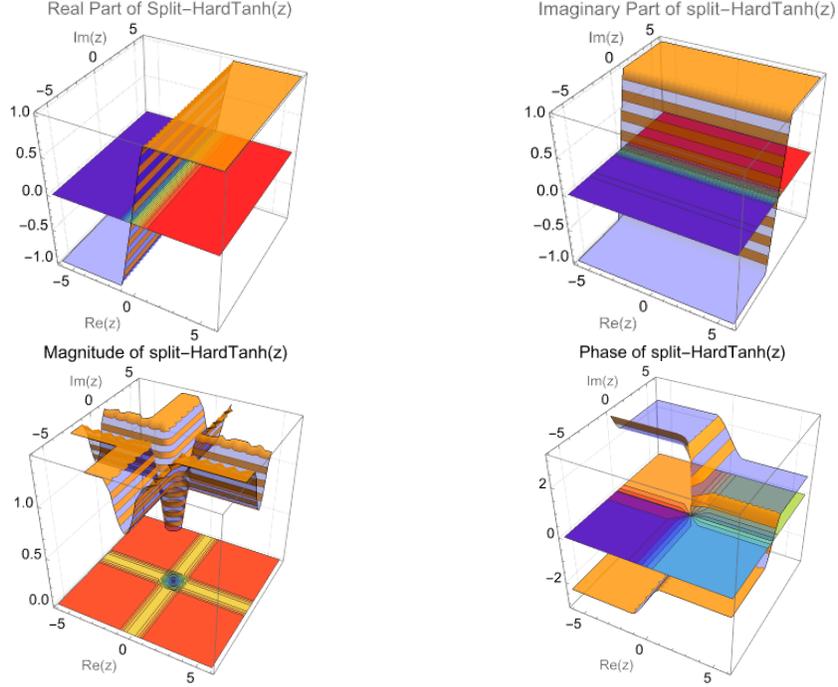

**Fig. 18.** Visualizations of the real, imaginary, magnitude, and phase parts of the Split-Hard Tanh function.

The HardTanh function is a modified version of the standard Tanh function that clips its output values to a certain range, typically $[-1, 1]$. Mathematically, the hard Tanh function can be defined as follows:

$$\sigma_{\text{Hard Tanh}}(u) = \begin{cases} -1, & u < -1 \\ 1, & u > 1 \\ u, & -1 \leq u \leq 1 \end{cases} \quad (107)$$

As a result, the $\sigma_{\text{Split–Hard Tanh}}(z)$ function is a modified version of the $\sigma_{\text{Split–Tanh}}(z)$ function, Fig. 18.

One of the main advantages of the Split-Hard Tanh function is its computational efficiency. Unlike the Split-Tanh function, which can be computationally expensive, the Split-Hard Tanh function relies on simple linear operations (addition and absolute value), making it faster to compute. This efficiency can be particularly beneficial in scenarios where computational resources are limited or when training large NNs.

### 7.7 Split-CReLU

The Split-CReLU [60, 61] AF is an extension of the ReLU AF for real numbers, but it is applied separately to the real and imaginary parts of a complex number. The general form of the Split-CReLU AF is given by:

$$\begin{aligned}\sigma_{\text{Split–CReLU}}(z) &= \sigma^{\Re}_{\text{Split–CReLU}}(x) + i\sigma^{\Re}_{\text{Split–CReLU}}(y) \\ &= \max(\Re(z), 0) + i\max(\Im(z), 0),\end{aligned} \quad (108.1)$$

where $z = x + iy$ and $\sigma^{\Re}_{\text{Split–CReLU}}$ is a real-valued function,

$$\sigma^{\Re}_{\text{Split–CReLU}}(u) = \text{ReLU}(u) = \max(u, 0), \quad (108.2)$$

for any $u \in \mathbb{R}$. The result of applying the Split-ReLU to a complex input is a complex number with potentially different real and imaginary components.

Like the traditional ReLU, the Split-ReLU induces sparsity in the representation. Any negative values in either the real or imaginary part are set to zero, effectively activating only positive components, see Fig. 19. This sparsity can be beneficial in certain scenarios, such as feature selection.

Split-CReLU satisfies the Cauchy-Riemann equations when both the real and imaginary parts are at the same time either strictly positive or strictly negative. This implies that Split-CReLU meets the Cauchy-Riemann criteria when $\theta_z \in (0, \pi/2)$ or $\theta_z \in (\pi, 3\pi/2)$.

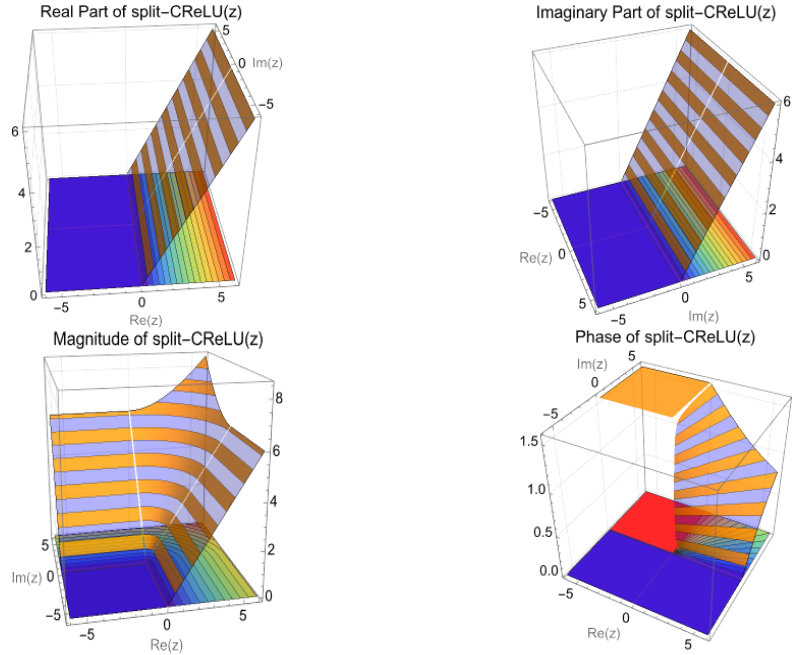

**Fig. 19.** Visualizations of the real, imaginary, magnitude, and phase parts of the Split-CReLU function.

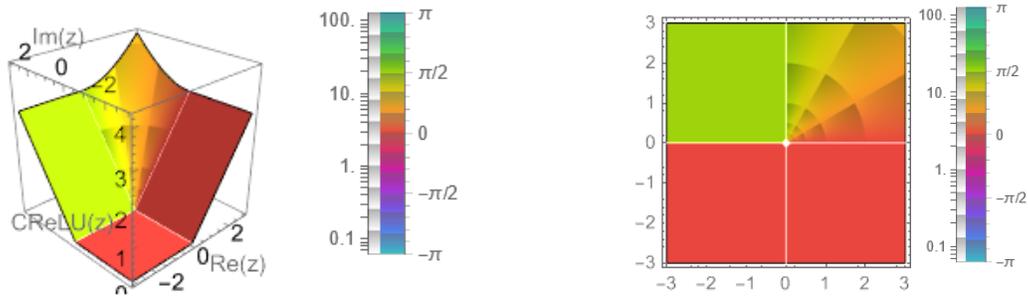

**Fig. 20.** Left panel: ComplexPlot3D generates a 3D plot of Abs[Split-CReLU] colored by arg[Split-CReLU] over the complex rectangle with corners $-3 - 3i$ and $3 + 3i$. Using "CyclicLogAbsArg" to cyclically shade colors to give the appearance of contours of constant Abs[Split-CReLU] and constant arg[Split-CReLU]. Right panel: ComplexPlot generates a plot of arg[Split-CReLU] over the complex rectangle with corners $-3 - 3i$ and $3 + 3i$. Using "CyclicLogAbsArg" to cyclically shade colors to give the appearance of contours of constant Abs[Split-CReLU] and constant arg[Split-CReLU].

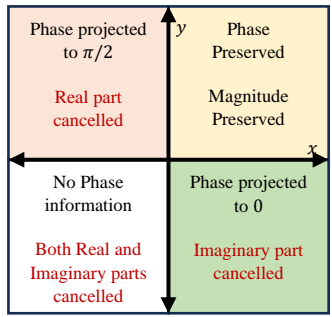

**Fig. 21.** Phase information encoding for Split-CReLU Function. The $x$-axis represents the real part and the $y$-axis axis represents the imaginary part. Split-CReLU discriminates the complex information into 4 regions where in two of which, phase information is projected to $0$ and $\pi/2$ and not canceled.

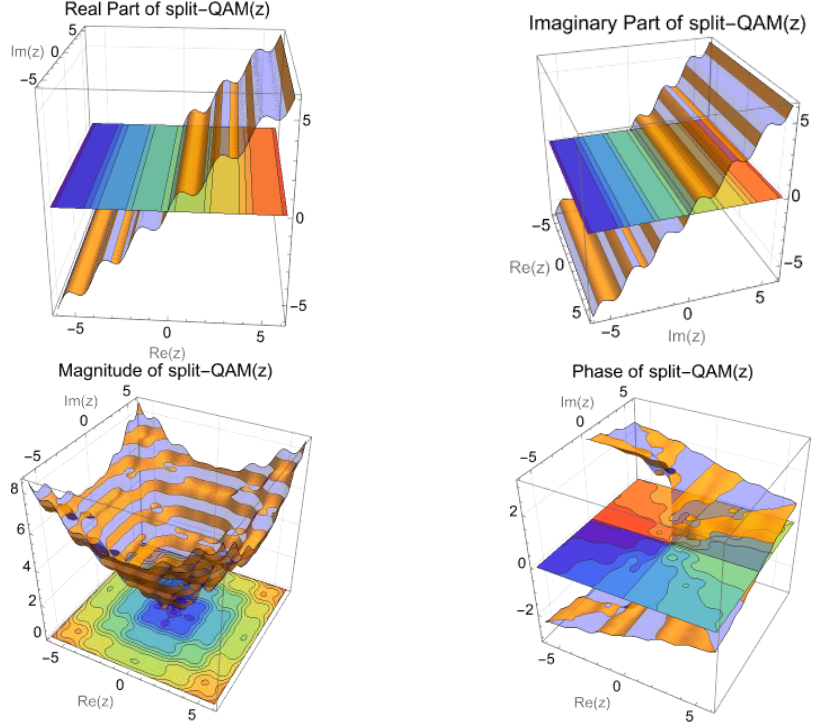

**Fig. 22.** Visualizations of the real, imaginary, magnitude, and phase parts of the Split-QAM function.

Split-CReLU discriminates the complex information into 4 regions where in two of which, phase information is projected and not canceled. This allows Split-CReLU to discriminate information easier with respect to phase information than the other AFs, see Figs. 20 and 21. CReLU has flexibility manipulating phase as it can either set it to zero or $\pi/2$, or even delete the phase information (when both real and imaginary parts are canceled) at a given level of depth in the network.

*7.8 Split-QAM Function*

Split-Quadrature Amplitude-Modulation (Split-QAM) [40] AF was defined by

$$\begin{aligned}\sigma_{\text{Split-QAM}}(z) &= \sigma_{\text{Split-QAM}}^{\Re}(x) + i\sigma_{\text{Split-QAM}}^{\Re}(y) \\ &= \Re(z) + \alpha \sin(\pi\, \Re(z)) + i[\Im(z) + \alpha \sin(\pi\, \Im(z))],\end{aligned} \quad (109.1)$$

where $z = x + iy$ and $\sigma_{\text{Split-QAM}}^{\Re}$ is a real-valued function,

$$\sigma_{\text{Split-QAM}}^{\Re}(u) = u + \alpha \sin(\pi u), \quad (109.2)$$

for any $u \in \mathbb{R}$ and the slope parameter $\alpha$ that determines the degree of nonlinearity is a positive real constant.

The Split-QAM CVAF was developed to handle Quadrature Amplitude Modulation (QAM) signals of any constellation size. The multi-saturation characteristic of the $S$-shape output exhibited by this function is particularly suited to $M$-ary QAM signals with discrete amplitudes, enhancing the network's robustness to noise due to the small $\Delta y$ for large $\Delta x$, as illustrated in Fig. 22. The Split-QAM AF modulates both the real and imaginary parts of the complex input using sinusoidal terms, with the parameter $\alpha$ controlling the strength of this modulation. This parameter requires careful tuning based on the specific task. This type of AF is beneficial for NNs in communication systems where modulation plays a crucial role in signal processing. However, depending on the application, the additional computational cost introduced by the sinusoidal terms may need to be considered.

*7.9 Amplitude-Phase-Type Function*

The Amplitude-Phase-Type function (APTF) [62-68] is expressed as

$$\sigma_{\text{APTF}}(z) = \text{Tanh}(|z|)\, e^{i \arg(z)}, \quad (110)$$

where $|z|$ denotes the magnitude (amplitude) and $\arg(z)$ represents the phase of the complex variable $z$.

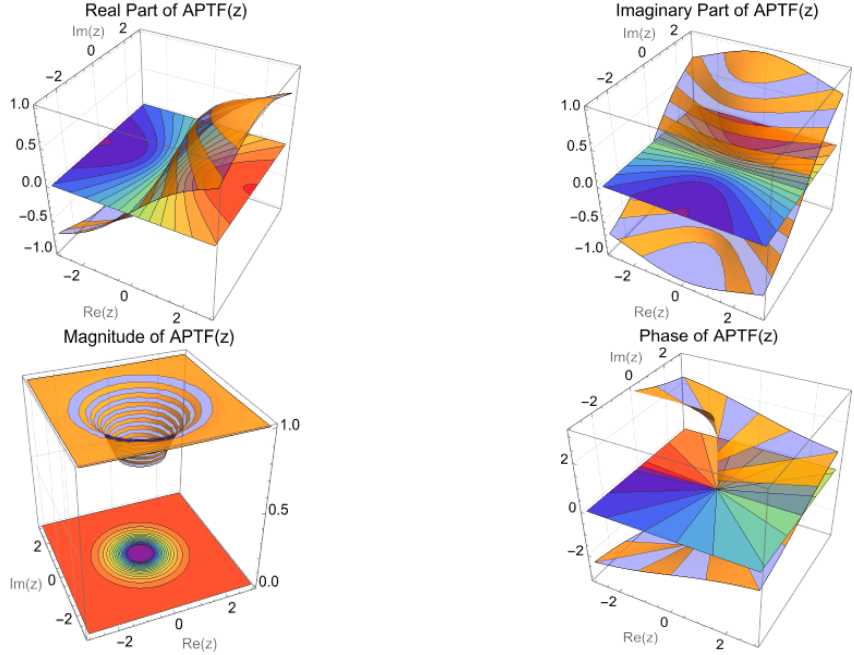

**Fig. 23.** Visualizations of the real, imaginary, magnitude, and phase parts of the Amplitude-phase-type function.

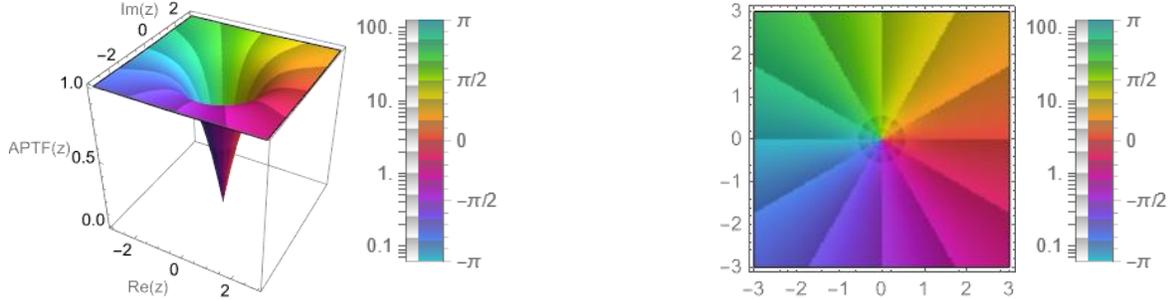

**Fig. 24.** Left panel: The ComplexPlot3D generates a 3D plot of Abs[$\sigma_{\text{APTF}}(z)$] colored by arg[$\sigma_{\text{FATF}}(z)$] over the complex rectangle with corners $z_{min} = -3 - 3i$ and $z_{max} = 3 + 3i$. Using "CyclicLogAbsArg" to cyclically shade colors to give the appearance of contours of constant Abs[$\sigma_{\text{APTF}}(z)$] and constant arg[$\sigma_{\text{APTF}}(z)$]. Right panel: ComplexPlot generates a plot of arg[$\sigma_{\text{APTF}}(z)$] over the complex rectangle with corners $-3 - 3i$ and $3 + 3i$. Using "CyclicLogAbsArg" to cyclically shade colors to give the appearance of contours of constant Abs[$\sigma_{\text{APTF}}(z)$] and constant arg[$\sigma_{\text{APTF}}(z)$].

Unlike the real-imaginary-type AF, this formulation emphasizes saturation in amplitude while keeping the phase unchanged. The term Tanh($|z|$) in the APTF emphasizes saturation in amplitude. The hyperbolic tangent function squashes the input values between $-1$ and $1$, promoting saturation when the amplitude of $z$ is high. The visual representation in Figs 23 and 24 illustrates the distinctive shape of the amplitude-phase-type AF, emphasizing its point symmetry about the origin $(0, i0)$, which is clearly observed in amplitude and phase figures.

The symmetry around the origin implies that the function's behavior remains consistent regardless of how the coordinate axes are oriented. In other words, if you rotate the entire coordinate system, the APTF will still exhibit the same shape and properties (amplitude). This property is advantageous in scenarios where information processing involves rotation around the origin, as the APTF remains unaffected by the angle of rotation, making it particularly suitable for tasks where the input data or features may undergo rotational transformations.

The amplitude-phase-type AF finds its niche in the processing of wave-related information. By associating the wave amplitude with the amplitude of the complex variable and the wave phase with the phase of the neural variable, this CVAF becomes a valuable tool for applications involving electromagnetic waves, light waves, sonic waves, ultrasonic waves, quantum waves, and other wave-related phenomena. Beyond its utility in wave processing, the amplitude-phase-type AF is well-suited for applications where point symmetry concerning the origin is essential.

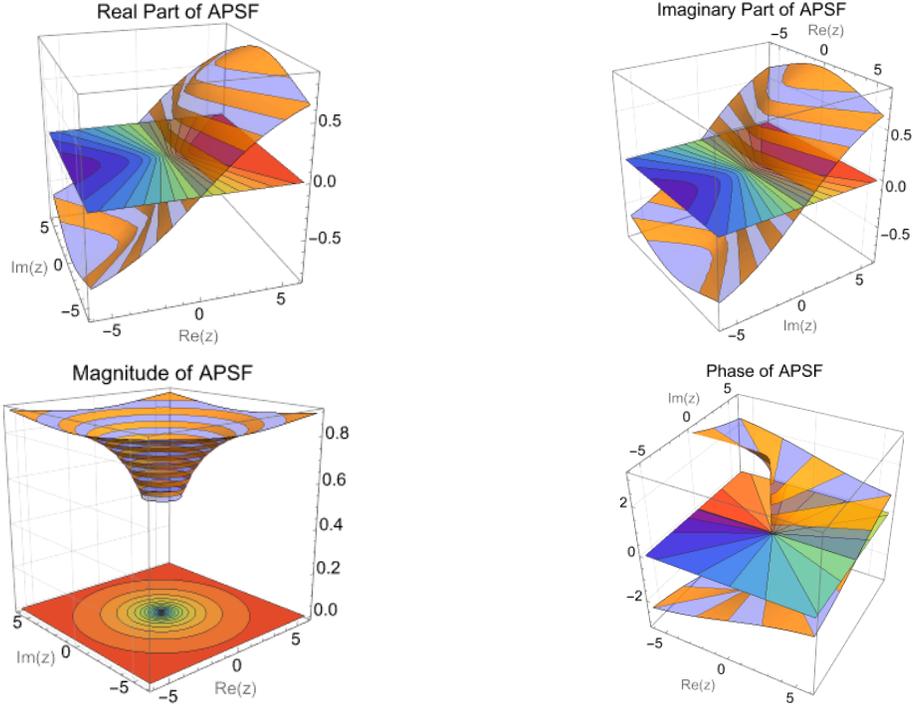

**Fig. 25.** Visualizations of the real, imaginary, magnitude, and phase parts of the Amplitude-Phase Sigmoidal function.

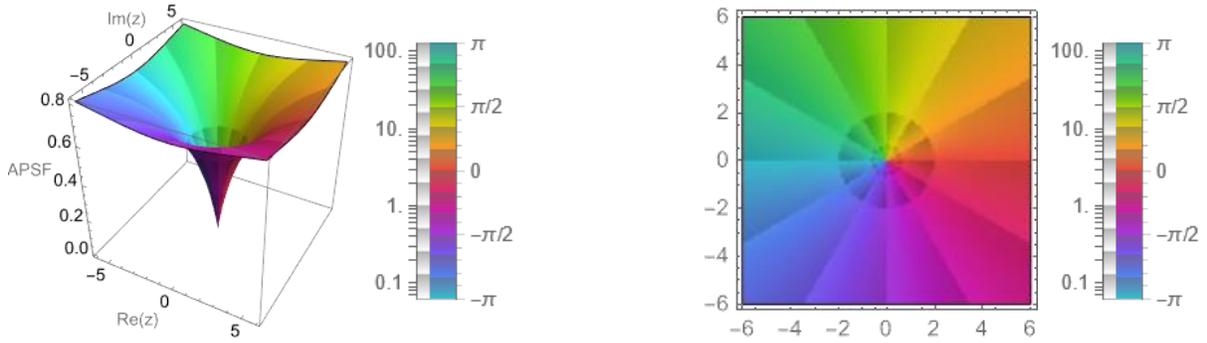

**Fig. 26.** Left panel: The ComplexPlot3D generates a 3D plot of Abs[$\sigma_{\text{APSF}}(z)$] colored by arg[$\sigma_{\text{APSF}}(z)$] over the complex rectangle with corners $z_{min} = -6 - 6i$ and $z_{max} = 6 + 6i$. Using "CyclicLogAbsArg" to cyclically shade colors to give the appearance of contours of constant Abs[$\sigma_{\text{APSF}}(z)$] and constant arg[$\sigma_{\text{APSF}}(z)$]. Right panel: ComplexPlot generates a plot of arg[$\sigma_{\text{APSF}}(z)$] over the complex rectangle with corners $-6 - 6i$ and $6 + 6i$. Using "CyclicLogAbsArg" to cyclically shade colors to give the appearance of contours of constant Abs[$\sigma_{\text{APSF}}(z)$] and constant arg[$\sigma_{\text{APSF}}(z)$].(with parameters; $a = 2$ and $b = 1$)

### *7.10 Amplitude- Phase Sigmoidal Function*

Another common class of non-analytic CVAFs is the Amplitude-Phase Sigmoidal Function (APSF) [20] popularized by

$$\sigma_{\text{APSF}}(z) = \frac{z}{a + \frac{1}{b}|z|} = \left(\frac{b}{ab + |z|}\right)z, \tag{111}$$

where $a$ and $b$ are real positive constants. This function has the property of mapping a point $z = x + iy = (x, y)$ on the complex plane to a unique point $\sigma_{\text{APSF}}(z) = (u = \frac{x}{a+|z|/b}, v = \frac{y}{a+|z|/b})$ on the open disc $\{z: |z| < b\}$. The division by $(ab + |z|)$ in the denominator ensures that the magnitude is scaled such that the output lies in the open disc $\{z: |z| < b\}$. This is because $ab + |z|$ is always greater than $|z|$ due to the positivity of both $a$ and $b$. The inequality $ab + |z| > |z|$ guarantees that $1 > \frac{|z|}{ab+|z|}$, which implies $b > \left(\frac{b}{ab+|z|}\right)|z| = |\sigma_{\text{APSF}}(z)|$. Therefore, the resulting magnitude is always scaled down to be less than $b$. The APSF is designed

to map complex numbers to points within a specific open disc, and its magnitude is scaled to ensure that the output always lies within that disc.

The APSF has the property of monotonically squashing the magnitude $|z|$ to a value $|\sigma_{\text{APSF}}(z)|$ within the interval $[0, b)$. This means that as the magnitude of $z$ increases, the output magnitude $|\sigma_{\text{APSF}}(z)|$ increases as well, but it is always constrained to be less than $b$. The Sigmoid function maps a real number $x$ to a point in the interval $(0,1)$. The hyperbolic tangent function maps a real number $x$ to a point in the interval $(-1,1)$. In a similar vein, the APSF maps complex numbers to a point with magnitude in the interval $[0, b)$, exhibiting a squashing behavior akin to the Sigmoid and hyperbolic tangent functions. So, the APSF can be seen as the natural generalization of real-valued squashing functions such as the Sigmoid. The visual representation in Figs. 25 and 26 illustrates the distinctive shape of the APSF, emphasizing its point symmetry about the origin $(0, i0)$, which is clearly observed in amplitude and phase figures.

The parameter $a$ in the APSF controls the steepness of the function. Specifically, it influences how quickly the function approaches its limiting value of $b$ as the magnitude of $z$ increases. For larger values of $a$, the function approaches its limiting value more gradually, resulting in a smoother transition. Conversely, for smaller values of $a$, the function approaches its limiting value more rapidly, leading to a steeper transition. The steepness parameter $a$ provides a way to adjust the sensitivity of the function to changes in the input, allowing for customization based on the specific characteristics desired for a given application or task in a NN or mathematical model. The term $z$ at the end of the APSF expression plays a crucial role in preserving the phase of the input. Multiplying $z$ by $\frac{b}{ab+|z|}$ scales the magnitude without changing the angle $\theta$, preserving the phase information.

The partial derivatives $\frac{\partial u}{\partial x}, \frac{\partial u}{\partial y}, \frac{\partial v}{\partial x}$ and $\frac{\partial v}{\partial y}$ are

$$\frac{\partial u}{\partial x} = \begin{cases} \frac{b(y^2 + ab|z|)}{|z|(ab + |z|)^2}, & |z| \neq 0 \\ \frac{1}{a}, & |z| = 0 \end{cases}, \tag{112.1}$$

$$\frac{\partial u}{\partial y} = \begin{cases} -\frac{bxy}{|z|(ab + |z|)^2}, & |z| \neq 0 \\ 0, & |z| = 0 \end{cases}, \tag{112.2}$$

$$\frac{\partial v}{\partial x} = \begin{cases} \frac{bxy}{|z|(ab + |z|)^2}, & |z| \neq 0 \\ 0, & |z| = 0 \end{cases}, \tag{112.3}$$

$$\frac{\partial v}{\partial y} = \begin{cases} \frac{b(x^2 + ab|z|)}{|z|(ab + |z|)^2}, & |z| \neq 0 \\ \frac{1}{a}, & |z| = 0 \end{cases}. \tag{112.4}$$

The special definitions of the partial derivatives when $|z| = 0$, i.e., at $z = (0,0)$, correspond to their limits as $z \to (0,0)$. Being thus defined, the singularities at the origin are removed and all partial derivatives exist and are continuous for all $z \in \mathbb{C}$.

### 7.11 Complex Cardioid

In their 1992 paper [20], Georgiou and Koutsougeras presented the activation, $\sigma_{\text{APSF}}(z)$, that attenuates the magnitude of the signal while preserving the phase. This CVAF was designed to be particularly useful when it is important to maintain the phase information of the input signal while adjusting the magnitude in a controlled manner. In [69], the authors refer to this CVAF as SigLog as it modifies the magnitude by applying the Sigmoid of the log of the magnitude:

$$\sigma_{\text{SigLog}}(z) = \frac{z}{1 + |z|} = \sigma_{\text{Sigmoid}}(\log(|z|)) e^{i \arg z}, \tag{113.1}$$

$$\sigma_{\text{Sigmoid}}(z) = \frac{1}{1 + e^{-z}}. \tag{113.2}$$

$\sigma_{\text{SigLog}}(z)$ is a special case from $\sigma_{\text{APSF}}(z)$ with $a = 1$ and $b = 1$. $|z|$ represents the magnitude of the complex number $z$, which is calculated as the distance of $z$ from the origin in the complex plane. The division by $(1 + |z|)$ ensures that the magnitude is scaled such that the output lies in the unit circle. This is because the denominator $(1 + |z|)$ is always greater than $|z|$, so the resulting magnitude is always scaled down to be less than 1. When $|z|$ is small, the attenuation is minimal, and when $|z|$ is large, the

attenuation becomes more significant. This implies that the magnitude of the complex number is attenuated in a manner that depends on its original magnitude.

The combination of the logarithmic function and the Sigmoid activation serves to control the magnitude attenuation. Let us prove the equality, $\frac{z}{1+|z|} = \sigma_{\text{sigmoid}}(\log(|z|))e^{i \arg z}$,

$$
\begin{aligned}
\sigma_{\text{sigmoid}}(\log(|z|))e^{i \arg z} &= \frac{1}{1 + e^{-\log(|z|)}} e^{i \arg z} \\
&= \frac{1}{1 + \frac{1}{|z|}} e^{i \arg z} \\
&= \frac{|z|}{|z| + 1} e^{i \arg z} \\
&= \frac{z}{|z| + 1}.
\end{aligned}
\tag{114}
$$

Complex Cardioid AF [69] is CVAF sensitive to the input phase rather than the input magnitude. The Complex Cardioid is defined as:

$$
\sigma_{\text{CCardioid}}(z) = \frac{1}{2}(1 + \cos[\arg z])z, \tag{115.1}
$$

$$
\frac{\partial}{\partial z}\sigma_{\text{CCardioid}}(z) = \frac{1}{2} + \frac{1}{2}\cos[\arg z] + \frac{i}{4}\sin[\arg z]. \tag{115.2}
$$

The function $\cos[\arg z]$ introduces phase sensitivity. The cosine function oscillates between $-1$ and $1$, meaning it varies based on the angle of $z$. This phase sensitivity is a key feature of Complex Cardioid activation. The factor $\frac{1}{2}(1 + \cos[\arg z])$ is responsible for amplitude modulation. When $\cos[\arg z] = 1$, the factor becomes 1, resulting in no attenuation of the magnitude. When $\cos[\arg z] = -1$, the factor becomes 0, leading to complete attenuation. The modulation is controlled by the cosine term, which depends on the phase of the input complex number. The $z$ term at the end ensures that the output phase remains the same as the input phase. Multiplying $z$ does not alter the phase but contributes to scaling the final output. The overall effect is that the output magnitude is attenuated based on the input phase ($\arg(z)$), while the output phase remains equal to the input phase.

The modulation of magnitude based on phase sensitivity might offer benefits in certain types of data where the relationship between magnitude and phase is significant. This CVAF could be useful in NNs where preserving phase information is crucial, such as in signal processing tasks or applications involving waveforms.

The real- and imaginary- part of $\sigma_{\text{CCardioid}}(z)$ AF, as well as the amplitude and the phase of the AF are visually depicted in Figs. 27 and 28 showcasing its characteristic shape. Input values that lie on the positive real axis are scaled by one. This means there is no attenuation; the magnitude remains unchanged. For the positive real axis, $z$ is a real number with zero imaginary part, i.e., $z = x$ where $x > 0$. The argument $\arg(z)$ for a positive real number is zero because it lies directly along the positive real axis. This means that $\cos[\arg z] = \cos(0) = 1$. The Complex Cardioid AF preserves the magnitude of input values on the positive real axis, resulting in an output that is equal to the input for these specific values.

Input values that lie on the negative real axis are scaled by zero. This implies complete attenuation. For the negative real axis, $z$ is a real number with zero imaginary part, i.e., $z = -x$ where $x > 0$. The argument $\arg(z)$ for a negative real number is $\pi$ because it lies directly along the negative real axis. This means that $\cos[\arg z] = \cos(\pi) = -1$. The amplitude modulation factor is 0 for the negative real axis. The Complex Cardioid AF effectively "turns off" or attenuates the magnitude for input values on the negative real axis, resulting in an output of zero for these specific values.

For input values with nonzero imaginary components, the scaling factor varies gradually from one to zero. As the complex number rotates in phase from the positive real axis towards the negative real axis, the cosine term ($\cos[\arg(z)]$) changes accordingly, leading to a gradual reduction in the scaling factor.

When the input values are restricted to real values, the Complex Cardioid function is simply the ReLU AF. When the input values are restricted to real numbers (zero imaginary component), the argument $\arg(z)$ becomes 0 for positive real numbers and $\pi$ for negative real numbers. $\sigma_{\text{cardioid}}(z) = z, z > 0$ and $\sigma_{\text{cardioid}}(z) = 0, z < 0$. Hence, Complex Cardioid AF is a phase sensitive complex extension of ReLU.

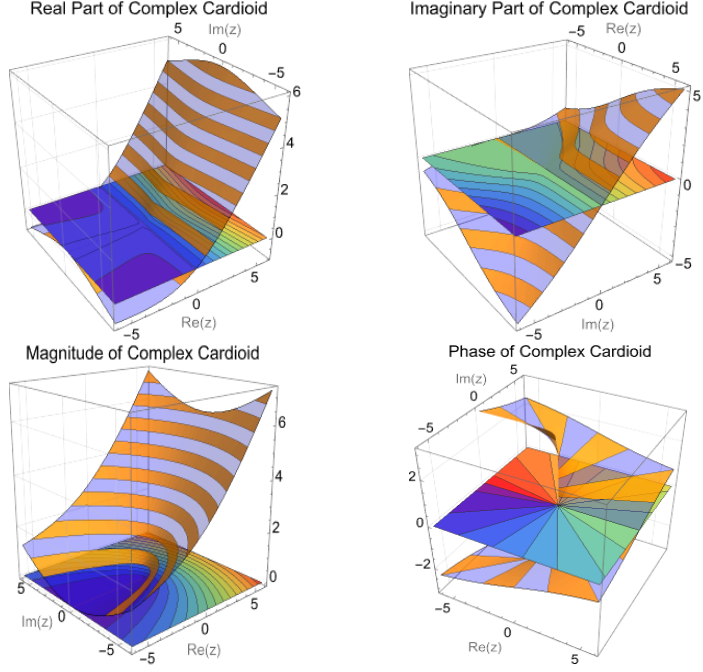

**Fig. 27.** Real part, imaginary part, amplitude, and phase of the $\sigma_{\text{CCardioid}}(z)$ AF.

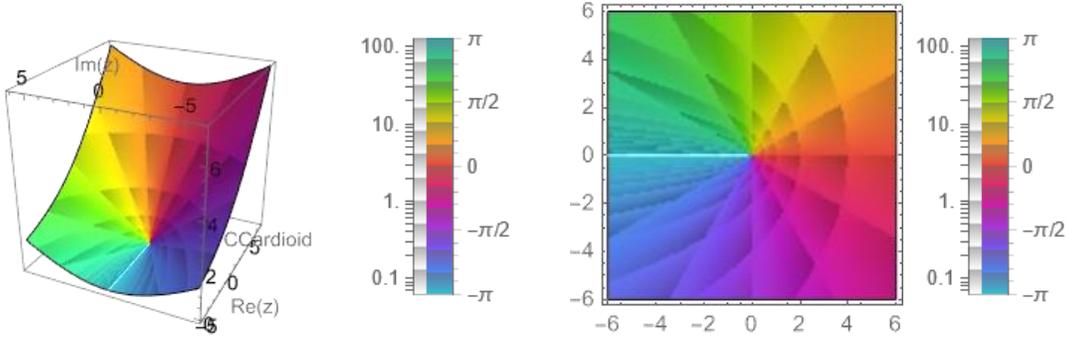

**Fig. 28.** Left panel: The ComplexPlot3D generates a 3D plot of Abs[$\sigma_{\text{CCardioid}}(z)$] colored by arg[$\sigma_{\text{CCardioid}}(z)$] over the complex rectangle with corners $z_{min} = -6 - 6i$ and $z_{max} = 6 + 6i$. Using "CyclicLogAbsArg" to cyclically shade colors to give the appearance of contours of constant Abs[$\sigma_{\text{CCardioid}}(z)$] and constant arg[$\sigma_{\text{CCardioid}}(z)$]. Right panel: ComplexPlot generates a plot of arg[$\sigma_{\text{CCardioid}}(z)$] over the complex rectangle with corners $-6 - 6i$ and $6 + 6i$. Using "CyclicLogAbsArg" to cyclically shade colors to give the appearance of contours of constant Abs[$\sigma_{\text{CCardioid}}(z)$] and constant arg[$\sigma_{\text{CCardioid}}(z)$].

### *7.12 modReLU*

modReLU [60, 61, 70-72] is a pointwise nonlinearity, $\sigma_{\text{modReLU}}(z): \mathbb{C} \to \mathbb{C}$, which affects only the absolute value of a complex number, defined as

$$\sigma_{\text{modReLU}}(z) = \begin{cases} \left(\dfrac{|z| + b}{|z|}\right) z, & |z| + b \geq 0 \\ 0, & |z| + b < 0 \end{cases}, \quad (116.1)$$

where $b \in \mathbb{R}$ is a bias parameter of the nonlinearity. Note that the modReLU is similar to the ReLU in spirit, in fact more concretely,

$$\sigma_{\text{modReLU}}(z) = \sigma_{\text{ReLU}}(|z| + b)\frac{z}{|z|} = \max(|z| + b, 0)\frac{z}{|z|} = \sigma_{\text{ReLU}}(|z| + b)e^{i\theta}. \quad (116.2)$$

The $z$ term at the end of $\left(\frac{|z|+b}{|z|}\right)z$ is responsible for preserving the phase of the complex number. Since $z$ is multiplied back, it doesn't alter the phase of the complex number. The term $\frac{|z|+b}{|z|}$ is responsible for amplitude modulation. This part scales the magnitude of the complex number while introducing a bias $b$.

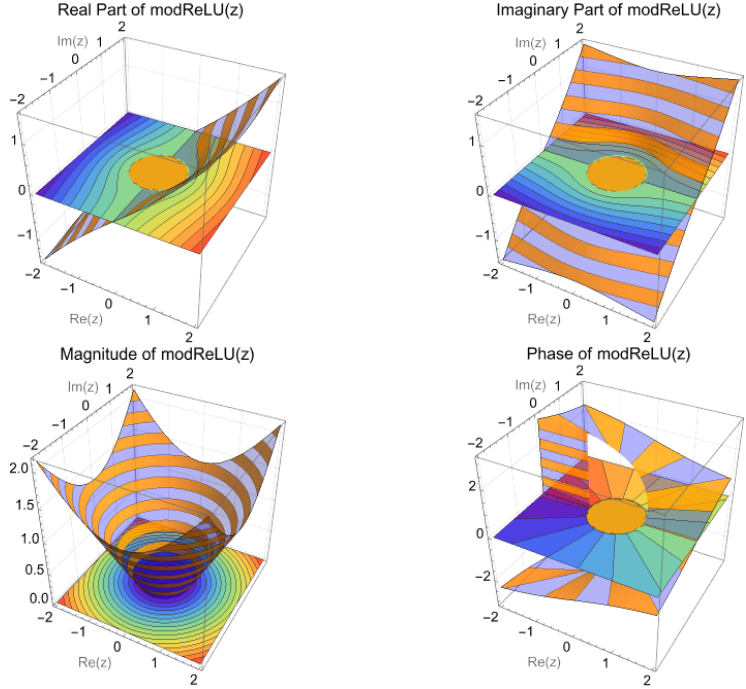

**Fig. 29.** Real part, imaginary part, amplitude, and phase of the $\sigma_{\text{modReLU}}(z)$ AF. ($b = -0.7$)

The ReLU operation $\sigma_{\text{ReLU}}(|z| + b) = \max(|z| + b, 0)$ ensures that the function becomes zero for negative inputs ($|z| + b < 0$) and retains the amplitude-modulated $z$ for non-negative inputs ($|z| + b \geq 0$). As $|z|$ is always positive, a bias $b$ is introduced in order to create a "dead zone" of radius $b$ around the origin 0 where the neuron is inactive, and outside of which it is active. Note that, modReLU does not satisfy the Cauchy-Riemann equations and thus is not holomorphic.

Case $b < 0$, see Fig. 29: In this case, the conditions are as follows: (1) $|z| + b \geq 0$: This condition is satisfied when $|z| \geq -b$. (2) $|z| + b < 0$: This condition is satisfied when $|z| < -b$. Therefore, for $b < 0$, the modReLU function is:

$$\sigma_{\text{modReLU}}(z) = \begin{cases} \left(\dfrac{|z| + b}{|z|}\right) z, & |z| \geq -b \\ 0, & |z| < -b \end{cases}. \tag{117}$$

This means that when $b < 0$, the modReLU function scales the complex number $z$ by a factor determined by the modified magnitude only when the original magnitude $|z|$ is greater than $-b$. Otherwise, the output is 0.

Case $b > 0$: In this case, the condition $|z| + b \geq 0$ is always satisfied, as $|z|$ is always non-negative, and adding a positive value $b$ to it results in a non-negative quantity. Therefore, the modReLU function simplifies to: $\sigma_{\text{modReLU}}(z) = \bigl((|z| + b)/|z|\bigr) z$. This means that when $b > 0$, the modReLU function scales the complex number $z$ by a factor determined by the modified magnitude ($|z| + b/|z|$). The "dead zone" has disappeared. Fig. 30 represents the phase information encoding for the modReLU AF. Fig. 31 compares the amplitudes of modReLU in two cases $b < 0$ and $b > 0$.

*7.13 Fully Complex Tanh Function*

The hyperbolic tangent function, Tanh (z), is easily defined as the ratio between the hyperbolic sine and the cosine functions (or expanded, as the ratio of the half-difference and half-sum of two exponential functions in the points $z$ and $-z$). The formula for the hyperbolic tangent of a complex number $z$ is given by:

$$\sigma_{\text{FCTanh}}(z) = \text{Tanh}(z) = \frac{e^z - e^{-z}}{e^z + e^{-z}} = \frac{\sinh z}{\cosh z}, \tag{118.1}$$

$$\frac{\partial}{\partial z} \sigma_{\text{FCTanh}}(z) = \text{sech}^2(z). \tag{118.2}$$

Complex hyperbolic tangent function was proposed as a "Fully Complex" AF [21, 23].

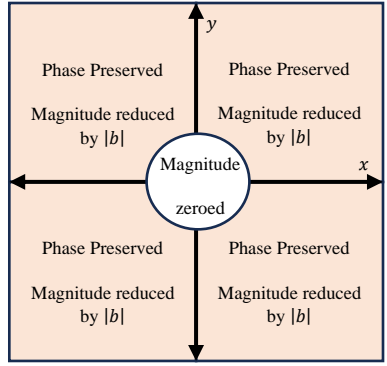

**Fig. 30.** Phase information encoding for the modReLU AF. The $x$-axis represents the real part and the $y$-axis axis represents the imaginary part. The case where $b < 0$ for modReLU. The radius of the white circle is equal to $|b|$. In the case where $b \geq 0$, the whole complex plane would be preserving both phase and magnitude information and the whole plane would have been colored with orange. We can see the modReLU, the complex representation is discriminated into two regions, i.e., the one that preserves the whole complex information (colored in orange) and the one that cancels it (colored in white).

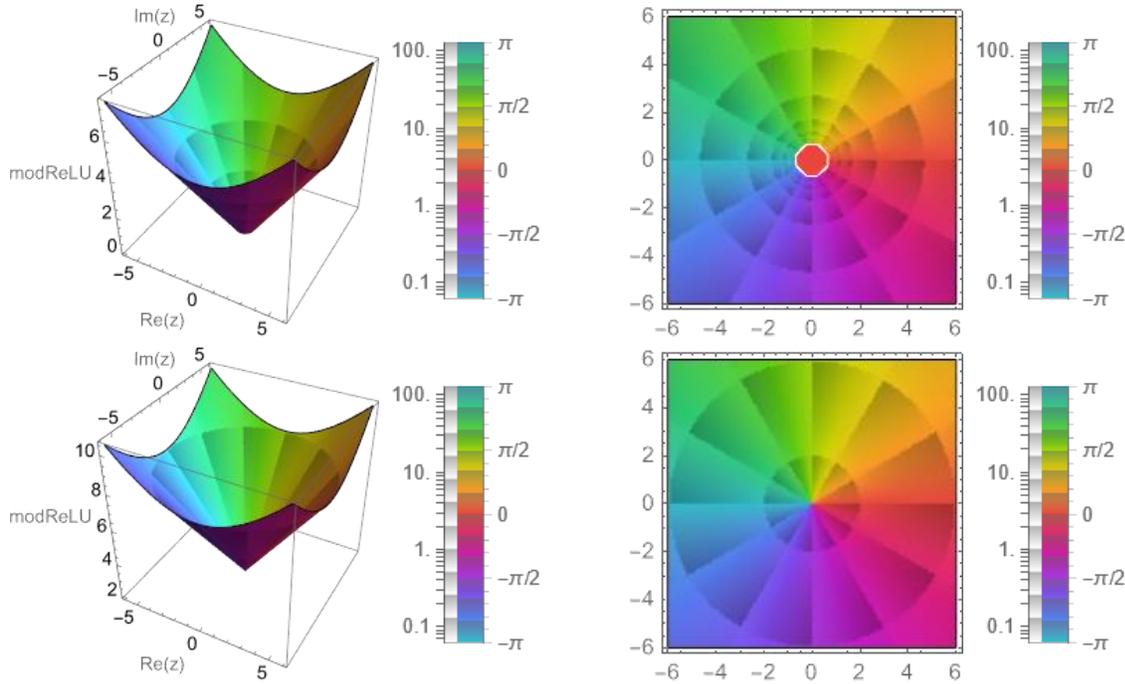

**Fig. 31.** Upper panel left: The ComplexPlot3D generates a 3D plot of $\text{Abs}[\sigma_{\text{modReLU}}(z)]$ with $b = -0.7$ colored by $\arg[\sigma_{\text{modReLU}}(z)]$ over the complex rectangle with corners $z_{min} = -6 - 6i$ and $z_{max} = 6 + 6i$. Using "CyclicLogAbsArg" to cyclically shade colors to give the appearance of contours of constant $\text{Abs}[\sigma_{\text{modReLU}}(z)]$ and constant $\arg[\sigma_{\text{modReLU}}(z)]$. Upper panel right: ComplexPlot generates a plot of $\arg[\sigma_{\text{modReLU}}(z)]$ with $b = -0.7$ over the complex rectangle with corners $-6 - 6i$ and $6 + 6i$. Using "CyclicLogAbsArg" to cyclically shade colors to give the appearance of contours of constant $\text{Abs}[\sigma_{\text{modReLU}}(z)]$ and constant $\arg[\sigma_{\text{modReLU}}(z)]$. Note that, the bias $b$ is introduced in order to create a "dead zone" of radius $|b|$ around the origin 0 where the neuron is inactive, and outside of which it is active. Lower panel: The 3D and 2D plots show the $\text{Abs}[\sigma_{\text{modReLU}}(z)]$ with $b = 2$. Note that, the "dead zone" has disappeared.

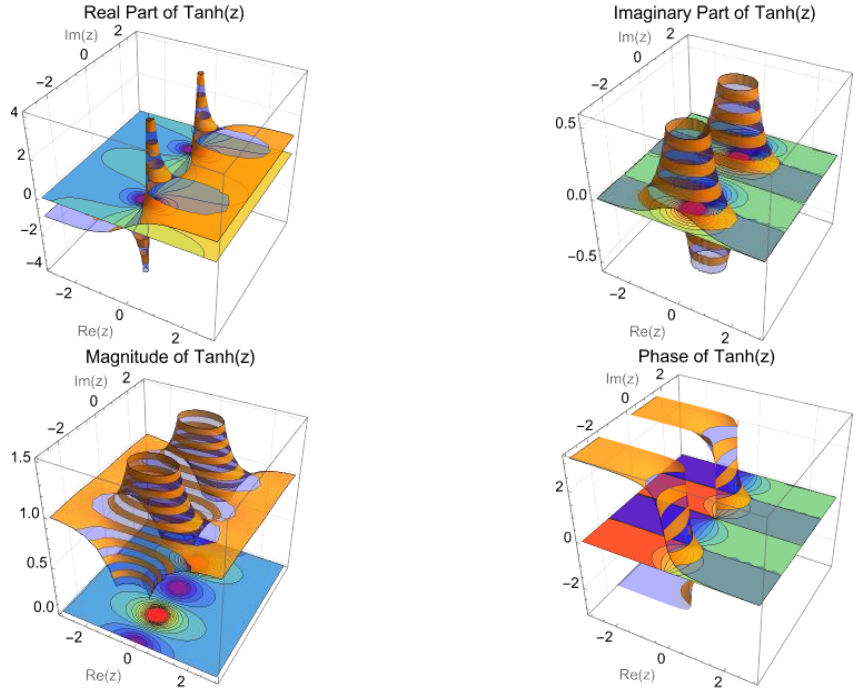

**Fig. 32.** Complex $\text{Tanh}(z)$ as a function of complex variables, real-part, imaginary-part, amplitude, and phase.

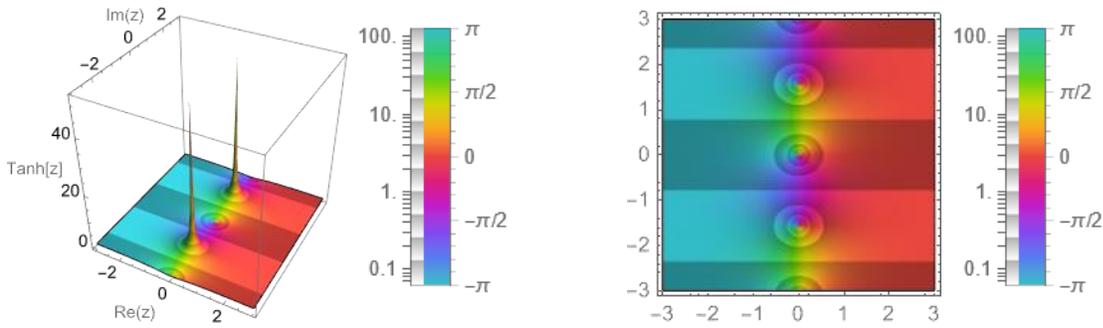

**Fig. 33.** Left panel: The ComplexPlot3D generates a 3D plot of $\text{Abs}[\sigma_{\text{FCTanh}}(z)]$ colored by $\arg[\sigma_{\text{FCTanh}}(z)]$ over the complex rectangle with corners $z_{min} = -3 - 3i$ and $z_{max} = 3 + 3i$. Using "CyclicLogAbs" to cyclically shade colors to give the appearance of contours of constant $\text{Abs}[\sigma_{\text{FCTanh}}(z)]$. Right panel: ComplexPlot generates a plot of $\arg[\sigma_{\text{FCTanh}}(z)]$ over the complex rectangle with corners $-3 - 3i$ and $3 + 3i$.

Note that singular points do exist at $z = (n + 1/2)\pi i$, $\forall n \in \mathbb{N}$, where $\cosh z$ has zeros, the denominator of the last formula equals zero. Figs 32 and 33 illustrate the shape of the function, which significantly deviates from the "saturation" feature. While this function is differentiable at almost every point, it diverges to infinity, deviating from the expected saturation behavior.

Instead of boundedness, $\text{Tanh}(z)$ function has well-defined but not necessarily bounded first order derivatives almost everywhere in $\mathbb{C}$. Since they are bounded almost everywhere, the rare existence of singular points hardly poses a problem in learning, and the singular points can be handled separately. Since $\text{Tanh}(z)$ is analytic and bounded almost everywhere in the complex plane, when trained by backpropagation, it can easily outperform the non-analytic split complex AF in convergence speed and achievable minimum squared error when the domain is bounded around the unit circle.

The non-saturating behavior of CVAFs poses a fundamental challenge to the development of CVNNs. This issue was the most serious reason that CVNNs were considered difficult to develop before. This observation challenges the intuition derived from real-valued Tanh functions and underscores the need for careful consideration when extending AFs into the complex domain. Understanding and resolving this issue are crucial for unlocking the full potential of CVNNs and leveraging their advantages in various applications.

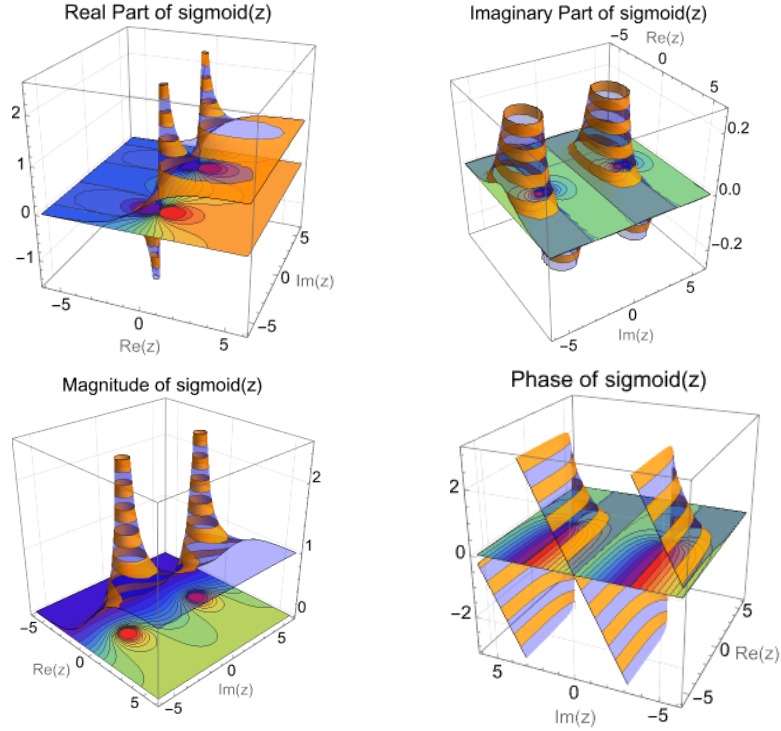

**Fig. 34.** Fully Complex Logistic-Sigmoidal Function as a function of a complex variable, real part, imaginary part, amplitude, and phase.

*7.14 Fully Complex Logistic-Sigmoidal Function*

The traditional Logistic Sigmoid function is commonly used in NNs to introduce non-linearity and map real-valued inputs to a range between 0 and 1. The fully complex Logistic Sigmoid AF [19] extends this concept to complex numbers, allowing for the incorporation of complex-valued inputs and weights in NNs. The function is defined as:

$$\sigma_{\text{FCSigmoid}}(z) = \frac{1}{1 + e^{-z}}, \tag{119}$$

here, $z$ is a complex number. Fig 34 shows $\sigma_{\text{FCSigmoid}}(z)$ in complex domain.

The use of the exponential function $e^{-z}$ in the $\sigma_{\text{FCSigmoid}}(z)$ involves potential issues with singularities when $z$ takes certain values. The denominator $1 + e^{-z}$ becomes zero when $e^{-z} = -1$. Solving for $z$, we find that this occurs when: $z = (2n + 1)i\pi$, where $n$ is an integer. To mitigate the singularity problem, practitioners often resort to scaling the input data to a manageable region in the complex plane. This scaling can be achieved by normalizing or restricting the input values to a certain range. By doing so, you effectively limit the range of possible values for the real and imaginary parts of the complex input $z$. This scaling helps in maintaining numerical stability during training and avoids situations where the exponential function produces extreme values. It is a common practice in CVNN implementations to preprocess the input data to ensure that it falls within a reasonable range.

*7.15 Fully Complex ETFs*

A single-valued function is said to have a singularity at a point if the function is not analytic at the point [13, 14]. The behavior of a complex function near a singular point can be classified into three main types: removable singularities, poles, and essential singularities. Here are some key concepts and details related to classification of singularity points:

- An isolated singularity is a type of singularity for a complex function that is surrounded by a neighborhood where the function is analytic. In other words, an isolated singularity is a point at which the function is not analytic, but there exists a deleted neighborhood (a punctured disk or an open set excluding the singularity itself) where the function is analytic. Mathematically, let $U \in \mathbb{C}$ be an open set and $z_0 \in U$. Suppose that $f: U \setminus \{z_0\} \to \mathbb{C}$ is holomorphic. In this situation, we say that $f$ has an isolated singular point (or isolated singularity) at $z_0$. The implication of the phrase is usually just that $f$ is defined and holomorphic on some such "deleted neighborhood" of $z_0$.

- A removable singularity is a type of singularity that can be "removed" or "filled in" by redefining the function at that point in such a way that the function becomes analytic at that point. In other words, a function has a removable singularity at a point if the singularity can be eliminated by assigning a value to the function at that point or by modifying the function in a small neighborhood of that point. Mathematically, if $f(z)$ has a removable singularity at $z_0$, it means that there exists a continuous function $g(z)$ such that $g(z) = f(z)$ for all $z$ in some deleted neighborhood of $z_0$, except possibly at $z_0$ itself. The function $g(z)$ is then analytic at $z_0$.
- Graphically, if you were to plot a function with a removable singularity, it might appear to have a "hole" or a point where the function is not defined. However, by redefining the function at that point or making a modification, you can smooth out the graph, and the function becomes well-behaved and analytic in the entire neighborhood.
- Example: Consider the function $f(z) = \frac{\sin(z)}{z}$, which has a singularity at $z = 0$. This singularity is removable because we can define a new function $g(z)$ as:

$$g(z) = \begin{cases} \frac{\sin(z)}{z}, & z \neq 0 \\ 1, & z = 0 \end{cases}, \quad (120)$$

 (by defining $f(0)$ to be 1 which is the limit of $f(z)$ as $z$ tends to 0). The function $g(z)$ is defined and continuous at $z = 0$ and is equal to $f(z)$ in its domain. Thus, the singularity at $z = 0$ is removable.
 Note that, if $f(z)$ has an isolated singularity at $z_0$, the singularity is said to be removable if $\lim_{z \to z_0} f(z)$ exists.

- Riemann Removable Singularities Theorem [13]: Let $f: U \setminus \{z_0\} \to \mathbb{C}$ be holomorphic and bounded. Then $\lim_{z \to z_0} f(z)$ exists. The function $g: U \to \mathbb{C}$ defined by

$$g(z) = \begin{cases} f(z), & z \neq 0, \\ \lim_{t \to z_0} f(t), & z = 0, \end{cases} \quad (121)$$

 is holomorphic.

- A pole is a type of singularity that occurs when a function approaches infinity at a certain point. If $\lim_{z \to z_0} f(z) \to \infty$, while $f(z)$ is analytic in a deleted neighborhood of $z = z_0$, then $f(z)$ is not removable but has a pole at $z = z_0$. Poles are characterized by the fact that the function becomes unbounded as the variable approaches a specific complex number. Equivalently, $f$ has a pole of order $n$ at $z_0$ if $n$ is the smallest positive integer for which $(z - z_0)^n f(z) = g(z)$ is holomorphic at $z_0$.
 Example: The basic example of a pole is $f(z) = 1/z^n$, which has a single pole of order $n$ at $z = 0$.

- An isolated singularity that is neither removable nor a pole is said to be an isolated essential singularity.
- This classification helps us understand the local behavior of a complex function near a singular point. Removable singularities are "fixable," poles represent a certain kind of divergence, and essential singularities indicate more complex and intricate behaviors.
- Furthermore, A branch point [14] is a point in the complex plane where a multi-valued function ceases to be single-valued. In other words, a branch point of a function is a point in the complex plane whose complex argument can be mapped from a single point in the domain to multiple points in the range. Suppose that we are given the function $w = z^{1/2}$. Suppose further that we allow $z$ to make a complete circuit (counter-clockwise) around the origin starting from point $A$. We have $z = re^{i\theta}$, $w = \sqrt{r}e^{i\theta/2}$ so that at $A$, $\theta = \theta_1$ and $w = \sqrt{r}e^{i\theta_1/2}$. After a complete circuit back to $A$, $\theta = \theta_1 + 2\pi$ and $w = \sqrt{r}e^{i(\theta_1+2\pi)/2} = -\sqrt{r}e^{i\theta_1/2}$. Thus, we have not achieved the same value of $w$ with which we started. However, by making a second complete circuit back to $A$, i.e., $\theta = \theta_1 + 4\pi$, $w = \sqrt{r}e^{i(\theta_1+4\pi)/2} = \sqrt{r}e^{i\theta_1/2}$ and we then do obtain the same value of $w$ with which we started. We can describe the above by stating that if $0 \leq \theta < 2\pi$, we are on one branch of the multiple-valued function $z^{1/2}$, while if $2\pi \leq \theta < 4\pi$, we are on the other branch of the function. It is clear that each branch of the function is single-valued. In order to keep the function single-valued, we set up an artificial barrier such as $OB$ where $B$ is at infinity (although any other line from $O$ can be used), which we agree not to cross. This barrier is called a branch line or branch cut. Hence, a branch cut is a curve or a line in the complex plane along which a multi-valued function is modified to make it single-valued. By introducing a branch cut, we exclude certain paths from the domain of the function, allowing us to define a single-valued function.

A number of ETFs derivable from the entire exponential function $e^z$ that are analytic are defined as "Fully Complex" AFs [23, 73, 74] and are shown to provide a parsimonious structure for processing data in the complex domain, and address most of the shortcomings of the traditional approach. The following ETFs are identified to provide adequate non-linearity as an AF.

**Circular functions**

$$\tan(z) = \sigma_{\text{FCtan}}(z) = \frac{e^{iz} - e^{-iz}}{i(e^{iz} + e^{-iz})}, \tag{122}$$

$$\sin(z) = \sigma_{\text{FCsin}}(z) = \frac{e^{iz} - e^{-iz}}{2i}. \tag{123}$$

**Inverse circular functions**

$$\arctan(z) = \sigma_{\text{FCarctan}}(z) = \int_0^z \frac{dt}{1+t^2}, \tag{124}$$

$$\arcsin(z) = \sigma_{\text{FCarcsin}}(z) = \int_0^z \frac{dt}{(1-t)^{\frac{1}{2}}}, \tag{125}$$

$$\arccos(z) = \sigma_{\text{FCarccos}}(z) = \int_z^1 \frac{dt}{(1-t^2)^{\frac{1}{2}}}. \tag{126}$$

**Hyperbolic functions**

$$\text{Tanh}(z) = \sigma_{\text{FCTanh}}(z) = \frac{\sinh(z)}{\cosh(z)} = \frac{e^z - e^{-z}}{e^z + e^{-z}}, \quad \sinh(z) = \sigma_{\text{FCsinh}}(z) = \frac{e^z - e^{-z}}{2}. \tag{127}$$

**Inverse hyperbolic functions**

$$\text{arcTanh}(z) = \sigma_{\text{FCarcTanh}}(z) = \int_0^z \frac{dt}{1-t^2}, \quad \text{arcsinh}(z) = \sigma_{\text{FCarcsinh}}(z) = \int_0^z \frac{dt}{(1+t^2)^{\frac{1}{2}}}. \tag{128}$$

Figs 35 through 43 show the magnitude of these ETFs in the vicinity of unit circle. These figures show two distinctive patterns of inevitable unboundedness property when real-valued ETFs are generalized into complex-valued ETFs. The ETFs divide into two classes, based on their types of singularity: a set of bounded squashing AFs over the bounded domain and a set of AFs with unbounded isolated singularities.

First category of unboundedness is observable in the tangent function family (Tan(z), Tanh(z), ArcTan(z), and ArcTanh(z)), Figs 35 through 38, that possesses finite number of point set discontinuities in a bounded domain. Note that ArcTan(z) has isolated singularity at $\pm i$ and ArcTanh(z) has isolated singularity at $\pm 1$. Moreover, Tanh(z) has isolated essential singularities at every $(1/2 + n)\pi i$, $n \in \mathbb{N}$, since it is asymptotically $+\infty$ as $(1/2 + n)\pi i$ is approached from below and to $-\infty$ as $(1/2 + n)\pi i$ is approached from above along the imaginary axis. Similarly, Tan(z) has isolated essential singularities at every $(1/2 + n)\pi$. Usually, these singular points and discontinuities at non-zero points do not pose a problem in training when the domain of interest is bounded within a circle of radius $\pi/2$. If the domain is larger including these irregular points, the training process tends to become more sensitive to the size of the learning rate. In this case, the initial random weights need to be bounded in a small radius, typically below 0.1 to avoid oscillation in the gradient-based updates.

Second category of unboundedness is observable in the inverse sine and cosine family including ArcSin(z), ArcCos(z), and ArcSinh(z) functions that exhibit unbounded but decreasing rate of magnitude growth as they move away from the origin, as shown in Figs 39 through 41. The inverse sine and cosines have removable singularities represented as ridges along the real or imaginary axis outside the unit circle, which is defined as the branch cuts, as illustrated in Figs 39 through 41. This branch-type discontinuity results from the definition of the integral where the branch should not be crossed. However, ArcSin(z) and ArcSinh(z) functions are the most radially symmetric functions in magnitude. ArcCos(z) function is asymmetric along the real axis and has a discontinuous zero point at $z = 1$. When the domain is bounded, ArcSin(z), ArcCos(z), and ArcSinh(z) are naturally bounded with squashing function characteristics while providing more discriminating nonlinearity than the split-Tanh function. It has been observed that the radial symmetricity as well as the nonlinear phase response of ArcSinh(z) and ArcSin(z) functions tend to provide efficient nonlinear approximation capability.

Sin(z) and Sinh(z) functions exhibit continuous magnitude but are unbounded functions with convex behavior along the imaginary and real axes, respectively. Figs 42 and 43 represent Sin(z) and Sinh(z) functions, respectively. Also, note that Sin(z) is equivalent to Sin(x) along the real axis, while Sinh(z) is equivalent to Sin(x) along the imaginary axis. Therefore, the bounded squashing property for Sin(z) and Sinh(z) functions is available within a radius of $\pi/2$ from the origin. Therefore, these functions, when bounded in a radius of π/2 can be effective nonlinear approximators. Consequently, Sin(z) and Sinh(z), ArcSin(z), ArcCos(z), and ArcSinh(z) functions (the sine family) can be classified as complex squashing functions within a bounded domain.

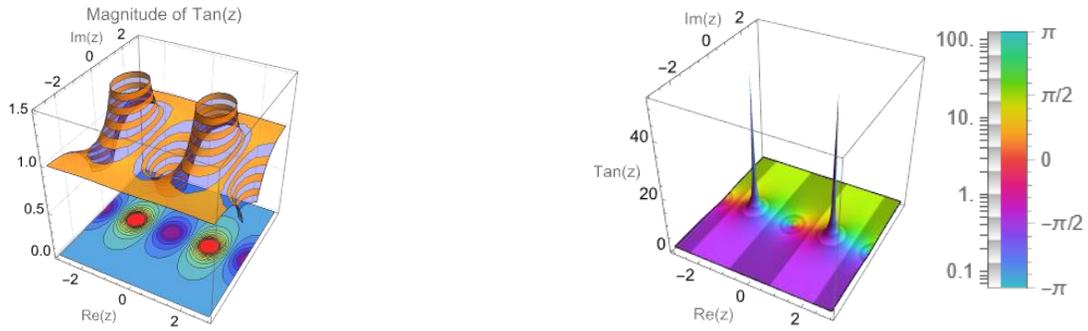

**Fig. 35.** Left panel: 3D and contour plot of the magnitude of $\sigma_{FCtan}(z)$. Right panel: The ComplexPlot3D generates a 3D plot of Abs$[\sigma_{FCtan}(z)]$ colored by arg$[\sigma_{FCtan}(z)]$ over the complex rectangle with corners $z_{min} = -3 - 3i$ and $z_{max} = 3 + 3i$. Additionally, the color shading is achieved using the "CyclicLogAbs" function to cyclically shade colors and give the appearance of contours of constant Abs$[\sigma_{FCtan}(z)]$.

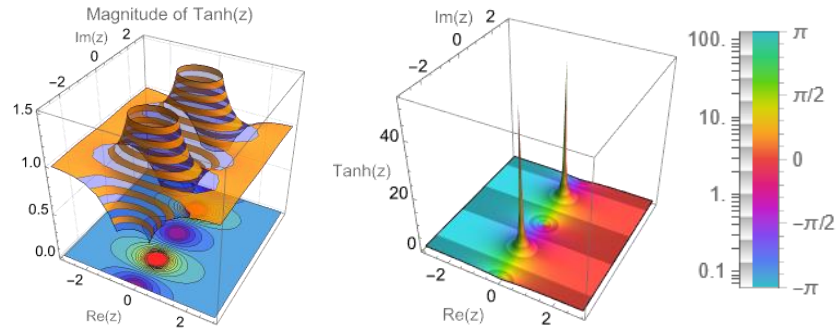

**Fig. 36.** Same as Fig. 35 but for the function $\sigma_{FCTanh}(z)$.

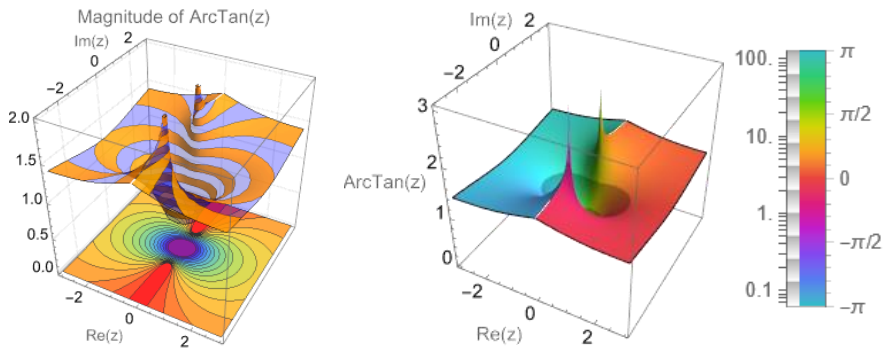

**Fig. 37.** Same as Fig. 35 but for the function $\sigma_{FCArcTan}(z)$.

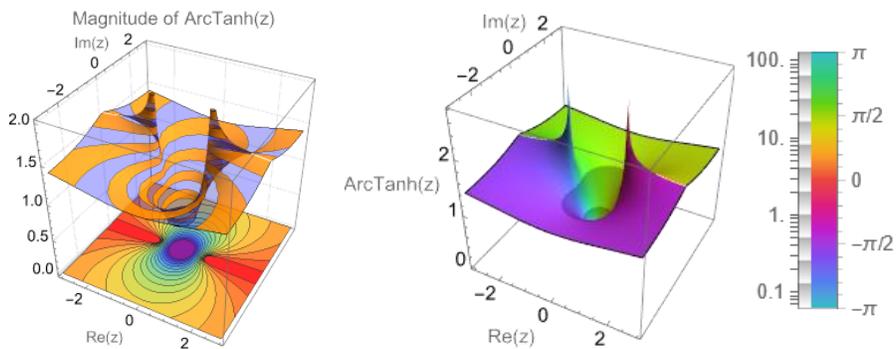

**Fig. 38.** Same as Fig. 35 but for the function $\sigma_{FCArcTanh}(z)$.

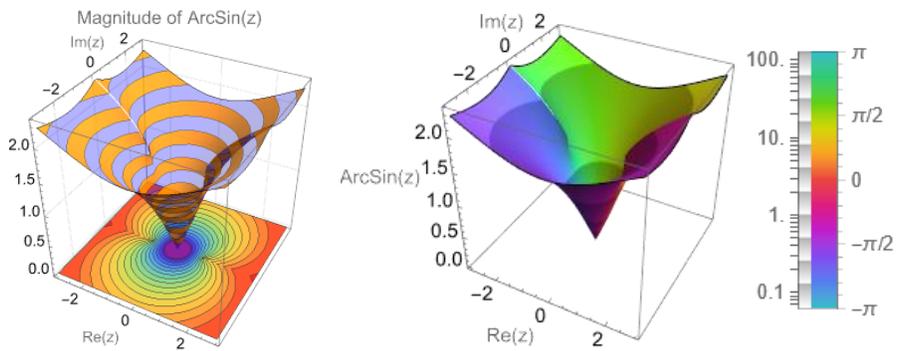

**Fig. 39.** Same as Fig. 35 but for the function $\sigma_{\text{FCArcSin}}(z)$.

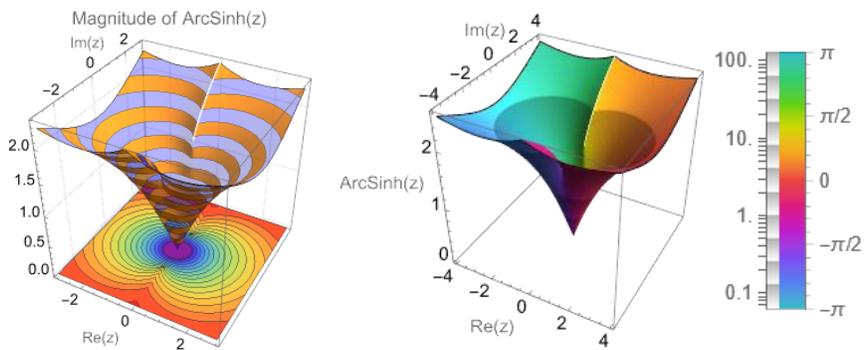

**Fig. 40.** Same as Fig. 35 but for the function $\sigma_{\text{FCArcSinh}}(z)$.

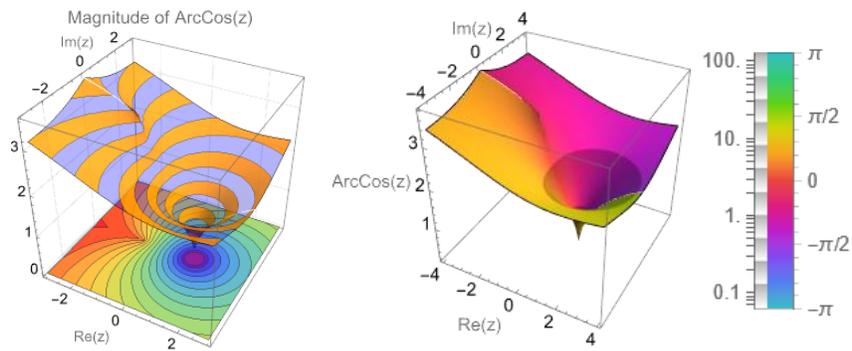

**Fig. 41.** Same as Fig. 35 but for the function $\sigma_{\text{FCArcCos}}(z)$.

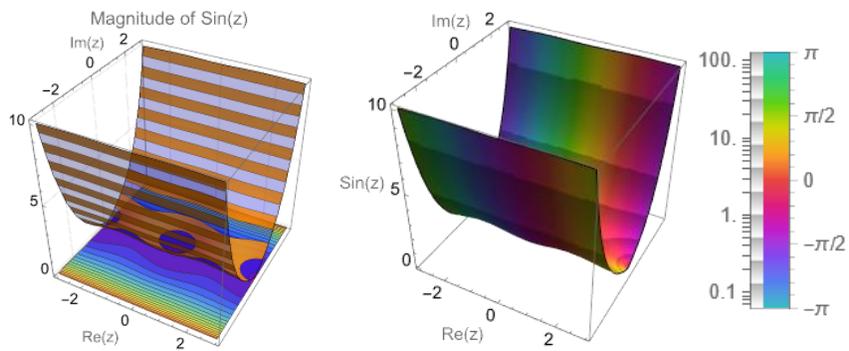

**Fig. 42.** Same as Fig. 35 but for the function $\sigma_{\text{FCSin}}(z)$.

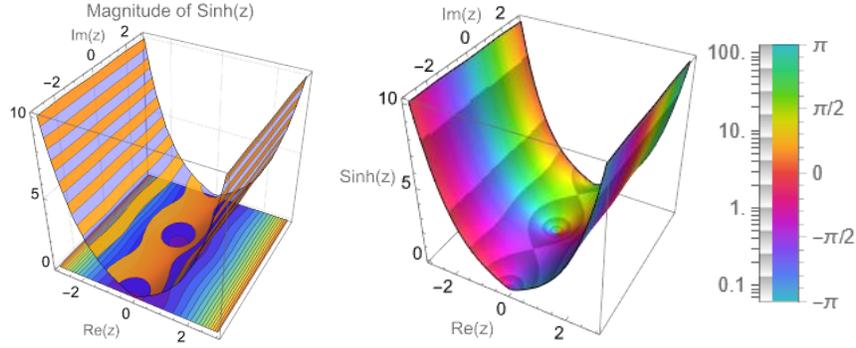

**Fig. 43.** Same as Fig. 35 but for the function $\sigma_{\text{FCSinh}}(z)$.

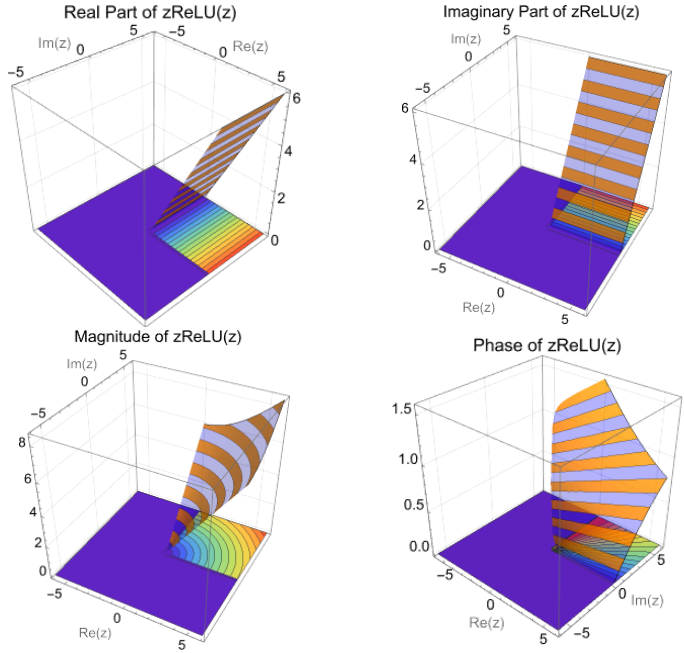

**Fig. 44.** Complex $\sigma_{\text{zReLU}}(z)$: real and imaginary parts, amplitude, and phase.

### 7.16 zReLU, z3ReLU, zPReLU and z3PReLU

The traditional ReLU function used in RVNNs is popular for introducing non-linearity by outputting the input if it is positive; otherwise, it outputs zero. Extending this concept to complex numbers allows for the incorporation of complex-valued inputs and weights in NNs. [17] proposed a complex-valued ReLU as:

$$\sigma_{\text{zReLU}}(z) = \begin{cases} z, & \text{Re}(z) > 0 \text{ and } \text{Im}(z) > 0 \\ 0, & \text{otherwise} \end{cases} = \begin{cases} z, & \arg(z) \in [0, \pi/2] \\ 0, & \text{otherwise} \end{cases}. \quad (129)$$

Figs 44 and 45 represent the real and imaginary parts, amplitude, and phase of $\sigma_{\text{zReLU}}(z)$. [75] proposed z3 Rectified Linear Unit (z3ReLU). The formula for the z3ReLU is given by:

$$\sigma_{\text{z3ReLU}}(z) = \begin{cases} z, & \text{Re}(z) > 0 \text{ or } \text{Im}(z) > 0 \\ 0, & \text{otherwise} \end{cases}. \quad (130)$$

When compared to similar CVAFs such as zReLU, the $\sigma_{\text{z3ReLU}}$ AF preserves more of the input magnitude and phase but still provides nonlinearity by rectifying inputs that have negative-valued real or imaginary segments, see Fig. 46.

Moreover, z Parametric Rectified Linear Unit (zPReLU) [76] CVAF was defined as follows:

$$\sigma_{\text{zPReLU}}(z) = \begin{cases} z, & \arg(z) \in [0, \pi/2] \\ \alpha z, & \text{otherwise} \end{cases}. \quad (131)$$

This nonlinearity is similar to zReLU. It has one complex-valued trainable parameter $\alpha$. The input, $z$, is multiplied with $\alpha$ if it does not lie in the first quadrant. Additionally, the z3 Parametric Rectified Linear Unit (z3PReLU) [76] AF is an extension of the previous $\sigma_{\text{zPReLU}}$, and it is defined as follows:

$$\sigma_{\text{z3PReLU}}(z) = \begin{cases} z, & \arg(z) \in [0, \pi/2] \\ \alpha_1 z, & \arg(z) \in (\pi/2, \pi] \\ \alpha_2 z, & \arg(z) \in (\pi, 3\pi/2] \\ \alpha_3 z, & \arg(z) \in (3\pi/2, 2\pi] \end{cases}. \tag{132}$$

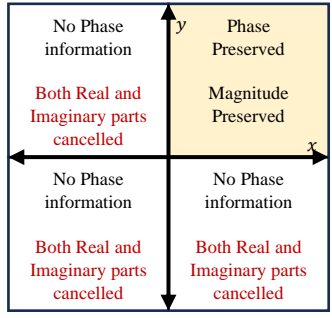

**Fig. 45.** Phase information encoding for the zReLU AF. The $x$-axis represents the real part and the $y$-axis axis represents the imaginary part. The complex representation is discriminated into two regions, i.e., the one that preserves the whole complex information (colored in orange) and the one that cancels it (colored in white).

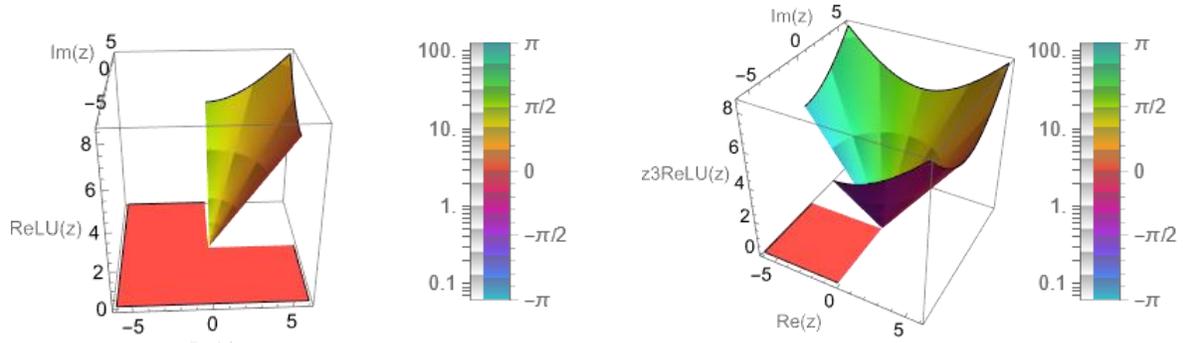

**Fig. 46.** Left panel: The ComplexPlot3D generates a 3D plot of Abs[$\sigma_{\text{zReLU}}(z)$] colored by arg[$\sigma_{\text{zReLU}}(z)$] over the complex rectangle with corners $z_{min} = -6 - 6i$ and $z_{max} = 6 + 6i$. Additionally, the color shading is achieved using the "CyclicLogAbsArg" function to cyclically shade colors and give the appearance of contours of constant Abs[$\sigma_{\text{zReLU}}(z)$] and constant arg[$\sigma_{\text{zReLU}}(z)$]. Right panel: Same as left figure but for the function $\sigma_{\text{z3ReLU}}(z)$.

This nonlinearity has three complex-valued trainable parameters. The input, $z$, is multiplied with different complex numbers depending on its quadrant.

### 7.17 Fully Complex Exponential Function

The ETFs used in fully complex NNs may exhibit sensitivity to weight initializations and learning rates. Additionally, these functions may suffer from singularities in the finite region of the complex plane, which can complicate the training process. An exponential function was proposed as a CVAF for the nonlinear processing of complex-valued data [16, 77].

The exponential function,

$$\sigma_{\text{FCExp}}(z) = e^z, \tag{133}$$

is entire in $\mathbb{C}$ since $\sigma_{\text{Exp}}(z) = \sigma'_{\text{FCExp}}(z) = e^z$.

A graphical representation (see Figs 47 and 48) provides a visual understanding of the behavior of the complex-valued exponential function. The plot illustrates the real and imaginary parts, amplitude, and phase of the AF. Notably, the amplitude is shown as continuous, increasing, and bounded within a specific region of the complex domain, underscoring its suitability for NN applications.

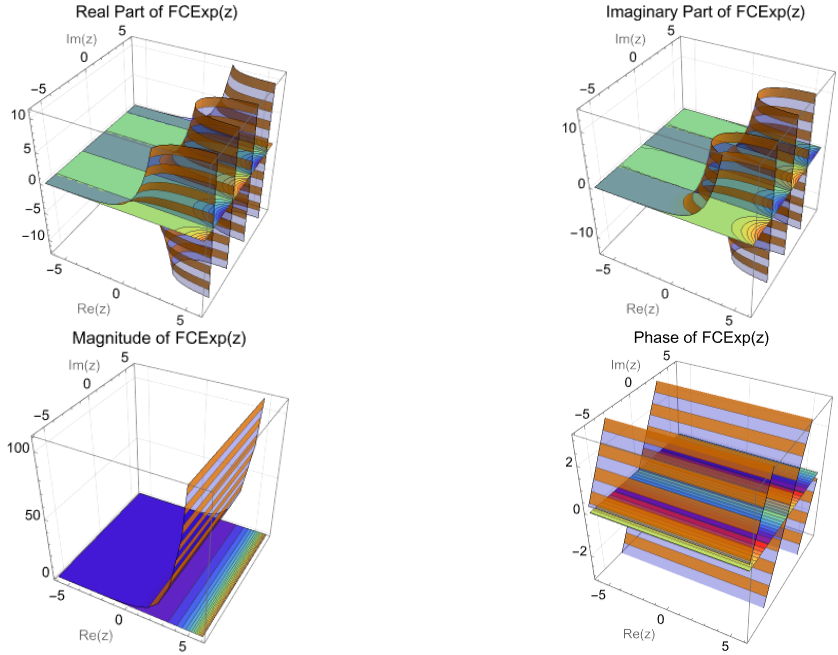

**Fig. 47.** Complex $\sigma_{\text{FCExp}}(z)$ as a function of complex variables, real-part, imaginary-part, amplitude, and phase.

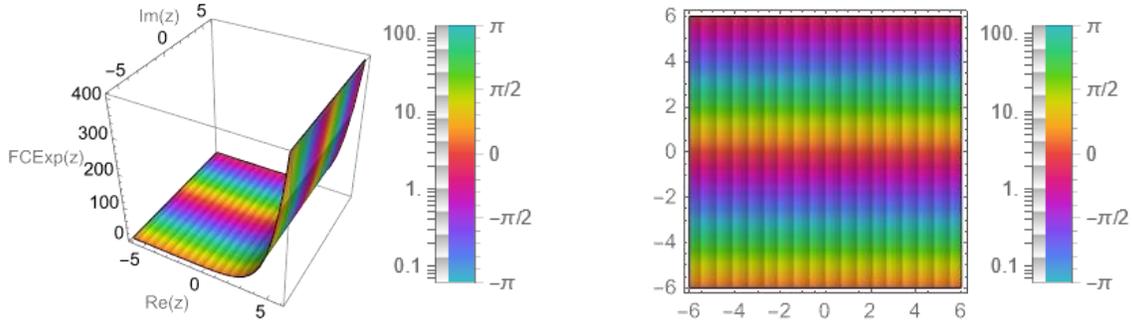

**Fig. 48.** Left panel: The ComplexPlot3D generates a 3D plot of Abs[$\sigma_{\text{FCExp}}(z)$] colored by arg[$\sigma_{\text{FCExp}}(z)$] over the complex rectangle with corners $z_{min} = -6 - 6i$ and $z_{max} = 6 + 6i$. Additionally, the color shading is achieved using the "CyclicLogAbs" function to cyclically shade colors and give the appearance of contours of constant Abs[$\sigma_{\text{FCExp}}(z)$]. Right panel: ComplexPlot generates a plot of arg[$\sigma_{\text{FCExp}}(z)$] over the complex rectangle with corners $-6 - 6i$ and $6 + 6i$, with using the "CyclicLogAbs" function.

The complex-valued exponential function exhibits an essential singularity at $\pm\infty$. To mitigate the challenges associated with these singularities, it is strategic to constrain the weights of the fully complex NN. By limiting the weights to a small ball of radius $r$ and controlling the number of hidden neurons, the network achieves a bounded behavior, thereby ensuring numerical stability during the training process.

## 8. New Complex Activation Functions

This section explores the extension of popular real-valued AFs—Exponential Linear Unit (ELU), Mish, SoftPlus, and Swish—to their complex counterparts, introducing Split-ELU, Split-Mish, Split-SoftPlus, and Split-Swish. These CVAFs operate by separately applying the real-valued AFs to the real and imaginary parts of the complex input, then recombining them into a complex output. This approach preserves the essential properties of the original AFs while enhancing control over phase and amplitude modifications, which are crucial in many complex-valued applications. We further generalize the real-valued AFs Mish and Swish to their fully complex counterparts, introducing Fully Complex Mish (FCMish), and Fully Complex Swish (FCSwish). Additionally, we propose a novel set of CVAFs designed to modulate the amplitude of complex inputs while preserving their phase, ensuring the output remains a valid complex number. The following sections provide a detailed explanation of these extendeds and novel CVAFs, their mathematical formulations, properties, and visual representations.

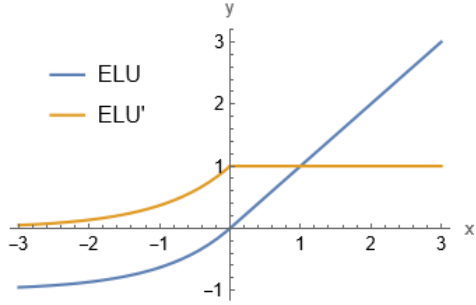

**Fig. 49.** The figure illustrates the ELU AF alongside its derivative, plotted over a range of $x$ from $-3$ to 3. The ELU function is designed to improve the convergence rate during training and reduce the vanishing gradient problem compared to traditional ReLU. It transitions from an exponential growth for negative values (where $\sigma_{ELU}(x) = \alpha(\exp(x) - 1)$ for $x < 0$) to a linear relationship for non-negative values (where $\sigma_{ELU}(x) = x$ for $x \geq 0$). The derivative of ELU shows how the rate of change adapts across different values of $x$.

## 8.1 ELU

Why ELU AF? Besides producing sparse codes, the main advantage of ReLU is that it alleviates the vanishing gradient problem since the derivative of 1 for positive values is not contractive. However, ReLU is non-negative and, therefore, has a mean activation larger than zero. Units with non-zero mean activations effectively act as biases for the next layer. If these non-zero means are not canceled out or balanced, it can lead to a bias shift effect as learning progresses through the network. The more correlated the units with non-zero mean activations are, the higher the potential for bias shift. Correlated activations can amplify each other's impact, leading to a cumulative effect on the bias shift in subsequent layers. AFs play a crucial role in shaping the distribution of activations. Good AFs aim to push activation means closer to zero. This property helps in reducing bias shift effects and contributes to more stable and efficient learning in NNs.

Centering activations around zero is known to speed up learning in NNs. This is because weight updates during training are influenced by the gradients of the AFs. When activations are centered, gradients are more balanced, and the optimization process is generally more effective. Batch Normalization (BN) is a technique that aims to address issues related to internal covariate shift. It normalizes the activations in a mini-batch, centering them around zero and scaling them to have a certain variance. This normalization can contribute to mitigating bias shift effects and accelerating training. Moreover, Projected Natural Gradient Descent is an optimization algorithm that leverages the natural gradient, which is the gradient of the loss function with respect to the model parameters, adjusted by the Fisher information matrix. The use of the natural gradient helps in more efficient optimization. Additionally, Projected Natural Gradient Descent involves projecting the parameters to ensure that the activations are centered, contributing to faster learning.

Using an AF that naturally pushes the mean activation toward zero is an alternative approach to achieve the benefits of centered activations. AFs that inherently maintain activations around zero can help mitigate bias shift and facilitate more stable learning in NNs. For example: (1) Tanh has been preferred over logistic functions. (2) Leaky ReLU (LReLU) that replaces the negative part of the ReLU with a linear function have been shown to be superior to ReLU. (3) Parametric ReLU (PReLU) generalizes LReLU by learning the slope of the negative part which yielded improved learning behavior on large image benchmark data sets. (4) Randomized ReLU (RReLU) which randomly samples the slope of the negative part raised the performance on image benchmark datasets and convolutional networks.

The real-valued ELU has negative values to allow for mean activations close to zero, but which saturates to a negative value with smaller arguments. ELU with $0 < \alpha$ [78] is given as,

$$\sigma_{ELU}(x) = \begin{cases} x, & x > 0, \\ \alpha(e^x - 1), & x \leq 0. \end{cases} \tag{134.1}$$

$$\frac{\partial}{\partial x}\sigma_{ELU}(x) = \begin{cases} 1, & x > 0 \\ \alpha e^x, & x \leq 0 \end{cases} = \begin{cases} 1, & x > 0, \\ \sigma_{ELU}(x) + \alpha, & x \leq 0, \end{cases} \tag{134.2}$$

having the output range in $[-1, \infty)$ where $\alpha$ is a learnable parameter. The ELU hyperparameter $\alpha$ controls the value to which an ELU saturates for negative net inputs (see Fig. 49). If $x$ keeps reducing past zero, eventually, the output of the ELU will be capped at $-1$, as the limit of $e^x$ as $x$ approaches negative infinity is 0. The value for $\alpha$ is chosen to control what we want this cap to be regardless of how low the input gets. This is called the saturation point. At the saturation point, and below, there is very little difference in the output of this function, and hence there's little to no variation (differential) in the information delivered from this node to the other node in the forward propagation.

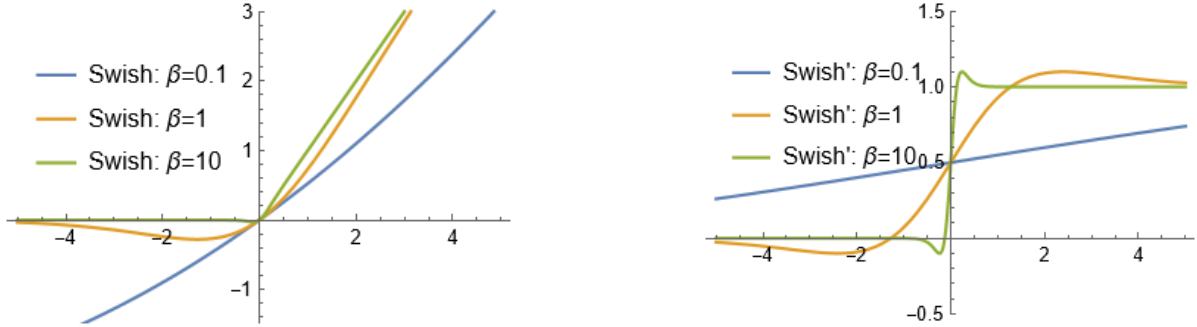

**Fig. 50.** Left panel: This plot visualizes the Swish AF, a smooth, non-monotonic function, for three different values of the parameter $\beta$ (0.1, 1, and 10) over a range of $x$ from $-5$ to $5$. The function is defined as $x\,\sigma_{\text{Sigmoid}}(\beta x)$. As $\beta$ increases, the Swish function transitions from a nearly linear behavior ($\beta = 0.1$) to a more pronounced ReLU-like curve ($\beta = 10$), illustrating its adaptability and the impact of $\beta$ on the activation output. Right panel: The plot shows the first derivatives of the Swish function for the same $\beta$ values (0.1, 1, and 10). Each derivative curve illustrates the rate of change of the Swish function at different points, highlighting how the responsiveness of the function varies with changes in $\beta$. For lower $\beta$, the derivative remains close to constant, while for higher $\beta$, the curve exhibits sharper changes, especially near zero.

The formulation of ELU involves an exponential term for negative inputs, which allows the network to capture information from both positive and negative sides of the input space, contributing to its ability to maintain mean activations close to zero. Note that, the $\alpha$ parameter in the ELU equation controls the slope of the function for negative inputs. It's often set to a small positive value like 1.0, but it can be tuned to suit the specific problem. The choice of $\alpha$ can impact the behavior of ELU, and in practice, it's often chosen through experimentation.

Key characteristics and benefits of the ELU AF include its ability to alleviate the vanishing gradient problem, similar to ReLUs, LReLUs, and PReLUs, by maintaining identity for positive values. ELUs offer improved learning characteristics compared to other AFs, as their negative values push mean unit activations closer to zero, akin to batch normalization but with lower computational complexity. This shift towards zero speeds up learning. While LReLUs and PReLUs also have negative values, they do not ensure a noise-robust deactivation state. ELUs, on the other hand, saturate to a negative value with smaller inputs, reducing forward propagated variation and information. ELUs help mitigate the "dying ReLU" problem, which occurs when ReLU units output zero for certain inputs, causing neurons to become inactive and impeding learning. ELU neurons avoid this issue by producing non-zero outputs for negative inputs. Moreover, ELU can handle outliers better than ReLU, as the exponential term allows it to capture extreme values without saturating (Robust to outliers). Additionally, ELU can lead to faster convergence and potentially better generalization performance compared to some other AFs, especially when dealing with complex datasets.

### 8.2 Swish

The introduction of Swish was part of a broader effort to explore and discover new AFs. The authors of the paper "Searching for AFs" [79] used a neural architecture search approach to automatically discover AFs that performed well on certain tasks. The paper contributed to the ongoing research in the deep learning community by highlighting the potential benefits of automatic methods for discovering NN architectures and components. The introduction of Swish and similar approaches demonstrated that there might be room for improvement beyond traditional AFs, and automatic search methods could be valuable in this exploration.

The real-valued Swish [79] is defined as
$$\sigma_{\text{Swish}}(x) = x\,\sigma_{\text{Sigmoid}}(\beta x), \tag{135}$$
where $\sigma_{\text{Sigmoid}}(x) = (1 + \exp(-x))^{-1}$ is the Sigmoid function and $\beta$ is either a constant or a trainable parameter. The output range of Swish is $(-\infty, \infty)$. Swish is a smooth and differentiable AF. Smoothness and differentiability are desirable properties in NN AFs because they facilitate gradient-based optimization during the training process. Fig. 50 plots the graph of Swish for different values of $\beta$. If $\beta = 0$, Swish becomes the scaled linear function $f(x) = \frac{x}{2}$. As $\beta \to \infty$, the Sigmoid component approaches a $0-1$ function, so Swish becomes like the ReLU function.

This suggests that Swish can be loosely viewed as a smooth function which nonlinearly interpolates between the linear function and the ReLU function. The degree of interpolation can be controlled by the model if $\beta$ is set as a trainable parameter. Like ReLU, Swish is unbounded above and bounded below. Unlike ReLU, Swish is smooth and non-monotonic. In fact, the non-monotonicity property of Swish distinguishes it from most common AFs.

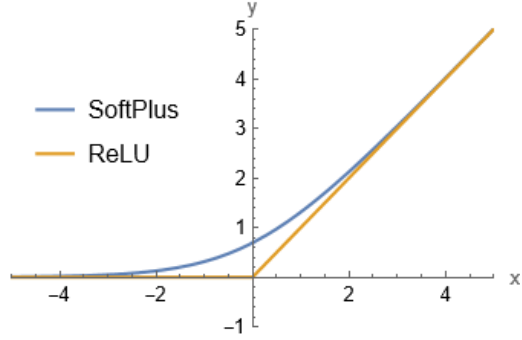

**Fig. 51.** The figure presents a comparative view of two AFs: SoftPlus, and ReLU, plotted over a range of $x$ from $-5$ to $5$. SoftPlus offers a smooth, gradual transition, acting as a continuously differentiable approximation of the ReLU function, which directly clamps all negative values to zero and linearly passes positive values.

The derivative of Swish is

$$\frac{\partial}{\partial x}\sigma_{\text{Swish}}(x) = \sigma_{\text{Sigmoid}}(\beta x) + x\,\beta\,\sigma_{\text{Sigmoid}}(\beta x)\left(1 - \sigma_{\text{Sigmoid}}(\beta x)\right)$$
$$= \beta\,\sigma_{\text{Swish}}(x) + \sigma_{\text{Sigmoid}}(\beta x)\left(1 - \sigma_{\text{Swish}}(x)\right). \quad (136)$$

The first derivative of Swish is shown in Fig. 50 for different values of $\beta$. The scale of $\beta$ controls how fast the first derivative asymptotes to $0$ and $1$.

When $\beta = 1$, the derivative has magnitude less than $1$ for inputs that are less than around $1.25$. Thus, the success of Swish with $\beta = 1$ implies that the gradient preserving property of ReLU (i.e., having a derivative of $1$ when $x > 0$) may no longer be a distinct advantage in modern architectures.

The most striking difference between Swish and ReLU is the non-monotonic "bump" of Swish when $x < 0$. The shape of the bump can be controlled by changing the $\beta$ parameter. The smaller and higher values of $\beta$ lead towards the linear and ReLU functions, respectively. Thus, it can control the amount of non-linearity based on the dataset and network complexity.

While Swish showed promise in terms of performance improvements, it is important to consider the computational cost. Swish involves the computation of the Sigmoid function, which might be more computationally expensive compared to simpler AFs like ReLU.

*8.3 SoftPlus*

The real-valued SoftPlus function [80] was proposed in 2001 and is mostly used in statistical applications. SoftPlus unit-based AF is also used in Deep Neural Networks (DNNs) [81]. As a smoothing version of the ReLU function, (as illustrated in Fig. 51), the SoftPlus function is defined as:

$$\sigma_{\text{SoftPlus}}(x) = \log(e^x + 1). \quad (137.1)$$

The derivative of the SoftPlus function with respect to its input $x$ can be computed as follows:

$$\frac{\partial}{\partial x}\sigma_{\text{SoftPlus}}(x) = \frac{e^x}{1 + e^x} = \sigma_{\text{Sigmoid}}(x). \quad (137.2)$$

This derivative can be helpful in training NNs using gradient-based optimization algorithms like backpropagation.

The SoftPlus has a number of advantages. It is smooth and differentiable everywhere, which makes it well-suited for gradient-based optimization techniques like backpropagation. This property makes the SoftPlus function more stable no matter when being estimated from the positive and negative directions, while ReLU has a discontinuous gradient at point $0$. The SoftPlus function maps its input to a range between $0$ and positive infinity, which can be useful in certain situations. It is a monotonic function, meaning that as the input $x$ increases, the output also increases.

Unlike some other AFs, such as the Sigmoid or Tanh, the SoftPlus function does not saturate for large positive inputs. Saturated AFs can lead to vanishing gradients and slow down the learning process. Additionally, the SoftPlus function has a non-zero gradient even when the input is negative. Unlike ReLU, which propagates no gradient for $x < 0$, SoftPlus can propagate gradients across all real inputs. This feature prevents the "dying ReLU" problem, where neurons become inactive and do not update their weights during training. Furthermore, the SoftPlus unit also outperforms the Sigmoid unit with the following aspects. The derivative of SoftPlus is a Sigmoid function. It means that the gradient of the SoftPlus unit approaches 1 when the input increases, which largely reduces the bad effects of the vanishing gradient problem.

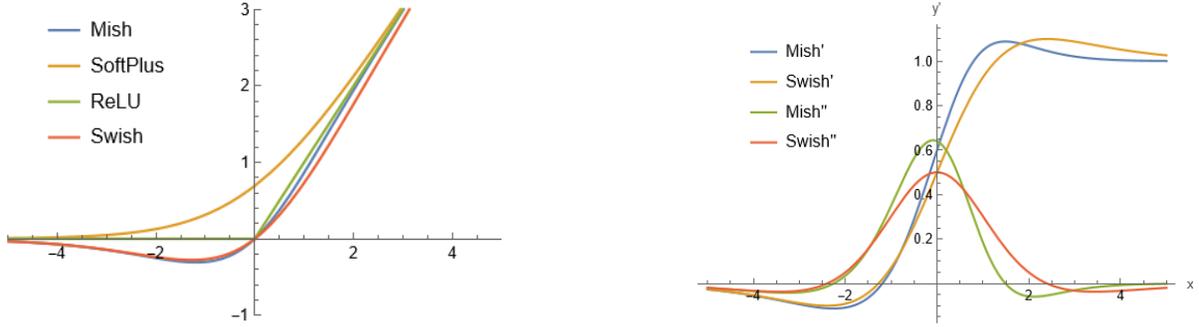

**Fig. 52.** Left panel: The figure illustrates a comparison of four advanced AFs: Mish, SoftPlus, ReLU, and Swish, each plotted over a range of $x$ from $-5$ to 5. Mish, defined as $x \operatorname{Tanh}(\ln(1 + e^x))$, showcases a smooth and non-monotonic curve that closely resembles the behavior of Swish but with subtle differences in negative input handling. SoftPlus, a smooth approximation of ReLU, shows a gentle logistic-like transition, emphasizing its continuous and differentiable nature. ReLU remains the simplest, with a direct zero-clamping behavior for negative inputs and linear response for positive values. Swish, defined as $x \cdot \sigma_{\text{Sigmoid}}(\beta x)$ with $\beta = 1$, combines aspects of linear and sigmoidal responses, providing a flexible shape that varies with the input value. Right panel: The figure presents a detailed comparison of the first and second derivatives of the Mish and Swish AFs, plotted over a range of $x$ from $-5$ to 5. The first derivatives, labeled as "Mish'" and "Swish'", show the rate of change of the respective AFs, illustrating how they respond to different input values. The second derivatives, labeled as "Mish''" and "Swish''", indicate the curvature or the rate of change of the slope of the AFs, providing deeper insights into their dynamic behavior.

It is important to note that, while there may be concerns about the hard saturation of ReLU and the resulting zero gradients, experimental evidence [82] suggests that this characteristic might not hinder supervised training as much as initially thought. The use of SoftPlus as an alternative does not necessarily provide a clear advantage in all cases, and the benefits of hard zeros in ReLU might outweigh the drawbacks in certain scenarios, particularly when some hidden units remain active during training. The hard non-linearities do not hurt so long as the gradient can propagate along some paths, i.e., that some of the hidden units in each layer are non-zero.

### *8.4 Mish*

The SoftPlus function is also used with the Tanh function in Mish AF [83]. The real-valued Mish, as visualized in Fig. 52, is a smooth, continuous, self-regularized, non-monotonic AF. Mathematically, it was defined as:

$$\sigma_{\text{Mish}}(x) = x \operatorname{Tanh}\big(\sigma_{\text{SoftPlus}}(x)\big) = x \operatorname{Tanh}(\ln(1 + e^x)). \tag{138.1}$$

Similar to Swish, Mish is bounded below and unbounded above with a range of $[\approx -0.31, \infty)$. The derivative of Mish, as shown in Fig. 52, can be defined as:

$$\frac{\partial}{\partial x}\sigma_{\text{Mish}}(x) = \frac{\omega e^x}{\delta}, \tag{138.2}$$

where, $\omega = 4(x + 1) + 4e^{2x} + e^{3x} + e^x(4x + 6)$ and $\delta = 2e^x + e^{2x} + 2$.

The Mish AF offers several advantages. It preserves a small amount of negative information, thereby eliminating the preconditions necessary for the Dying ReLU phenomenon and improving expressivity and information flow. Being unbounded above, Mish avoids saturation, which typically slows down training due to near-zero gradients. Additionally, its boundedness below results in strong regularization effects. Unlike ReLU, Mish is continuously differentiable, avoiding singularities and undesired side effects during gradient-based optimization. Mish's non-monotonic nature helps capture complex and non-linear relationships in data. Similar to the SoftPlus function, Mish is smooth and differentiable everywhere, making it well-suited for gradient-based optimization methods like backpropagation. The function also has a built-in self-regularization property, resisting very large input values and helping to avoid exploding gradients during training. It has demonstrated competitive performance in DNNs compared to other popular AFs, such as ReLU and LReLU, in various experimental settings. However, the increased complexity of Mish, due to its multiple functions, can be a limitation for DNNs.

### *8.5 Split: ELU, Mish, SoftPlus, and Swish*

Now, we extend the popular real-valued AFs ELU, Mish, SoftPlus, and Swish to their complex counterparts: Split-ELU, Split-Mish, Split-SoftPlus, and Split-Swish. These functions split the real and imaginary parts of the complex input, apply the AF separately to each part, and then combine them back into a complex number.

$$\sigma_{\text{Split-ELU}}(z) = \sigma_{\text{ELU}}[\Re(z)] + i\sigma_{\text{ELU}}[\Im(z)], \tag{139.1}$$
$$\sigma_{\text{Split-Mish}}(z) = \sigma_{\text{Mish}}[\Re(z)] + i\sigma_{\text{Mish}}[\Im(z)], \tag{139.2}$$
$$\sigma_{\text{Split-SoftPlus}}(z) = \sigma_{\text{SoftPlus}}[\Re(z)] + i\sigma_{\text{SoftPlus}}[\Im(z)], \tag{139.3}$$
$$\sigma_{\text{Split-Swish}}(z) = \sigma_{\text{Swish}}[\Re(z)] + i\sigma_{\text{Swish}}[\Im(z)]. \tag{139.4}$$

The split approach offers several benefits, including enhanced control over phase and amplitude modifications, crucial for domains where these elements carry significant information. For instance, Split-ELU (Figs. 53 and 54) uses the ELU function to provide a response for negative inputs, aiding in gradient stability and learning efficiency. Split-Mish (Figs. 55 and 56), with its non-monotonic smoothness, manages extreme values better, enhancing the network's ability to handle outliers. Split-SoftPlus (Figs. 57 and 58) offers a continuously differentiable approximation of ReLU, making it ideal for maintaining smooth gradient flow in complex-valued calculations. Lastly, Split-Swish (Figs. 59 and 60) incorporates a self-gating mechanism that regulates gradient flow, adding a form of light regularization beneficial for complex data.

The Split-ELU, Split-Mish, Split-SoftPlus, and Split-Swish CVAFs exhibit line symmetry with respect to both the real and imaginary axes. This symmetry means that these functions' behavior can be mirrored across these axes, simplifying their analysis and understanding. The real part of these AFs is symmetric with respect to the real axis ($\Im(z) = 0$). This means if you replace a complex number $z_1 = (a, b)$ with $z_2 = (a, -b)$, where $a$ and $b$ are the real and imaginary parts, respectively, the real part of the function remains unchanged. Mathematically, if $\sigma$ is the CVAF, then: $\Re(\sigma(a, b)) = \Re(\sigma(a, -b))$. This implies that the behavior of the real part of the function does not change when the imaginary part of the input is negated. This property can be particularly useful when analyzing or visualizing the real part of these functions because it reduces the problem to considering only one half of the imaginary plane. On the other hand, the symmetry of the imaginary part of these functions with respect to the imaginary axis ($\Re(z) = 0$) implies that if you replace $z_1 = (a, b)$ with $z_2 = (-a, b)$ the imaginary part of these functions value remains the same. The symmetry in these CVAFs implies that the neural dynamics associated with these CVAFs have special significance along the real and imaginary axes, $\Im(z) = 0$ and $\Re(z) = 0$. Along these axes, the behavior of the CVAFs is more predictable and structured. By knowing that the function behaves symmetrically with respect to both the real and imaginary axes, the complexity of studying the function's behavior is reduced. This simplification can be particularly beneficial in both theoretical analysis and practical applications, such as NN training and optimization, where understanding the behavior of AFs is crucial.

### *8.6 Fully Complex: Swish and Mish*

We define Fully Complex Swish (FCSwish) (Fig. 61) as
$$\sigma_{\text{FCSwish}}(z) = z\,\sigma_{\text{Sigmoid}}(z), \tag{140}$$
where $\sigma_{\text{Sigmoid}}(z) = (1 + \exp(-z))^{-1}$ is the Sigmoid function. The singularities of the FCSwish function occur at specific points in the complex plane. To understand these singularities, we first examine the conditions under which the Logistic Sigmoid function becomes undefined or singular. The singularities of $\sigma_{\text{Sigmoid}}(z)$ arise when its denominator equals zero, which happens when $1 + \exp(-z) = 0$. Solving for $z$, we get $\exp(-z) = -1$, leading to $-z = \ln(-1)$. In the complex plane, $\ln(-1) = i\pi$ for the principal branch, but due to the periodic nature of the complex exponential function, the logarithm can also be expressed as $i\pi + 2i\pi n$, where $n$ is any integer. Therefore, the singularities of the Sigmoid function occur at $z = i\pi + 2i\pi n$. Applying this to the FCSwish function, which includes the Logistic Sigmoid function as a factor, we find that the singularities of the FCSwish function also occur at these points. Thus, the singularities of the FCSwish function can be expressed as $z = i\pi + 2i\pi n$ for $n \in \mathbb{Z}$. This means that the FCSwish function has an infinite number of singularities, periodically spaced along the imaginary axis at intervals of $2i\pi$.

Similarly, we define Fully Complex Mish (FCMish) (Fig. 61) as
$$\sigma_{\text{FCMish}}(z) = z\,\text{Tanh}(\ln(1 + e^z)), \tag{141}$$
The singularities of the FCMish function occur at specific points in the complex plane, $z = 2i\pi n + \log[-1 \pm i]$, for $n \in \mathbb{Z}$.

While the singular points of the FCSwish and FCMish functions theoretically exist, they do not typically interfere with the training of NNs. This is because the domain of interest is usually bounded within a small circle, making the occurrence of these problematic points rare and manageable. Moreover, during the initial stages of training, the weights and biases are usually initialized to small random values, ensuring that the initial activations are also small. As training progresses, regularization techniques and learning rate schedules help keep the activations within a desirable range, further minimizing the risk of encountering singularities. If the network's parameters do start to produce inputs near these singular points, mechanisms like gradient clipping, normalization, and careful monitoring of the training process can be employed to mitigate potential issues.

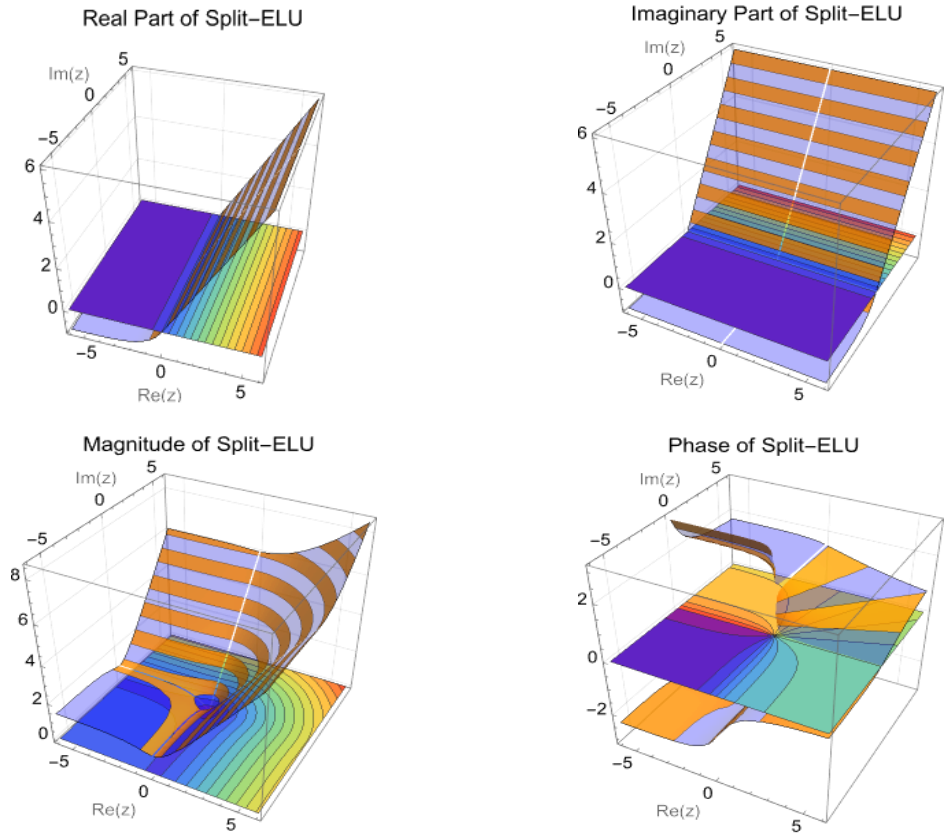

**Fig. 53.** Visualizations of the real, imaginary, magnitude, and phase parts of the Split-ELU function.

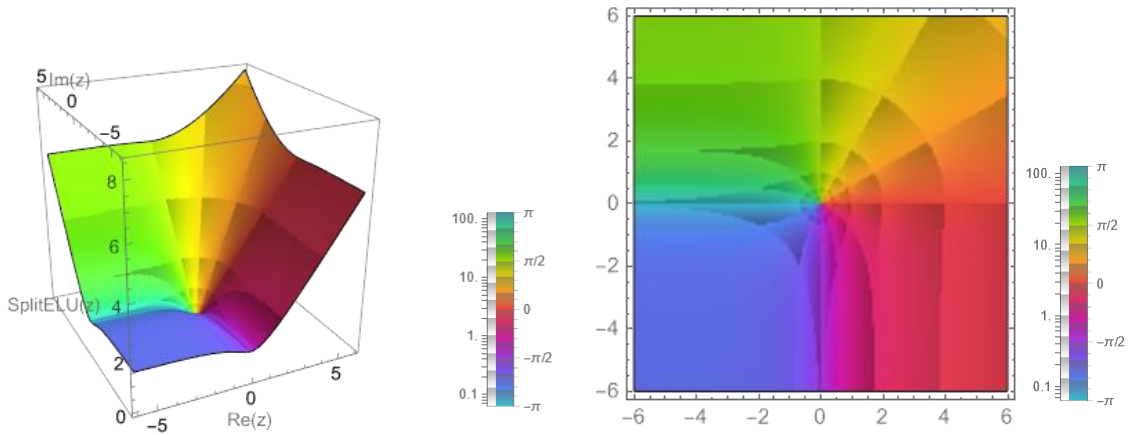

**Fig. 54.** Left panel: The ComplexPlot3D generates a 3D plot of Abs[$\sigma_{\text{Split-ELU}}(z)$] colored by arg[$\sigma_{\text{Split-ELU}}(z)$] over the complex rectangle with corners $z_{min} = -6 - 6i$ and $z_{max} = 6 + 6i$. Using "CyclicLogAbsArg" to cyclically shade colors to give the appearance of contours of constant Abs[$\sigma_{\text{Split-ELU}}(z)$] and constant arg[$\sigma_{\text{Split-ELU}}(z)$]. Right panel: ComplexPlot generates a plot of arg[$\sigma_{\text{Split-ELU}}(z)$] over the complex rectangle with corners $-6 - 6i$ and $6 + 6i$. Using "CyclicLogAbsArg" to cyclically shade colors to give the appearance of contours of constant Abs[$\sigma_{\text{Split-ELU}}(z)$] and constant arg[$\sigma_{\text{Split-ELU}}(z)$].

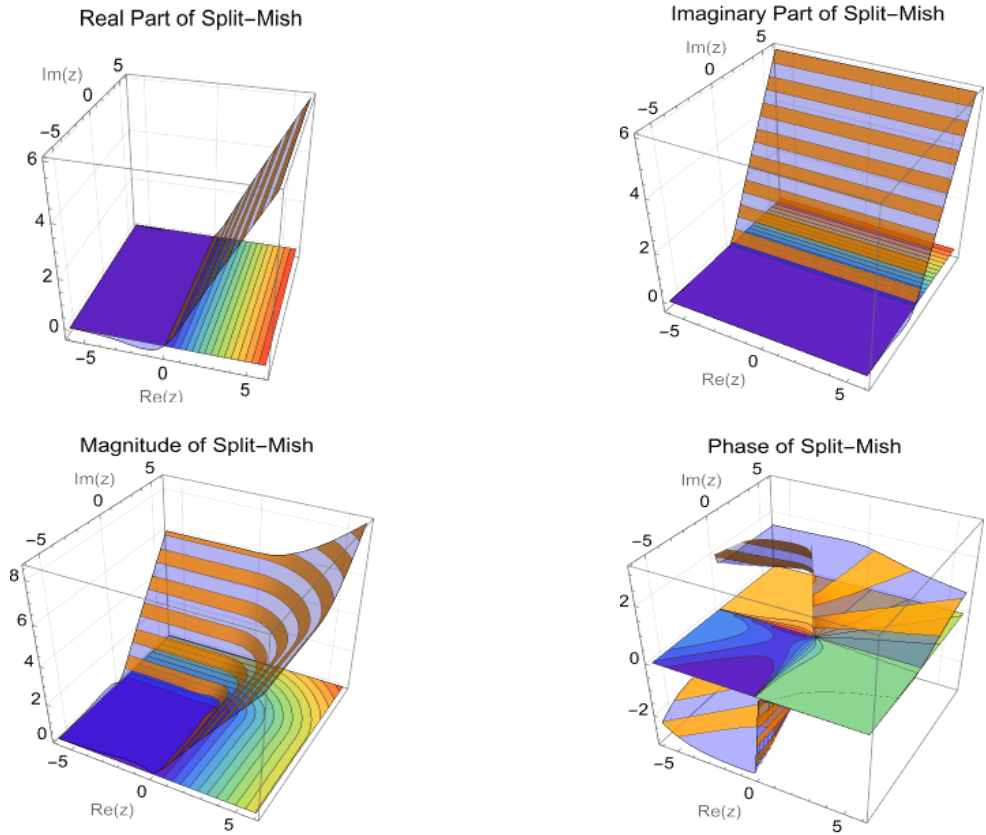

**Fig. 55.** Visualizations of the real, imaginary, magnitude, and phase parts of the Split-Mish function.

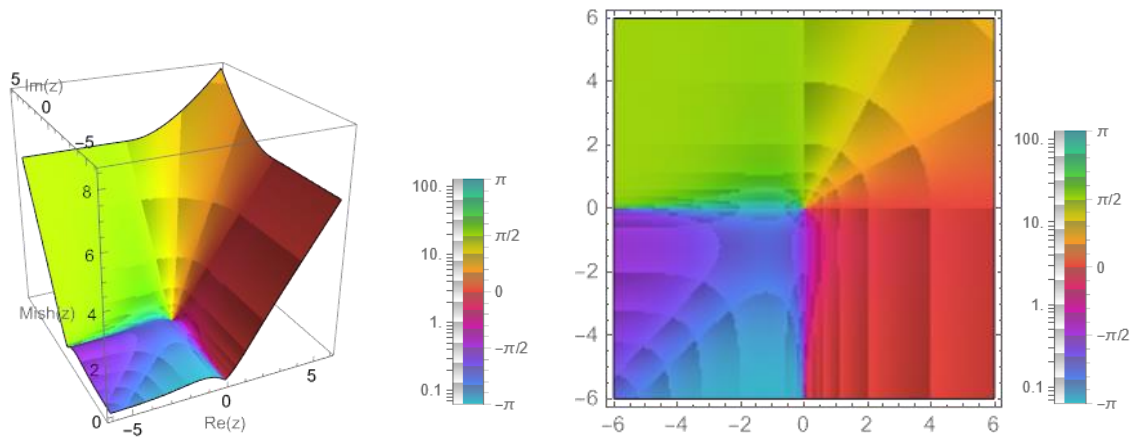

**Fig. 56.** Left panel: The ComplexPlot3D generates a 3D plot of Abs[$\sigma_{\text{Split-Mish}}(z)$] colored by arg[$\sigma_{\text{Split-Mish}}(z)$] over the complex rectangle with corners $z_{min} = -6 - 6i$ and $z_{max} = 6 + 6i$. Using "CyclicLogAbsArg" to cyclically shade colors to give the appearance of contours of constant Abs[$\sigma_{\text{Split-Mish}}(z)$] and constant arg[$\sigma_{\text{Split-Mish}}(z)$]. Right panel: ComplexPlot generates a plot of arg[$\sigma_{\text{Split-Mish}}(z)$] over the complex rectangle with corners $-6 - 6i$ and $6 + 6i$. Using "CyclicLogAbsArg" to cyclically shade colors to give the appearance of contours of constant Abs[$\sigma_{\text{Split-Mish}}(z)$] and constant arg[$\sigma_{\text{Split-Mish}}(z)$].

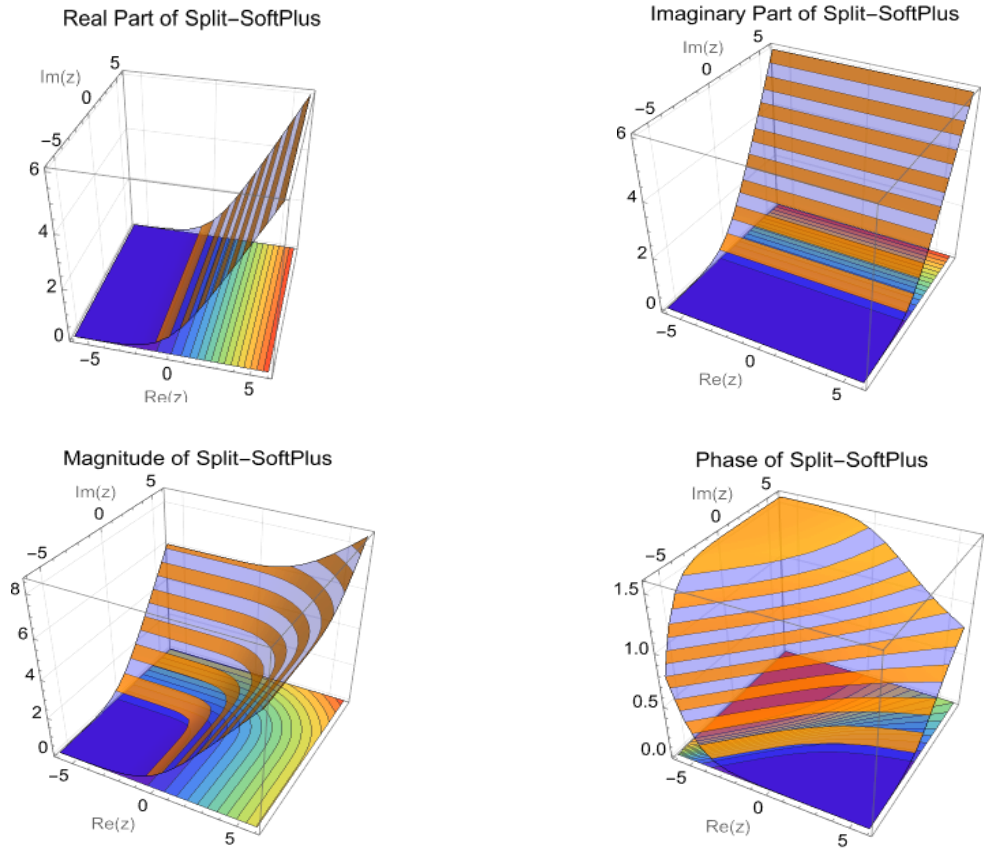

**Fig. 57.** Visualizations of the real, imaginary, magnitude, and phase parts of the Split-SoftPlus function.

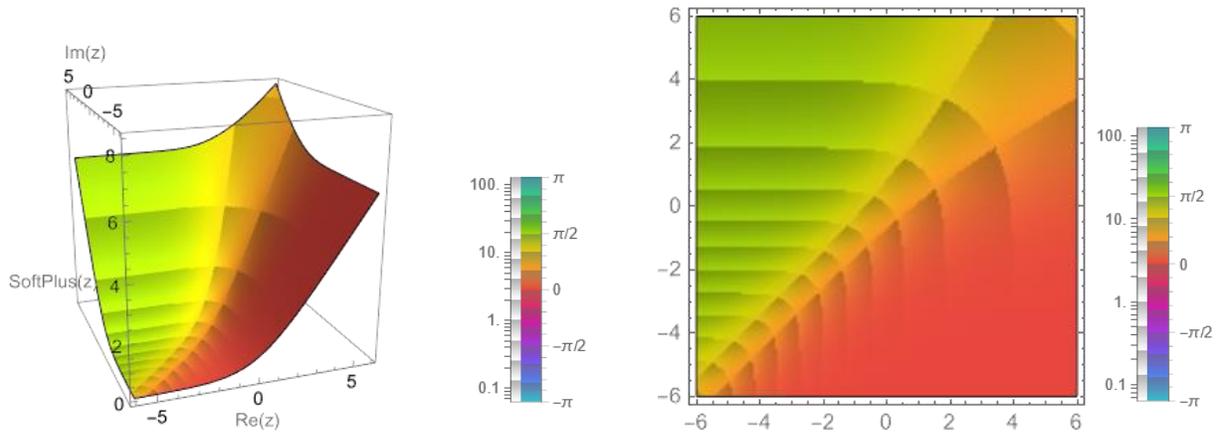

**Fig. 58.** Left panel: The ComplexPlot3D generates a 3D plot of Abs[$\sigma_{\text{Split-SoftPlus}}(z)$] colored by arg[$\sigma_{\text{Split-SoftPlus}}(z)$] over the complex rectangle with corners $z_{min} = -6 - 6i$ and $z_{max} = 6 + 6i$. Using "CyclicLogAbsArg" to cyclically shade colors to give the appearance of contours of constant Abs[$\sigma_{\text{Split-SoftPlus}}(z)$] and constant arg[$\sigma_{\text{Split-SoftPlus}}(z)$]. Right panel: ComplexPlot generates a plot of arg[$\sigma_{\text{Split-SoftPlus}}(z)$] over the complex rectangle with corners $-6 - 6i$ and $6 + 6i$. Using "CyclicLogAbsArg" to cyclically shade colors to give the appearance of contours of constant Abs[$\sigma_{\text{Split-SoftPlus}}(z)$] and constant arg[$\sigma_{\text{Split-SoftPlus}}(z)$].

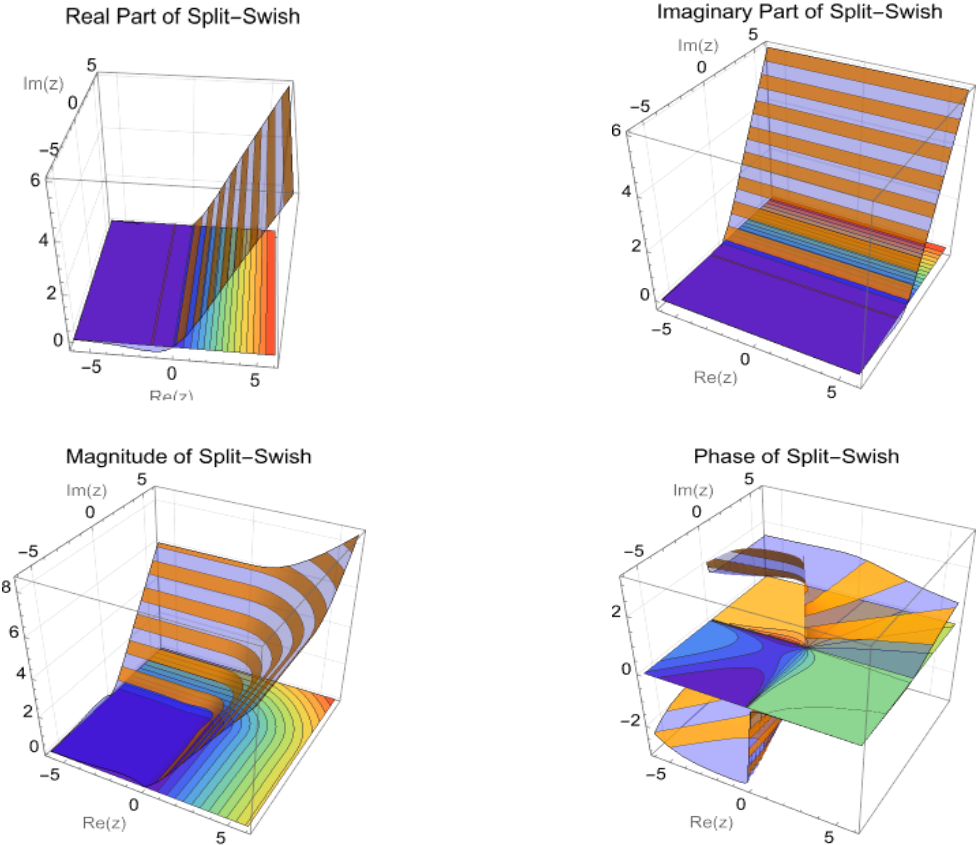

**Fig. 59.** Visualizations of the real, imaginary, magnitude, and phase parts of the Split-SoftPlus function.

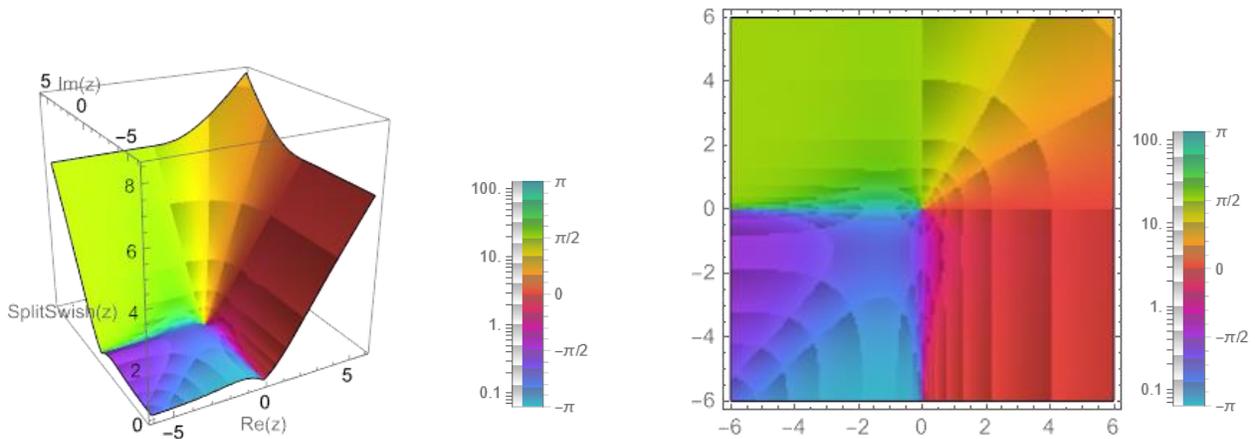

**Fig. 60.** Left panel: The ComplexPlot3D generates a 3D plot of Abs[$\sigma_{\text{Split-SoftPlus}}(z)$] colored by arg[$\sigma_{\text{Split-SoftPlus}}(z)$] over the complex rectangle with corners $z_{min} = -6 - 6i$ and $z_{max} = 6 + 6i$. Using "CyclicLogAbsArg" to cyclically shade colors to give the appearance of contours of constant Abs[$\sigma_{\text{Split-SoftPlus}}(z)$] and constant arg[$\sigma_{\text{Split-SoftPlus}}(z)$]. Right panel: ComplexPlot generates a plot of arg[$\sigma_{\text{Split-SoftPlus}}(z)$] over the complex rectangle with corners $-6 - 6i$ and $6 + 6i$. Using "CyclicLogAbsArg" to cyclically shade colors to give the appearance of contours of constant Abs[$\sigma_{\text{Split-SoftPlus}}(z)$] and constant arg[$\sigma_{\text{Split-SoftPlus}}(z)$].

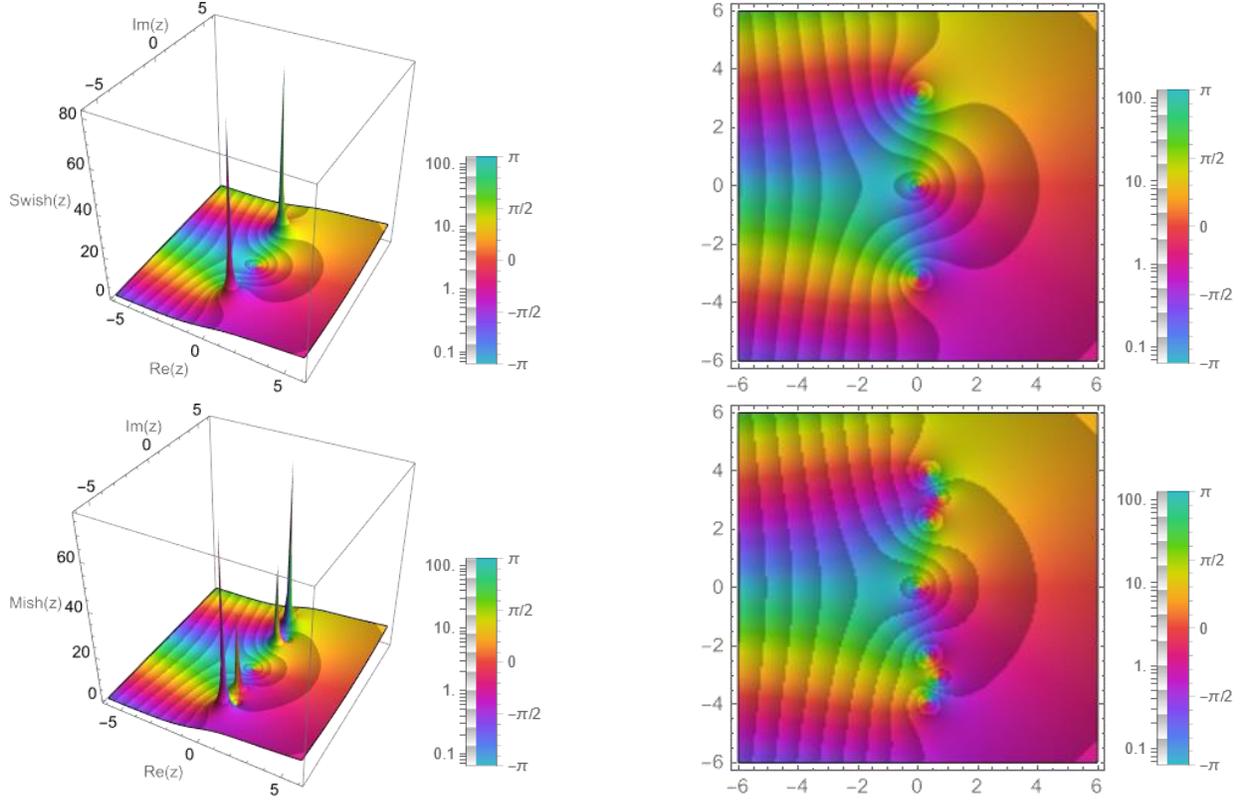

**Fig. 61.** Top panel (left): The ComplexPlot3D generates a 3D plot of Abs[$\sigma_{\text{FCSwish}}(z)$] colored by arg[$\sigma_{\text{FCSwish}}(z)$] over the complex rectangle with corners $z_{min} = -6 - 6i$ and $z_{max} = 6 + 6i$. Using "CyclicLogAbsArg" to cyclically shade colors to give the appearance of contours of constant Abs[$\sigma_{\text{FCSwish}}(z)$] and constant arg[$\sigma_{\text{FCSwish}}(z)$]. Top panel (right): ComplexPlot generates a plot of arg[$\sigma_{\text{FCSwish}}(z)$] over the complex rectangle with corners $-6 - 6i$ and $6 + 6i$. Using "CyclicLogAbsArg" to cyclically shade colors to give the appearance of contours of constant Abs[$\sigma_{\text{FCSwish}}(z)$] and constant arg[$\sigma_{\text{FCSwish}}(z)$]. Bottom panel: Same as top panel but for the function $\sigma_{\text{FCMish}}(z)$.

### 8.7 New Amplitude-Phase-Type Functions

In addition to extending existing real-valued AFs to complex domains, we introduce a set of novel CVAFs designed to enhance the performance and stability of CVNNs. These functions modulate the amplitude of the input complex number while preserving its phase, ensuring the output remains a valid complex number.

### 8.7.1 Complex Amplitude-Phase Piecewise Linear Scaling (CAP-PLS)

The CAP-PLS AF is defined as (Fig. 62):

$$\sigma_{\text{CAPPLS}}(z) = \min[|z|, a] \frac{z}{|z|}, \tag{142.1}$$

where $a$ is positive parameter. For a complex number $z$ such that $|z| < a$, the function $\sigma_{\text{CAPPLS}}(z)$ returns the original complex number $z$. However, for a complex number $z$ such that $|z| > a$, the function scales the complex number $z$ to have an absolute value of $a$ while maintaining its direction (phase). This behavior can be interpreted as a "cap" on the magnitude of $z$. If the magnitude of $z$ exceeds a certain threshold $a$, the function reduces the magnitude to $a$, preserving the direction. One of the main advantages of the CAP-PLS function is its computational efficiency. Unlike the APTF function, which can be computationally expensive (using Tanh), the CAP-PLS function relies on simple linear operations (min and absolute value), making it faster to compute. This efficiency can be particularly beneficial in scenarios where computational resources are limited or when training large NNs.

### 8.7.2 Complex Amplitude-Phase Exponential Scaling (CAP-ES)

The CAP-ES AF is formulated as (Fig. 63):

$$\sigma_{\text{CAPES}}(z) = \left(1 - e^{-|z|}\right) \frac{z}{|z|}. \tag{142.2}$$

For any complex number $z$, the function $\sigma_{\text{CAPES}}(z)$ modulates the magnitude of $z$ based on the exponential decay while preserving its direction. When $|z|$ is small, $e^{-|z|}$ is close to 1, making $1 - e^{-|z|}$ close to 0, and thus $\sigma_{\text{CAPES}}(z)$ is also small. As $|z|$ increases, $e^{-|z|}$ approaches 0, making $1 - e^{-|z|}$ approach 1, and $\sigma_{\text{CAPES}}(z)$ approximates $z$ normalized to unit magnitude. This formulation emphasizes saturation in amplitude while keeping the phase unchanged.

*8.7.3 Complex Amplitude-Phase ArcTan Scaling (CAP-ArcTanS)*

The CAP-ArcTanS AF is represented as (Fig. 64):

$$\sigma_{\text{CAPArcTanS}}(z) = \text{ArcTan}(|z|)\frac{z}{|z|}. \tag{142.3}$$

The function $\sigma_{\text{CAPArcTanS}}(z)$ modulates the magnitude of $z$ through $\text{ArcTan}(|z|)$ while keeping the direction unchanged. When $|z|$ is small, $\text{ArcTan}(|z|) \approx |z|$, making $\sigma_{\text{CAPArcTanS}}(z)$ approximately equal to $z$. As $|z|$ increases, $\sigma_{\text{CAPArcTanS}}(z)$ asymptotically approaches $\pi/2$, causing the magnitude of $\sigma_{\text{CAPArcTanS}}(z)$ to grow more slowly and ultimately saturate at this value. This behavior ensures controlled amplitude growth, which can be useful in preventing the explosion of values during processing while maintaining the phase information intact.

*8.7.4 Complex Amplitude-Phase Erf Attenuation (CAP-ErfA)*

The CAP-CErfA AF is (Fig. 65):

$$\sigma_{\text{APCErfA}}(z) = \text{erf}(|z|)\frac{z}{|z|}. \tag{142.4}$$

The function $\sigma_{\text{APCErfA}}(z)$ modulates the magnitude of $z$ through $\text{erf}(|z|)$ while preserving its direction. For small values of $|z|$, $\text{erf}(|z|) = |z|$, making $\sigma_{\text{APCErfA}}(z)$ approximately equal to $z$. As $|z|$ increases, $\text{erf}(|z|)$ approaches 1, causing the magnitude of $\sigma_{\text{APCErfA}}(z)$ to saturate and preventing it from growing indefinitely. This characteristic ensures that the amplitude of $\sigma_{\text{APCErfA}}(z)$ remains within a bounded range, thus controlling the growth of the magnitude while maintaining the phase information.

*8.7.5 Complex Amplitude-Phase Softplus (CAP-Softplus or modSoftplus)*

The CAP-Softplus AF or modSoftplus is mathematically specified as (Fig. 66):

$$\sigma_{\text{CAPSoftplus}}(z) = \sigma_{\text{modSoftplus}}(z) = \text{Log}\left[1 + e^{(|z|-a)}\right]\frac{z}{|z|}. \tag{142.5}$$

The function $\sigma_{\text{CAPSoftplus}}(z)$ modifies the magnitude of $z$ through $\text{Log}\left[1 + e^{(|z|-a)}\right]$ while keeping the direction unchanged. This transformation ensures that the amplitude grows smoothly and never becomes negative, which is particularly useful for stability in NN training. For values of $|z|$ close to $a$, $e^{(|z|-a)}$ is small, and thus $\text{Log}\left[1 + e^{(|z|-a)}\right] \approx \text{Log}[2]$, leading to a moderated amplitude. For large values of $|z|$, the term $\text{Log}\left[1 + e^{(|z|-a)}\right]$ asymptotically approaches $|z| - a$. It is a smoothing version of the modReLU function. While modReLU creates a hard threshold where the function output is zero for negative values of $|z| + b$ and retains the adjusted amplitude for non-negative values, CAP-Softplus smooths this transition. CAP-Softplus avoids the sharp transitions and potential instability of modReLU by using the logarithmic function to provide a smooth and continuous growth of the amplitude.

*8.7.6 Complex Amplitude-Phase Exponential Linear Unit (CAP-ELU or modELU)*

The CAP-ELU AF or modELU is defined as (Fig. 67):

$$\sigma_{\text{CAPELU}}(z) = \sigma_{\text{modELU}}(z) = \begin{cases} (|z| + b)\dfrac{z}{|z|}, & |z| < -b, \\ \alpha(e^{|z|+b} - 1)\dfrac{z}{|z|}, & \text{otherwise,} \end{cases} \tag{142.6}$$

where $b$ is negative parameter, $b < 0$. This function combines linear and exponential terms to modulate the magnitude of $z$. For $|z| < -b$, the function modifies the amplitude by simply adding $b$ to $|z|$. This ensures that for small magnitudes, the amplitude is shifted by $b$ but the phase of $z$ is preserved. For $|z| > -b$, the function applies an exponential transformation. This allows the amplitude to grow exponentially with $|z|$, modulated by the parameter $\alpha$, while still preserving the phase of $z$. As $|z|$ is always positive, a bias $b$ is introduced to create a "Modulation Region" of radius $b$ around the origin 0 where the amplitude is shifted by $b$, and outside of which the amplitude is grow exponentially with $|z|$. This behaver like modReLU AF where the ReLU operation $\sigma_{\text{ReLU}}(|z| + b) = \max(|z| + b, 0)$ ensures that the function becomes zero for inputs $(|z| + b < 0)$ and retains the amplitude-modulated $z$ for inputs $(|z| + b \geq 0)$. In modReLU AF, as $|z|$ is always positive, a bias $b$ is introduced in order to create a

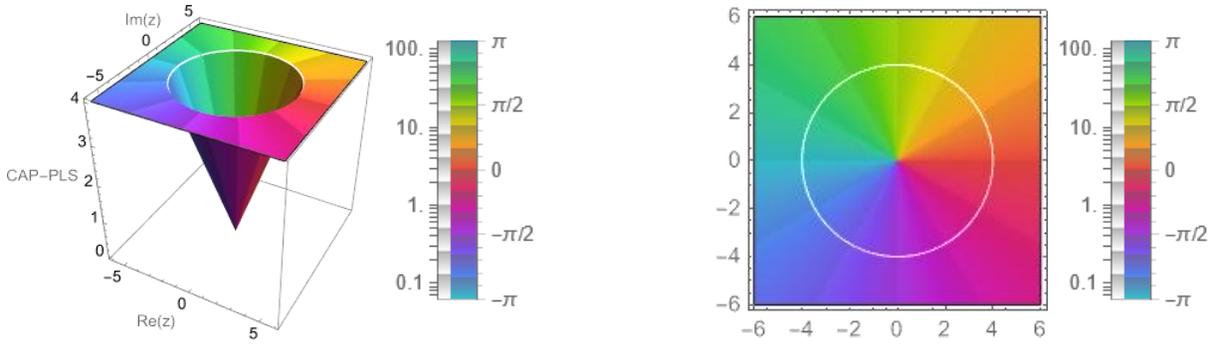

**Fig. 62.** Top panel: The ComplexPlot3D generates a 3D plot of Abs[$\sigma_{CAP-PLS}(z)$] colored by arg[$\sigma_{CAP-PLS}(z)$] over the complex rectangle with corners $z_{min} = -6 - 6i$ and $z_{max} = 6 + 6i$. Using "CyclicLogAbsArg" to cyclically shade colors to give the appearance of contours of constant Abs[$\sigma_{CAP-PLS}(z)$] and constant arg[$\sigma_{CAP-PLS}(z)$]. Top panel (right): ComplexPlot generates a plot of arg[$\sigma_{CAP-PLS}(z)$] over the complex rectangle with corners $-6 - 6i$ and $6 + 6i$. Using "CyclicLogAbsArg" to cyclically shade colors to give the appearance of contours of constant Abs[$\sigma_{CAP-PLS}(z)$] and constant arg[$\sigma_{CAP-PLS}(z)$].

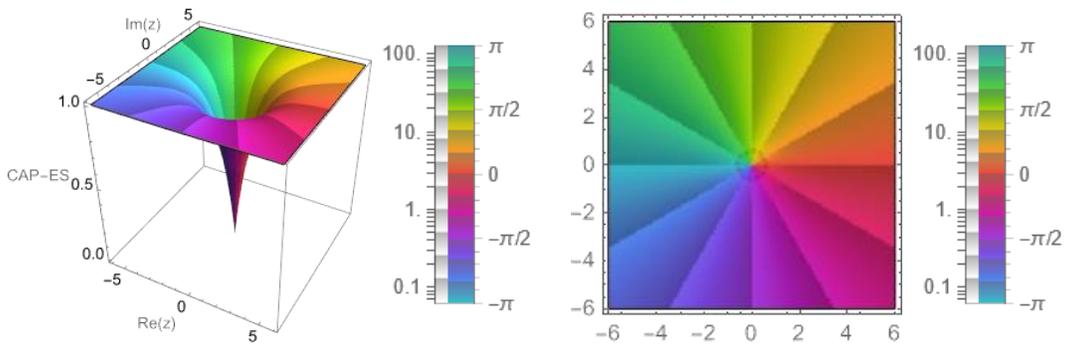

**Fig. 63.** Same as Fig. 62 but for the function $\sigma_{CAP-ES}(z)$.

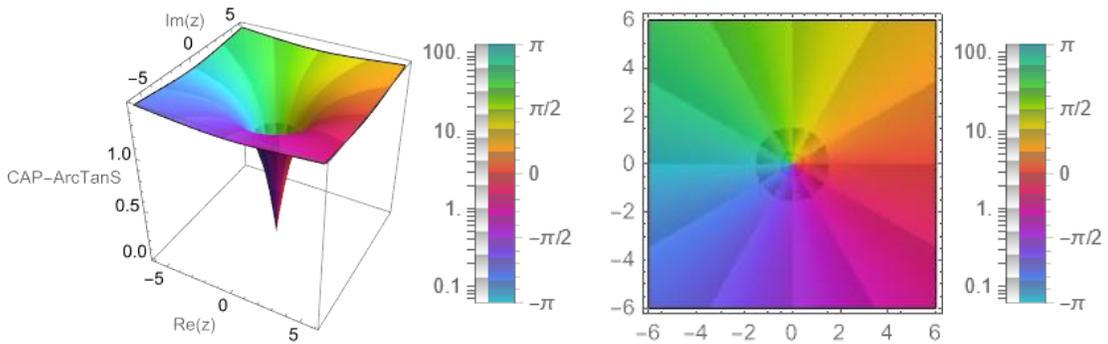

**Fig. 64.** Same as Fig. 62 but for the function $\sigma_{CAP-ArcTanS}(z)$.

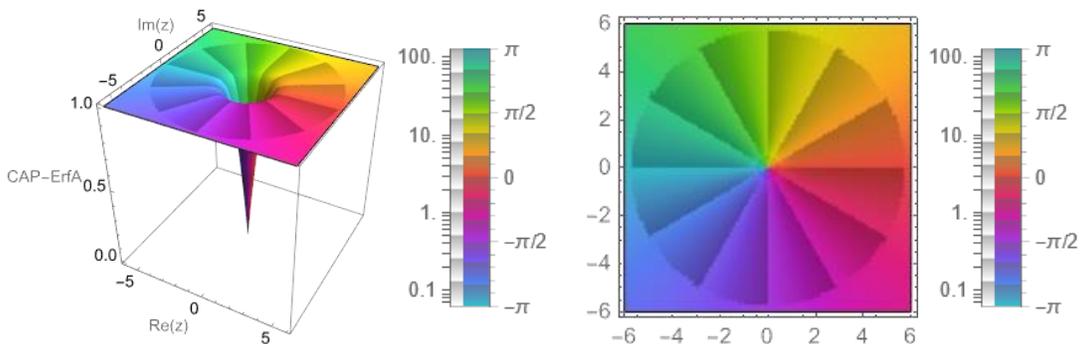

**Fig. 65.** Same as Fig. 62 but for the function $\sigma_{CAP-ErfA}(z)$.

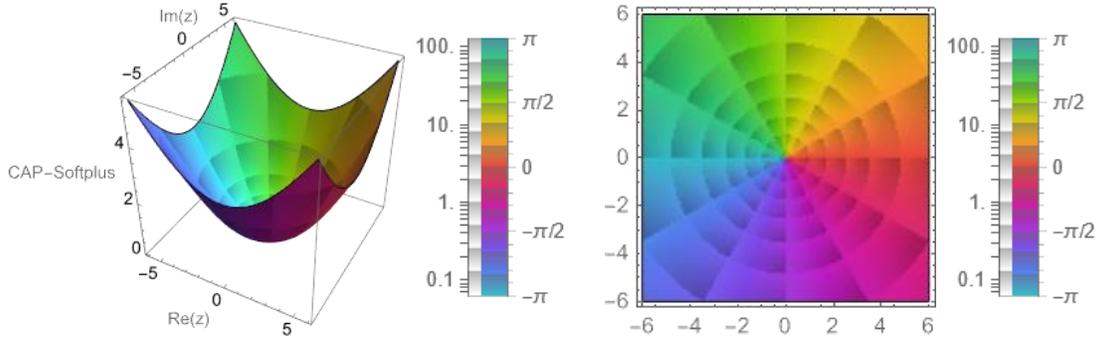

**Fig. 66.** Same as Fig. 62 but for the function $\sigma_{CAP-SoftPlus}(z)$.

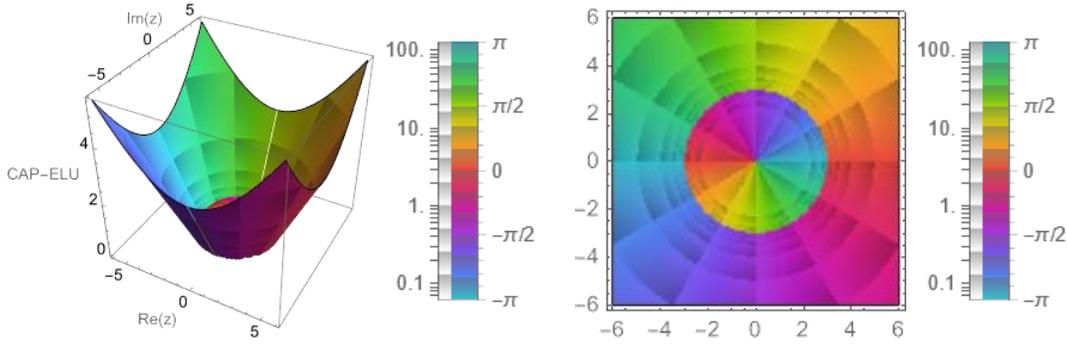

**Fig. 67.** Same as Fig. 62 but for the function $\sigma_{CAP-ELU}(z)$.

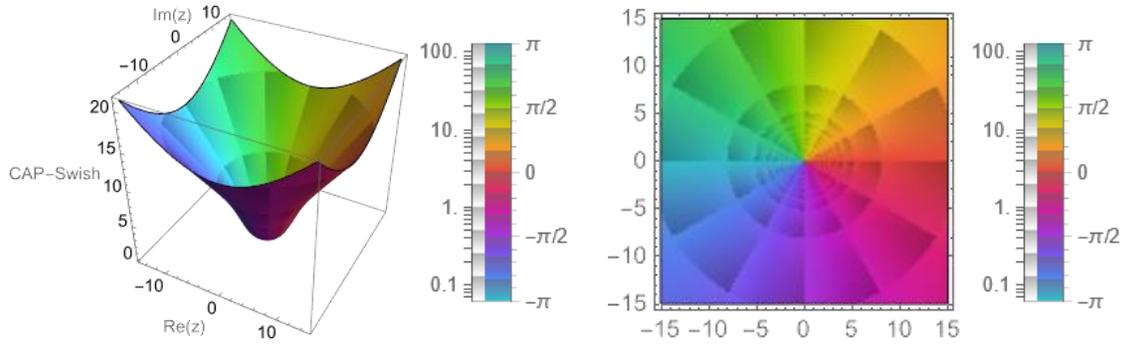

**Fig. 68.** Same as Fig. 62 but for the function $\sigma_{CAP-Swish}(z)$.

"dead zone" of radius $b$ around the origin $0$ where the neuron is inactive, and outside of which it is active. The CAP-ELU AF, however, goes a step further by combining linear and exponential terms. This combination allows the network to leverage the benefits of both linear adjustments for small magnitudes and exponential growth for larger magnitudes. The formulation of CAP-ELU allows the network to capture information from both $|z| < -b$ and $|z| > -b$ of the input space, contributing to its ability to maintain mean activations close to zero.

*8.7.7 Complex Amplitude-Phase Swish (CAP-Swish or modSwish)*

The CAP-Swish AF or modSwish is given by (Fig. 68):

$$\sigma_{CAPSwish}(z) = \frac{|z|}{1 + e^{-|z|+b}} \frac{z}{|z|}. \tag{142.7}$$

When $|z|$ is small, the term $e^{-|z|+b}$ is close to $e^b$, making $\frac{|z|}{1+e^{-|z|+b}} \approx \frac{|z|}{1+e^b}$, thus moderating the amplitude for small magnitudes. As $|z|$ increases, $e^{-|z|+b}$ approaches zero, and $\frac{|z|}{1+e^{-|z|+b}}$ approximates $|z|$, allowing the amplitude to grow smoothly and without abrupt changes. As $|z|$ is always positive, a bias $b$ is introduced to create a "Modulation Region" of radius $b$ around the origin 0 where the amplitude is moderating, and outside of which the amplitude is grow linearly with $|z|$. This behaver like modReLU AF where the ReLU operation $\sigma_{\text{ReLU}}(|z|+b) = \max(|z|+b, 0)$ ensures that the function becomes zero for inputs ($|z|+b < 0$) and retains the amplitude-modulated $z$ for inputs ($|z|+b \geq 0$).

CAP-Swish (or modSwish) is a versatile AF that combines the benefits of smoothness, non-linearity, and directional consistency. Its design, incorporating a sigmoid-like modulation around zero and linear growth for larger inputs, makes it suitable for various CVNN architectures and training scenarios.

In general, the above proposed AFs share several important properties that make them well-suited for use in CVNNs. All functions preserve the phase of the input $z$ by multiplying the modulated magnitude with $z/|z|$, maintaining the complex nature of the input. Most functions ensure that the output is bounded or grows slowly with increasing $|z|$, preventing the propagation of excessively large values and contributing to the stability of the training process.

## 9. Conclusion

In conclusion, this paper provides an extensive survey of CVNNs, highlighting the significant advancements in their AFs and learning algorithms. Despite the computational and implementation challenges associated with CVNNs, their potential to outperform RVNNs in various applications is undeniable. The survey covers the adaptation of the backpropagation algorithm to the complex domain, presenting three distinct CBP algorithms. These algorithms address the unique requirements of training NNs with complex-valued inputs, weights, and outputs. Furthermore, the paper emphasizes the crucial role of CVAFs in the design and training of CVNNs. It explores both fully complex AFs, which strive for boundedness and differentiability, and split AFs, which offer a practical compromise despite not preserving analyticity.

Moreover, this survey not only provides a comprehensive overview of the current state of CVNNs but also introduces new CVAFs, contributing to the ongoing research and development in this field. The paper extends popular real-valued AFs—ELU, Mish, SoftPlus, and Swish—into their complex counterparts: Split-ELU, Split-Mish, Split-SoftPlus, and Split-Swish. These CVAFs apply the real-valued AFs separately to the real and imaginary parts of the complex input, preserving the original properties while enhancing phase and amplitude control. We also introduce FCMish and FCSwish, generalizing Mish and Swish to their fully complex forms. Additionally, we propose novel CVAFs that modulate amplitude while preserving phase. We provide a comprehensive analysis of the mathematical formulations, properties, and visual representations of these extended and novel CVAFs. As the field progresses, the insights and advancements discussed in this paper will serve as a resource for researchers and practitioners aiming to leverage the power of CVNNs.